\documentclass[11pt,letterpaper]{article}
\usepackage[T1]{fontenc}
\usepackage[margin=1in]{geometry}
\usepackage{amsmath,amssymb}
\usepackage{graphicx}
\usepackage{microtype}
\usepackage{needspace}
\usepackage{enumitem}
\setlist{topsep=4pt,itemsep=2pt,parsep=0pt}
\RequirePackage{newtxtext}
\RequirePackage[amsthm]{newtxmath}
\RequirePackage{bm}
\RequirePackage{endnotes}

\usepackage{booktabs}   %

\usepackage{algorithm,algorithmic}
\usepackage{caption} %
\usepackage{placeins} %
\usepackage{tikz}
\usetikzlibrary{arrows.meta,backgrounds,calc,decorations.pathreplacing,fit,positioning}

\tikzset{
	salenode/.style={draw=black!65, rounded corners=2pt, line width=.6pt,
		align=center, inner xsep=7pt, inner ysep=5pt, font=\small},
	salearrow/.style={-{Latex[length=2mm]}, draw=black!70, line width=.65pt},
	saleedge/.style={font=\scriptsize, align=center, text=black!80},
	saleverified/.style={salenode, fill=green!11, draw=green!42!black},
	salereportable/.style={salenode, fill=orange!14, draw=orange!58!black},
	saleseller/.style={salenode, fill=blue!8, draw=blue!48!black},
	saleoutcome/.style={salenode, fill=black!5, draw=black!55}
}

\makeatletter
\newenvironment{breakablealgorithm}
{%
	\Needspace{8\baselineskip}
\begin{center}
		\refstepcounter{algorithm}%
		\hrule height.8pt depth0pt \kern2pt%
		\renewcommand{\caption}[1]{%
			{\raggedright\textbf{\fnum@algorithm:}\space ##1\par}%
			\addcontentsline{loa}{algorithm}{\fnum@algorithm: ##1}%
			\kern2pt\hrule\kern2pt%
		}
	}
	{%
		\kern2pt\hrule\relax%
	\end{center}
}
\makeatother

\usepackage{natbib}
\bibpunct[, ]{(}{)}{,}{a}{}{,}%
\def\BIBand{and}%

\usepackage{hyperref}
\hypersetup{
	colorlinks = true,
	linkcolor  = blue,
	citecolor  = blue,
	urlcolor   = blue,
}

\theoremstyle{plain}
\newtheorem{theorem}{Theorem}
\newtheorem{lemma}{Lemma}

\newtheorem{assumption}{Assumption}
\theoremstyle{definition}
\newtheorem{remark}{Remark}

\newtheorem{definition}{Definition}

\newcommand{\supp}{\operatorname{supp}}
\newcommand{\Reg}{\operatorname{Regret}}

\hypersetup{pdftitle={SALE: Source-Aware Dynamic Pricing with Strategic Buyers},pdfauthor={Jinjiang Zou, Yang Li, Zemin Zheng},pdfsubject={Research preprint}}
\title{SALE: Source-Aware Dynamic Pricing with Strategic Buyers}
\author{\begin{tabular}{@{}ccc@{}}
\makebox[.30\textwidth]{Jinjiang Zou\textsuperscript{\dag}} & \makebox[.30\textwidth]{Yang Li\textsuperscript{\dag}} & \makebox[.30\textwidth]{Zemin Zheng\textsuperscript{\dag}}\\[2pt]
{\footnotesize \href{mailto:zjj22chuang2@mail.ustc.edu.cn}{\textcolor{black}{\texttt{zjj22chuang2@mail.ustc.edu.cn}}}} & {\footnotesize \href{mailto:tjly@ustc.edu.cn}{\textcolor{black}{\texttt{tjly@ustc.edu.cn}}}} & {\footnotesize \href{mailto:zhengzm@ustc.edu.cn}{\textcolor{black}{\texttt{zhengzm@ustc.edu.cn}}}}
\end{tabular}\\[.65em]
{\small \textsuperscript{\dag}School of Management, University of Science and Technology of China, Hefei 230026, China}}
\date{}
\begin{document}
\maketitle

\begin{abstract}
In modern commerce, firms and platforms typically use customer features from different sources to personalize prices:  some features are verified using seller records or third-party data,  while others are self-reported and may be strategically manipulated by buyers seeking lower prices.
In this paper,  we study the problem of source-aware dynamic pricing with strategic buyers, where the buyer's valuation is a sparse linear function of her high-dimensional features plus market noise with an unknown distribution.
We develop the Source-Aware Localized exploration and calibrated Exploitation (SALE) policy, which personalizes exploration using past learning and verified features, and corrects for strategic manipulation using verified and reported features during exploitation.
Our novel personalized design aims to narrow the range of prices explored, reducing unnecessary exploration at excessively high or low prices.
To support learning under this design, we derive a new population equation that enables sparse valuation parameter estimation and, in turn, nonparametric estimation of the unknown noise distribution, even when buyer features are dependent over time.
By uncovering the intrinsic sparse structure, our analysis establishes estimation guarantees for both learning tasks using the same exploration data.
With these guarantees,   we show that SALE achieves a sublinear expected regret bound, explicitly quantifying the benefit of using verified features to narrow exploration intervals.
Several numerical studies further demonstrate the appealing performance of SALE.
\end{abstract}


\section{Introduction}\label{sec:Intro}

Effective personalized pricing depends on accurately learning how buyer features affect willingness to pay
\citep{NambiarSimchiLeviWang2019, ElmachtoubGuptaHamilton2021, TangQiFangShi2025}. In the era of big data, advances in data collection and the growth of online marketplaces give firms and platforms
access to a large set of features for each buyer \citep{Javanmard2019,WangWangSunCheng2024, BanFanHanJagabathulaKeskinZhou2026}. Such high-dimensional buyer features are common in dynamic-pricing applications, including online retail \citep{Mikians2013,FisherGallinoLi2017}, consumer lending \citep{PhillipsSimsekVanRyzin2015,JiangWeiChanHamdi2022}, and insurance \citep{SoleymanianWeinbergZhu2019,GuelmanGuillenPerezMarin2014}. For example, a consumer-lending platform observes a borrower's credit and repayment histories, account activity, employment information, and responses to a detailed loan application.
When financial histories are tracked over time and detailed application information is collected, a borrower's profile can include hundreds or thousands of candidate features. The high-dimensional  features in these
profiles also come from different sources \citep{Guo2024Customization,WernerfeltTuchmanShapiroMoakler2024, DiMaggioRatnadiwakara2026}. Some buyer features are verified using platform records or third-party data, whereas others are reported directly by
buyers \citep{EllicksonKarReeder2022,Pursiainen2026}.

These differences in feature sources also affect buyers' opportunities for strategic manipulation. When features are self-reported, buyers seeking lower prices may strategically manipulate their reports at a cost
\citep{Bechavod2022,LiLi2022,LiuSun2026Fairness}. For example, an automobile insurance applicant may strategically underreport odometer mileage to seek a lower premium  \citep{GaiPuntoni2021}.
Verified features, by contrast, are checked against seller records or third-party data and cannot be altered through misreporting.
In the automobile insurance example,  insurers can also use verified risk classifications based on applicants' claims histories, rather than relying solely on applicants' self-reported information\citep{JeziorskiKrasnokutskayaCeccarini2019}.
From a pricing perspective,  verified features provide the seller with reliable information for assessing  buyers' likely valuations, while strategic reporting makes it harder to learn how true buyer features affect willingness to pay \citep{Liu2025}.
The seller must therefore protect valuation learning from strategic behavior while making full use of verified information to personalize prices. Figure~\ref{fig:source-information} below illustrates the source-dependent information flow in personalized pricing.

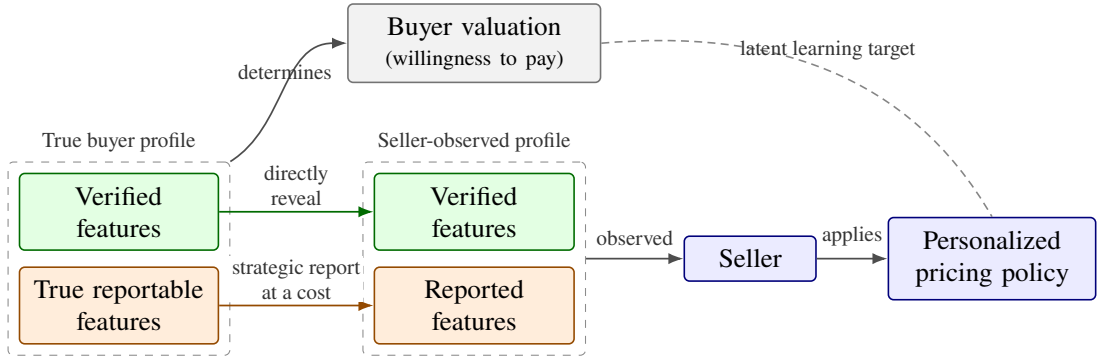
\begin{figure}[H]
\centering
\begin{tikzpicture}[x=1cm,y=1cm]
	\node[saleverified, text width=2.35cm, inner xsep=4pt]
		(true-verified) at (0,0.62)
		{Verified\\features};
	\node[salereportable, text width=2.35cm, inner xsep=4pt]
		(true-reportable) at (0,-0.62)
		{True reportable\\features};
	\node[draw=black!45, dashed, rounded corners=2pt,
		fit=(true-verified)(true-reportable), inner sep=4pt,
		label={[saleedge]above:True buyer profile}] (true-profile) {};

	\node[saleverified, text width=2.15cm]
		(observed-verified) at (4.70,0.62)
		{Verified\\features};
	\node[salereportable, text width=2.15cm]
		(reported) at (4.70,-0.62)
		{Reported\\features};
	\node[draw=black!45, dashed, rounded corners=2pt,
		fit=(observed-verified)(reported), inner sep=4pt,
		label={[saleedge]above:Seller-observed profile}]
		(observed-profile) {};

	\node[saleseller, text width=1.25cm]
		(seller) at (8.35,0)
		{Seller};
	\node[saleseller, text width=2.25cm]
		(policy) at (11.55,0)
		{Personalized\\pricing policy};
	\node[saleoutcome, text width=2.85cm]
		(valuation) at (4.70,2.85)
		{Buyer valuation\\[-1pt]
		\scriptsize (willingness to pay)};

	\draw[salearrow, draw=green!42!black]
		(true-verified.east) --
		node[saleedge, fill=white, inner sep=1pt, above=1pt]
		{directly\\reveal}
		(observed-verified.west);

	\draw[salearrow, draw=orange!58!black]
		(true-reportable.east) --
		node[saleedge, fill=white, inner sep=1pt, above=1pt]
		{strategic report\\at a cost}
		(reported.west);

	\draw[salearrow]
		(true-profile.north east)
		to[out=32,in=180]
		node[saleedge, midway, above]
		{determines}
		(valuation.west);

	\draw[salearrow]
		(observed-profile.east) --
		node[saleedge, above=1pt]
		{observed}
		(seller.west);

	\draw[salearrow]
		(seller.east) --
		node[saleedge, above=1pt]
		{applies}
		(policy.west);

	\draw[draw=black!48, densely dashed, line width=.6pt]
		(valuation.east)
		to[out=5,in=120]
		node[saleedge, midway, above]
		{latent learning target}
		(policy.north);
\end{tikzpicture}

\caption{Information structure in source-aware personalized pricing. Buyer valuation depends on true features, whereas the seller bases pricing decisions on verified features and potentially manipulated reports.}
\label{fig:source-information}
\end{figure}

Existing studies of personalized pricing with strategic buyers typically use manipulation-cost models to characterize buyers' reporting behavior, without explicitly accounting for differences in information provenance \citep{LiLi2022,Liu2025,LiuSun2026Fairness}. However, directly applying these approaches to source-aware contextual pricing may not only incorrectly treat verified features as manipulable but also prevent the seller from using reliable buyer information for more refined personalized exploration. Among these studies on strategic pricing, \citet{Liu2025} and \citet{LiuSun2026Fairness} develop policies in low-dimensional feature settings,
and \citet{Liu2025} additionally assumes a known market-noise distribution.
Yet in practice, the number of potentially relevant buyer features can be large in the aforementioned  dynamic-pricing applications, and the distribution of unobserved market noise may not be known in advance \citep{FanGuoYu2024, WangChen2025, ChenChenGaoLi2026}. These considerations motivate further advances in contextual dynamic pricing with strategic buyers.

In this paper, we develop a source-aware framework for contextual dynamic pricing with strategic buyers, distinguishing verified from reportable features. Within this framework, we consider a high-dimensional semiparametric setting in which buyer valuations follow a sparse linear model and the market-noise distribution is unknown.
To address strategic manipulation while making full use of verified information, we develop the Source-Aware Localized exploration and calibrated Exploitation (SALE) pricing policy.
In the exploration phase, SALE uses current verified features and historical information to localize price exploration from the buyer-independent full price domain to adaptive, buyer-specific pricing intervals.
In the exploitation phase, it uses both verified and reported features to calibrate prices to correct for strategic manipulation.
Moreover, we derive a new population equation linking binary purchase feedback from localized exploration to the sparse valuation parameters. Based on this equation, we construct an \(\ell_1\)-regularized estimator and establish its consistency under conditions allowing buyer features to be dependent over time. We then use the estimated parameters to learn the unknown noise distribution nonparametrically from the same observations. To establish convergence guarantees in the high-dimensional setting, we uncover the intrinsic sparse structure of the resulting nonparametric estimation problem through a localized empirical-process analysis. These guarantees finally yield a sublinear expected regret bound for SALE.

\subsection{Main Contributions}

The main contributions of this paper are threefold. First, our work provides the first study of source-aware contextual dynamic pricing with strategic buyers, making a practically meaningful addition to the existing literature.  Moreover, we study the  strategic dynamic pricing problem in a high-dimensional semiparametric setting,  extending beyond the low-dimensional settings of \citet{Liu2025} and \citet{LiuSun2026Fairness} and relaxing the assumption of a known market-noise distribution in \citet{Liu2025}. We also emphasize that our framework allows buyer arrivals to be dependent over time, better reflecting markets with returning buyers, and is, to the best of our knowledge, the first to accommodate such dependence in contextual dynamic pricing with strategic feature reporting.

Second, the proposed SALE policy adopts a novel buyer-specific exploration design that uses verified information while keeping exploration prices unaffected by reported features. Compared with the feature-independent exploration commonly used in prior dynamic pricing studies \citep{FanGuoYu2024, Liu2025, ChenChenGaoLi2026}, SALE concentrates price exploration within narrower personalized intervals, which naturally reduces per-period exploration costs.
A further methodological innovation is an unbiased response construction for buyer-specific exploration. It turns binary feedback from different exploration intervals into a common population equation for high-dimensional sparse parameter estimation without requiring a known noise distribution. The estimated parameters then allow us to learn that distribution nonparametrically from residuals constructed on the same sample.

Third, we establish high-dimensional sparse estimation guarantees under non-i.i.d. buyer arrivals for an \(\ell_1\)-regularized least-squares estimator,  in contrast to the likelihood-based estimators studied by \citet{Javanmard2019} and \citet{WangWangSunCheng2024}.  For nonparametric noise learning, we develop a localized empirical-process analysis that exploits intrinsic sparse structure to establish uniform estimation guarantees, thereby extending the semiparametric analysis of \citet{FanGuoYu2024} to the high-dimensional setting.
With these estimation guarantees, we establish SALE’s self-recovery property, which keeps localized exploration valid with high probability. A distinctive feature of this result is that it guarantees recovery from failures of localized exploration caused by estimation errors, a difficulty absent from the feature-independent exploration adopted by \citet{FanGuoYu2024}, \citet{ChenChenGaoLi2026}, and \citet{Liu2025}, among others.
Building on these results,  we derive a sublinear expected regret bound for SALE,  with an explicit factor quantifying the benefit of using verified information for personalized pricing.

\subsection{Related Work}

\paragraph{Pricing with Strategic Buyers.}
Our work is most closely related to studies of personalized pricing in which buyers can strategically manipulate the information available to sellers.
In a static setting, \citet{LiLi2022} formulate a personalized-pricing game in which consumers may manipulate their data to obtain lower prices, while the firm chooses how much data to collect and whether to disclose its collection scope.
In a dynamic setting,
\citet{Liu2025} study a low-dimensional parametric contextual-pricing model with a known market-noise distribution, and \citet{LiuSun2026Fairness} consider strategic misreporting of a private binary group label under constraints on price disparities across groups.
In contrast, our framework for contextual dynamic pricing explicitly distinguishes verified from reportable features within a high-dimensional buyer profile and examines how this distinction affects pricing policy and valuation learning.

Related research on pricing with strategic buyers also considers the intertemporal incentives.
One stream examines forward-looking buyers who time purchases in anticipation of future prices \citep{Su2007,LevinMcGillNediak2008,ChenFarias2018,ChenFariasTrichakis2019}, with related studies incorporating social learning, inventory, or resource constraints \citep{PapanastasiouSavva2016,ChenShi2019,AbhishekDoganJacquillat2021}.
A second stream studies repeated posted-price or contextual-auction interactions in which buyers may distort purchase decisions, bids, or valuation reports to influence future offers \citep{AminRostamizadehSyed2014,Drutsa2020,GolrezaeiJavanmardMirrokni2020,ChenGaoGeWang2022,GolrezaeiJailletLiang2023,HuhKandasamy2026}.
These studies focus on buyers' purchase timing and their influence on future prices, whereas we examine how information provenance constrains strategic feature reporting and affects valuation learning and personalized pricing.

\paragraph{Pricing with High-Dimensional Features.}
High-dimensional parametric contextual pricing has attracted considerable attention in recent years, with representative studies exploiting sparsity and regularization to estimate valuation or demand parameters from high-dimensional covariates \citep{Javanmard2019,BanKeskin2021,WangWangSunCheng2024}.
Related work allows demand parameters to change at unknown times \citep{ZhaoJiangYuChen2026}, studies growing effective dimensionality \citep{BanFanHanJagabathulaKeskinZhou2026}, and characterizes how regret depends on the selling horizon and feature dimension in a linear setting \citep{ChaiDuanFanWang2024}. These studies address different aspects of learning and pricing with high-dimensional features while retaining a parametric model structure.

High-dimensional semiparametric contextual pricing methods typically retain a parametric valuation index while leaving the market-noise distribution unspecified.
\citet{FanGuoYu2024} use penalized variable selection followed by least-squares refitting on the selected support to reduce high-dimensional semiparametric pricing to a low-dimensional problem under conditions ensuring variable selection consistency.
\citet{GongYouZhang2025} develop a general framework for contextual pricing with an unknown noise distribution and obtain regret guarantees for sparse high-dimensional linear valuation models through a regression oracle.
More recently, \citet{ChaiDuanFanWang2026} exploit feature diversity to reuse exploitation data for nonparametric noise estimation, with regret guarantees covering sparse linear valuation models. \citet{FanHanLvXuZhou2026} learn the oracle pricing map directly and use Lasso estimation to extend their approach to sparse high-dimensional linear valuation models.
These studies do not account for strategic feature reporting, whereas we use verified information to support buyer-specific exploration and effective valuation learning with strategic buyers.

\paragraph{Related Topics in Pricing.}
Noncontextual dynamic pricing has been extensively studied, with representative work adopting Bayesian formulations \citep{FariasVanRoy2009,HarrisonKeskinZeevi2011} or frequentist learning methods \citep{BroderRusmevichientong2012,KeskinZeevi2014}.
With low-dimensional features, contextual dynamic pricing has been studied under parametric demand or valuation models \citep{QiangBayati2016,CohenLobelPaesLeme2020,ChenSimchiLeviWang2021} and semiparametric models with unknown noise distributions \citep{WangChen2025,HanXuWenHanLobelZhou2026}.
Further work addresses model misspecification through randomized price shocks \citep{NambiarSimchiLeviWang2019} and model uncertainty through randomization over price vectors in robust multiproduct pricing \citep{GuanMisic2026}.

Related topics also include research on revenue decisions.
For example, \citet{ElmachtoubGuptaHamilton2021} compare the profits from personalized pricing with those from charging a common price, and \citet{Golrezaei2014} develop personalized assortment policies for revenue maximization under inventory constraints.
Further work on learning consumer responses to guide revenue decisions ranges from nonparametric choice modeling with limited data \citep{Farias2013} to contextual demand estimation with nonparametric price confounders \citep{WangLiu2025}.
More broadly, decisions can affect both market behavior and the availability of information, with representative research studying customer goodwill and future budgets allocated to a platform in repeated interactions \citep{CalmonCiocanRomero2020}, the role of decisions in information discovery \citep{VayanosGeorghiouYu2025}, and experimental design and inference under equilibrium effects \citep{WagerXu2021}.

\subsection{Organization and notation}

The rest of the paper is organized as follows. Section~\ref{sec:Setting} introduces the problem setting. Section~\ref{sec:Algorithms} presents the SALE policies, and Section~\ref{sec:Regret} gives their regret guarantees.
Section~\ref{sec:Experiments} reports the numerical results. Section~\ref{sec:Discussion} concludes the paper and outlines several directions for future work.
The theoretical proofs, as well as additional experiments and implementation details for SALE, are provided in the Supplementary Material.

We use the following notation. For a positive integer \(n\), write \([n]=\{1,\ldots,n\}\). For a vector \(\mathbf v\), \(\|\mathbf v\|_q\) denotes its \(\ell_q\)-norm, \(\|\mathbf v\|_0\) its number of nonzero
coordinates, and \(\supp(\mathbf v)\) its support. For a symmetric matrix \(\mathbf M\), \(\lambda_{\min}(\mathbf M)\) and \(\lambda_{\max}(\mathbf M)\) denote its smallest and largest eigenvalues, respectively, while \(\|\mathbf M\|_{\mathrm{op}}\) and \(\|\mathbf M\|_{\max}\) denote its operator and entrywise maximum norms.
When \(\mathbf M\) is positive definite, write \(\|\mathbf v\|_{\mathbf M}=(\mathbf v^\top\mathbf M\mathbf v)^{1/2}\).
For a set \(A\), \(\operatorname{int}(A)\) denotes its interior. For a function \(f\) on an interval \(A\), \(\|f\|_{\infty,A}\) and \(\operatorname{Lip}_A(f)\) denote its
supremum norm and Lipschitz seminorm. For an event \(E\), \(\mathbb I(E)\) denotes its indicator. For two nonnegative scalar sequences \(\{a_n\}_{n\ge1}\) and \(\{b_n\}_{n\ge1}\), write \(a_n\lesssim b_n\) if
there is a constant \(C>0\), independent of \(n\), such that \(a_n\le Cb_n\) for all \(n\). Define \(a_n\gtrsim b_n\) analogously, and write \(a_n\asymp b_n\) when both comparisons hold. The notation
\(\widetilde O(\cdot)\) suppresses logarithmic factors. %

\section{Problem Setting}\label{sec:Setting}
As is standard in the dynamic pricing literature (e.g., \citealp{FanGuoYu2024, Liu2025}), we consider a seller that sells a single product over \(T\) periods, where \(T\) denotes the length of the horizon and may be unknown to the seller. Within this framework,  a distinctive feature of our setting is that buyer features differ in their verifiability.
At time \(t\), a buyer arrives with two types of true features: verified features \(\mathbf x_{t,\mathrm V}^0\in\mathcal B_{\mathrm V}\subseteq
\mathbb R^{d_{\mathrm V}}\) and reportable features \(\mathbf x_{t,\mathrm R}^0\in\mathcal B_{\mathrm R}\subseteq \mathbb R^{d_{\mathrm R}}\).  The seller observes the verified features directly, while the buyer may strategically manipulate the reportable features from their true values \(\mathbf x_{t,\mathrm R}^0\) to the reported values \(\mathbf x_{t,\mathrm R}\) to seek a lower price.
Let \(d=d_{\mathrm V}+d_{\mathrm R}\), and denote the buyer's full true feature vector, with an intercept as its last coordinate, by
\[
\mathbf x_t^0=
\big((\mathbf x_{t,\mathrm V}^0)^\top,
(\mathbf x_{t,\mathrm R}^0)^\top,1\big)^\top
\in\mathbb R^{d+1}.
\]
We allow for high-dimensional buyer features. To account for repeat buyers, we also allow non-i.i.d. buyer arrivals: buyer features may be temporally dependent and follow an exogenous data-generating process unknown to the seller.

To choose a price for the arriving buyer, the seller needs to assess the buyer’s willingness to pay, which is not directly observed.
Following the literature on high-dimensional contextual dynamic pricing (e.g., \citealp{Javanmard2019}),  we model the buyer's valuation $v_t$ as
\begin{equation}\label{eq:valuation-model}
v_t=\boldsymbol\theta_0^\top\mathbf x_t^0+z_t,
\end{equation}
where
\(\boldsymbol\theta_0=(\boldsymbol\beta_{\mathrm V,0}^\top, \boldsymbol\beta_{\mathrm R,0}^\top,\alpha_0)^\top\in\Theta\subseteq\mathbb R^{d+1}\) is the unknown sparse valuation parameter and \(\{z_t\}_{t\ge1}\) is a mean-zero market-noise sequence independent of the true-feature sequence.
The sparsity is formulated by \(s_0=\|\boldsymbol\theta_0\|_0\ll d\), meaning that only a small subset of the buyer's features affects valuation.
We make the same assumptions about the noise and valuations as \citet{FanGuoYu2024}.
The noise variables \(\{z_t\}_{t\ge1}\) are i.i.d. draws from an unknown distribution with cdf \(F\), supported on \((-\delta_z,\delta_z)\) for a known constant \(\delta_z>0\).
Buyer valuations satisfy \(v_t\in[\delta_v,B-\delta_v]\) for some \(\delta_v>0\), where \(B>0\) is a known upper bound.
Given this valuation bound, the seller posts a price \(p_t\in[0,B]\) according to a pricing policy \(\pi\) that uses past observations and the current buyer’s verified features and report. The seller then observes the binary sale outcome \(y_t=\mathbb I\{p_t\le v_t\}\).
If \(p_t\le v_t\), a sale occurs, \(y_t=1\), and the seller earns revenue \(p_t\); otherwise, no sale occurs, \(y_t=0\), and the revenue is zero.

To evaluate the seller's pricing policy  \(\pi\) , we compare it with a clairvoyant seller that knows \(\boldsymbol\theta_0\), \(F\), and the buyer's true features.  Under valuation
model \eqref{eq:valuation-model}, the probability of a sale at price \(p\in[0,B]\), conditional on the true features, is
\[
\mathbb P(v_t\ge p\mid\mathbf x_t^0)
=1-F(p-u_t),
\]
where \(u_t=\boldsymbol\theta_0^\top\mathbf x_t^0\) denotes the buyer's valuation excluding market noise.
The corresponding expected revenue is therefore
\begin{equation}\label{revenuet}
\operatorname{rev}_t(p)
=p\{1-F(p-u_t)\}.
\end{equation}
To express the oracle's optimal price as a function of the valuation component, define
\begin{equation}\label{eq:global-oracle-map}
g(u)\in\arg\max_{p\in[0,B]}p\{1-F(p-u)\}.
\end{equation}
If the maximum expected revenue is zero, set \(g(u)=0\).
The oracle therefore posts
\begin{equation*}
p_t^\ast=g(u_t)
=g(\boldsymbol\theta_0^\top\mathbf x_t^0).
\end{equation*}
Using this oracle as the benchmark, we measure the performance of \(\pi\) by its expected
cumulative regret relative to the oracle:
\begin{equation*}
\Reg_\pi(T)
=\mathbb E_\pi\!\left[
\sum_{t=1}^T
\left\{
p_t^\ast\mathbb I(p_t^\ast\le v_t)
-p_t\mathbb I(p_t\le v_t)
\right\}
\right].
\end{equation*}

\begin{remark}\label{zzjshi}%
For an optimal price \(g(u)\in(0,B)\) at which the expected
revenue function is differentiable, the first-order condition is
\[
1-F(g(u)-u)=g(u)F'(g(u)-u).
\]
Define
\[
\phi(z)=z-\frac{1-F(z)}{F'(z)}
\]
wherever \(F'(z)>0\). If \(g(u)-u\) is in this domain,  i.e., \(F'(g(u)-u)>0\), the first-order
condition can be rewritten as
\[
\phi(g(u)-u)=-u.
\]
When \(\phi\) is strictly increasing, this equation yields
\begin{equation}\label{eq:oracle-map-representation}
g(u)=u+\phi^{-1}(-u),
\end{equation}
where \(\phi^{-1}\) denotes the inverse of \(\phi\).
Regularity assumptions will be imposed in Section~\ref{sec:regularity-conditions} to ensure the validity of the above pricing characterization.
\end{remark}

\subsection{Source-Aware Feature Manipulation}\label{sec:manipulation-cost}
A buyer's report may alter the posted price, but it does not affect her valuation. After receiving the reported features, the seller observes
\(\mathbf x_t=\big((\mathbf x_{t,\mathrm V}^0)^\top, \mathbf x_{t,\mathrm R}^\top,1\big)^\top\) rather than the true features $\mathbf x_t^0=
\big((\mathbf x_{t,\mathrm V}^0)^\top, (\mathbf x_{t,\mathrm R}^0)^\top,1\big)^\top$.
Similar to \citet{ Liu2025},  we model the cost of manipulating the reportable features from
\(\mathbf x_{t,\mathrm R}^0\) to \(\mathbf x_{\mathrm R}\) as
\begin{equation*}
\mathrm{cost}(\mathbf x_{t, \mathrm R},\mathbf x_{t,\mathrm R}^0)
=\frac12(\mathbf x_{t, \mathrm R}-\mathbf x_{t,\mathrm R}^0)^\top
\mathbf A(\mathbf x_{t, \mathrm R}-\mathbf x_{t,\mathrm R}^0),
\end{equation*}
where the symmetric positive definite matrix \(\mathbf A\in\mathbb R^{d_{\mathrm R}\times d_{\mathrm R}}\) is common across buyers.
Unlike \citet{Liu2025}, our source-aware framework restricts the buyer's reporting decision and manipulation cost to reportable features rather than the full vector of buyer features, while holding verified features fixed.

The buyer’s reporting decision depends on both the manipulation cost and the seller’s pricing rule.
Since the valuation model is unknown to the seller,  this rule differs between exploration periods, when the seller collects data to learn the valuation model, and exploitation periods, when the seller uses the resulting estimates to set prices. More specifically, before the buyer reports, the seller announces whether the current period is used for exploration or exploitation and discloses the form of the corresponding pricing rule. During exploration, the pricing rule does not use the reported features \(\mathbf x_{t,\mathrm R}\). The buyer therefore has no incentive to manipulate and reports truthfully, with \(\mathbf x_{t,\mathrm R}=\mathbf x_{t,\mathrm R}^{0}\). During exploitation, the pricing rule uses the reported features \(\mathbf x_{t,\mathrm R}\). The buyer may therefore submit a manipulated report \(\mathbf x_{t,\mathrm R}\neq\mathbf x_{t,\mathrm R}^{0}\) to seek a lower price, incurring a manipulation cost. Figure 2 summarizes the seller’s pricing rules during exploration and exploitation.

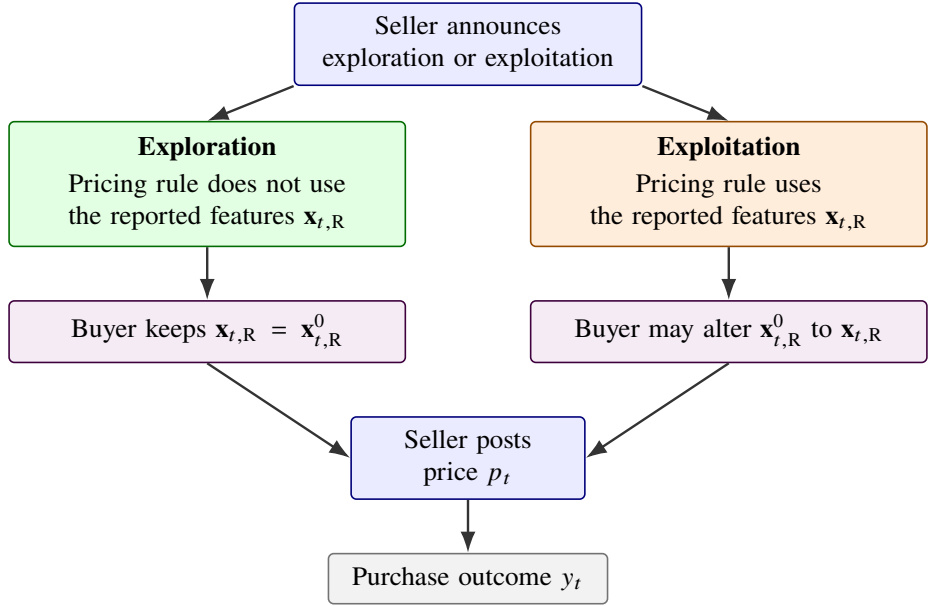
\begin{figure}[htbp]
\centering
\begin{tikzpicture}[x=1cm,y=1cm,
    reportarrow/.style={-{Latex[length=2.8mm,width=2mm]},
        draw=black!80, line width=.9pt}]
    \node[saleseller, text width=4.1cm] (announce) at (0,0)
        {Seller announces\\exploration or exploitation};

    \node[saleverified, text width=4.75cm, minimum height=1.65cm]
        (exploration) at (-3.45,-1.85)
        {\textbf{Exploration}\\[2pt]
        Pricing rule does not use\\the reported features $\mathbf x_{t,\mathrm R}$};

    \node[salenode, fill=violet!8, draw=violet!48!black,
        text width=4.75cm, minimum height=.8cm, below=7mm of exploration]
        (exploration-report)
        {Buyer keeps $\mathbf x_{t,\mathrm R}=\mathbf x_{t,\mathrm R}^0$};

    \node[salereportable, text width=4.75cm, minimum height=1.65cm]
        (exploitation) at (3.45,-1.85)
        {\textbf{Exploitation}\\[2pt]
        Pricing rule uses\\ the reported features $\mathbf x_{t,\mathrm R}$};

    \node[salenode, fill=violet!8, draw=violet!48!black,
        text width=4.75cm, minimum height=.8cm, below=7mm of exploitation]
        (exploitation-report)
        {Buyer may alter $\mathbf x_{t,\mathrm R}^0$ to $\mathbf x_{t,\mathrm R}$};

    \node[saleseller, text width=2.6cm,
        below=7mm of exploration-report, xshift=3.45cm] (post)
        {Seller posts\\price $p_t$};

    \node[saleoutcome, text width=3.2cm, below=7mm of post] (outcome)
        {Purchase outcome $y_t$};

    \draw[reportarrow] (announce.south west) -- (exploration.north);
    \draw[reportarrow] (announce.south east) -- (exploitation.north);
    \draw[reportarrow] (exploration.south) -- (exploration-report.north);
    \draw[reportarrow] (exploitation.south) -- (exploitation-report.north);
    \draw[reportarrow] (exploration-report.south) -- (post.west);
    \draw[reportarrow] (exploitation-report.south) -- (post.east);
    \draw[reportarrow] (post.south) -- (outcome.north);
\end{tikzpicture}
\caption{Pricing rules in exploration and exploitation.}
\label{fig:reporting-modes}
\end{figure}

To choose her optimal report during exploitation, the buyer needs to consider how her report would affect the price she expects to pay under the announced pricing rule. We assume that the buyer knows her valuation information $\boldsymbol{\theta}_0$ and $F$, and hence the pricing function $g$, whereas $\boldsymbol{\theta}_0$ and $F$ are unknown to the seller.
Although the seller discloses the form of the pricing rule, it does not reveal the estimates obtained from exploration data and used to set prices during exploitation.
Without access to the seller's estimates, the buyer uses the known valuation information to form her expected price.
Following \citet{Liu2025}, the buyer chooses her report to minimize her perceived price plus the manipulation cost:
\begin{equation}\label{eq:buyer-report-problem}
\mathbf x_{t,\mathrm R}\in
\arg\min_{\mathbf x_{\mathrm R}\in\mathbb R^{d_{\mathrm R}}}
\left\{
g\!\left(
\boldsymbol\beta_{\mathrm V,0}^{\top}\mathbf x_{t,\mathrm V}^{0}
+\boldsymbol\beta_{\mathrm R,0}^{\top}\mathbf x_{\mathrm R}
+\alpha_0
\right)
+\frac12
(\mathbf x_{\mathrm R}-\mathbf x_{t,\mathrm R}^{0})^{\top}
\mathbf A
(\mathbf x_{\mathrm R}-\mathbf x_{t,\mathrm R}^{0})
\right\}.
\end{equation}
The first-order condition yields
\begin{equation}\label{eq:manipulation}
\mathbf x_{t,\mathrm R}
=
\mathbf x_{t,\mathrm R}^{0}
-\mathbf A^{-1}\boldsymbol\beta_{\mathrm R,0}
g'(\boldsymbol\theta_0^{\top}\mathbf x_t),
\end{equation}
which also relates the true and reported valuation components:
\begin{equation}\label{eq:population-corrected-valuation}
\boldsymbol\theta_0^{\top}\mathbf x_t^{0}
=
\boldsymbol\theta_0^{\top}\mathbf x_t
+
\boldsymbol\beta_{\mathrm R,0}^{\top}
\mathbf A^{-1}\boldsymbol\beta_{\mathrm R,0}
g'(\boldsymbol\theta_0^{\top}\mathbf x_t).
\end{equation}

The seller, in turn, needs to account for this reporting behavior when setting prices.
\citet{Liu2025} show that pricing directly from manipulated features can incur linear regret in their setting.
In our model, even if the seller knew $\boldsymbol\theta_0$ and $F$, pricing at $g(\boldsymbol\theta_0^\top\mathbf x_t)$ need not yield the oracle price $g(\boldsymbol\theta_0^\top\mathbf x_t^0)$ since the manipulated report \(\mathbf x_{t,\mathrm R}\neq\mathbf x_{t,\mathrm R}^{0}\).
The above equality \eqref{eq:population-corrected-valuation} shows how the seller can correct the valuation component computed from the reported features before applying $g$.
Since $\boldsymbol\theta_0$ and $F$ are unknown to the seller, implementing
this correction requires estimates of the valuation parameter $\boldsymbol\theta_0$, the pricing function $g$, and its derivative.
The policy developed in Section~\ref{sec:Algorithms} learns these quantities from truthful exploration data and uses them to calibrate exploitation prices.

\section{SALE Policy}\label{sec:Algorithms}
We develop the Source-Aware Localized exploration and calibrated Exploitation (SALE) policy. We first present the SALE-K algorithm for a manipulation cost matrix known to the seller.
Appendix~\ref{sec:sale-unknown-a} further extends SALE-K to SALE-U when the manipulation cost matrix is unknown to the seller.

To balance exploration and exploitation, SALE-K, presented in Algorithm~\ref{alg:knownA}, adopts the episodic structure widely used in prior dynamic pricing studies \citep{FanGuoYu2024,Liu2025,ChenChenGaoLi2026}.
Specifically, SALE-K operates in episodes of increasing length, with episode $k$ having length $\ell_k=2^{k-1}\ell_0$, where $\ell_0$ is the initial episode length.
Episode $k$ starts at time $S_k=1+\sum_{j=1}^{k-1}\ell_j$ and consists of an exploration stage $I_k$ of length $a_k$, followed by an exploitation stage $I_k'$ of length $\ell_k-a_k$.
During $I_k$,  SALE-K uses two functions, $L_k(\cdot)$ and $U_k(\cdot)$, to map each buyer's verified features to the lower and upper endpoints of her personalized exploration interval.
These functions are initialized as  $L_1(\cdot)\equiv0$ and $U_1(\cdot)\equiv B$, so the first episode uses $[0,B]$ for all buyers.
For each arriving buyer, SALE-K draws the exploration price uniformly from the corresponding personalized interval.
Using the observations in $I_k$, SALE-K computes the parameter estimate $\widehat{\boldsymbol\theta}_k$, the estimated pricing function $\widehat g_k$, and its derivative $\widehat g_k'$.
It also uses $\widehat{\boldsymbol\theta}_k$ to construct the interval functions $L_{k+1}(\cdot)$ and $U_{k+1}(\cdot)$ for personalized exploration in episode $k+1$.
During $I_k'$, SALE-K uses $\mathbf A$, $\widehat g_k'$, and $\widehat{\boldsymbol\theta}_k$ to correct the estimated valuation component for strategic
manipulation based on the relationship in  \eqref{eq:population-corrected-valuation}.
SALE-K then applies $\widehat g_k$ to the corrected valuation component and truncates the resulting price to $[0,B]$.

\par\medskip
\begin{breakablealgorithm}
\caption{SALE-K (Known $\mathbf A$)}
\label{alg:knownA}
\begin{algorithmic}[1]
    \STATE \textbf{Input:} Price bound $B$; initial episode length $\ell_0$; manipulation-cost matrix $\mathbf A$.
    \STATE Initialize the personalized exploration interval $[L_1(\cdot), U_1(\cdot)]$, where constant functions $L_1(\cdot)\equiv0$ and $U_1(\cdot)\equiv B$.

    \FOR{each episode $k=1,2,\ldots$}
        \STATE Set $\ell_k=2^{k-1}\ell_0$ and
        $S_k=1+\sum_{j<k}\ell_j$.

        \STATE Set the exploration duration $a_k$ according to \eqref{eq:width-adaptive-schedule}.

        \STATE Set $I_k=\{S_k,\ldots,S_k+a_k-1\}$ and
        $I_k'=\{S_k+a_k,\ldots,S_k+\ell_k-1\}$.

        \STATE \textbf{Exploration phase:}
        \FOR{$t\in I_k$}
            \STATE Observe $\mathbf x_{t,\mathrm V}^0$ and the truthful
            report $\mathbf x_{t,\mathrm R}=\mathbf x_{t,\mathrm R}^0$.

            \STATE Draw
            $p_t\sim\mathrm{Unif}\!\left(
            L_k(\mathbf x_{t,\mathrm V}^0),
            U_k(\mathbf x_{t,\mathrm V}^0)\right)$
            and observe $y_t$.

        \ENDFOR

        \STATE Using observations from
        $I_k$,   compute $\widehat{\boldsymbol\theta}_k$ by
        \eqref{eq:least_squares}.

          \STATE Set the personalized exploration interval $[L_{k+1}(\cdot),U_{k+1}(\cdot)]$ for episode $k+1$, where the functions $L_{k+1}(\cdot)$ and $U_{k+1}(\cdot)$ are constructed using $\widehat{\boldsymbol\theta}_k$
             according to \eqref{eq:safe-exploration-interval}.

  \STATE Using observations from $I_k$, compute $\widehat g_k$ and $\widehat g_k'$
        by \eqref{eq:ghat}.

        \STATE \textbf{Exploitation phase:}
        \FOR{$t\in I_k'$}
            \STATE Observe $\mathbf x_{t,\mathrm V}^0$ and
            $\mathbf x_{t,\mathrm R}$, and post
            \begin{equation*}
                p_t=\min\!\left\{B,\max\!\left\{0,
                \widehat g_k\!\left(
                \widehat{\boldsymbol\theta}_k^\top\mathbf x_t
                +\widehat{\boldsymbol\beta}_{\mathrm R,k}^\top
                \mathbf A^{-1}
                \widehat{\boldsymbol\beta}_{\mathrm R,k}
                \widehat g_k'\!\left(
                \widehat{\boldsymbol\theta}_k^\top\mathbf x_t
                \right)
                \right)\right\}\right\}.
            \end{equation*}

            \STATE Observe $y_t$.
        \ENDFOR
    \ENDFOR
\end{algorithmic}
\end{breakablealgorithm}

\medskip

\begin{figure}[!ht]
\centering
\resizebox{\linewidth}{!}{%
\begin{tikzpicture}[
    box/.style={
        draw, rounded corners=2pt, align=center,
        font=\footnotesize, text width=1.65cm,
        minimum height=1.4cm, inner sep=3pt
    },
    interval/.style={
        box, text width=2.4cm,
        draw=green!50!black, fill=green!8
    },
    exploration/.style={box, draw=blue!65!black, fill=blue!5},
    exploitation/.style={box, draw=orange!75!black, fill=orange!10},
    estimate/.style={box, draw=black!50, fill=black!5},
    flow/.style={-{Stealth[length=1.8mm]}, thick},
    update/.style={
        flow, densely dashed, draw=violet!75!black,
        rounded corners=3pt
    },
    episode/.style={
        draw=black!20, fill=black!2, rounded corners=4pt,
        inner sep=6pt
    },
    period/.style={
        draw=black!65, semithick, decorate,
        decoration={brace,mirror,amplitude=4pt}
    },
    note/.style={font=\footnotesize, align=center},
    phase/.style={font=\footnotesize, below=3pt, inner sep=1pt}
]%
\node[interval, text width=1.5cm] (initial) at (-2.5,1.2) {
    $L_1(\cdot)\equiv0$\\[1pt]$U_1(\cdot)=B$
};

\node[interval] (intervalPrev) at (0.6,3.35) {
    Personalized\\exploration intervals\\[2pt]
    $[L_{k}(\cdot),U_{k}(\cdot)]$
};
\node[exploration] (explorePrev) at (0.6,1.2) {
    Uniform\\exploration\\[2pt]in $I_{k}$
};
\node[estimate] (estimatePrev) at (2.75,1.2) {
    Estimates\\[2pt]
    $\widehat{\boldsymbol\theta}_{k}$\\[1pt]
    $\widehat g_{k},\,\widehat g'_{k}$
};
\node[exploitation] (exploitPrev) at (4.9,1.2) {
    Calibrated\\exploitation\\[2pt]in $I_{k}'$
};

\node[interval] (intervalNow) at (7.9,3.35) {
    Personalized\\exploration intervals\\[2pt]
    $[L_{k+1}(\cdot),U_{k+1}(\cdot)]$
};
\node[exploration] (exploreNow) at (7.9,1.2) {
    Uniform\\exploration\\[2pt]in $I_{k+1}$
};
\node[estimate] (estimateNow) at (10.05,1.2) {
    Estimates\\[2pt]
    $\widehat{\boldsymbol\theta}_{k+1}$\\[1pt]
    $\widehat g_{k+1},\,\widehat g_{k+1}'$
};
\node[exploitation] (exploitNow) at (12.2,1.2) {
    Calibrated\\exploitation\\[2pt]in $I_{k+1}'$
};

\begin{scope}[on background layer]
    \node[episode,
        fit=(intervalPrev)(explorePrev)(estimatePrev)(exploitPrev)]
        (episodePrev) {};
    \node[episode,
        fit=(intervalNow)(exploreNow)(estimateNow)(exploitNow)]
        (episodeNow) {};
\end{scope}
\draw[flow] (intervalPrev.south) -- (explorePrev.north);
\draw[flow] (explorePrev.east) -- (estimatePrev.west);
\draw[flow] (estimatePrev.east) -- (exploitPrev.west);
\draw[flow] (intervalNow.south) -- (exploreNow.north);
\draw[flow] (exploreNow.east) -- (estimateNow.west);
\draw[flow] (estimateNow.east) -- (exploitNow.west);

\draw[update]
    (estimatePrev.north) -- (2.75,3.35)
    -- node[note, above=2pt, xshift=-2.5mm] {
        Construct $L_{k+1}(\cdot)$ \\ and $U_{k+1}(\cdot)$
        using $\widehat{\boldsymbol\theta}_{k}$
    } (intervalNow.west);
\draw[update]
    (estimateNow.north) -- (10.05,3.35)
    -- node[note, above=2pt, xshift=-11mm] {
       Construct $L_{k+2}(\cdot)$ \\ and $U_{k+2}(\cdot)$
        using $\widehat{\boldsymbol\theta}_{k+1}$
    } (14.5,3.35);
\node[font=\large] at (14.85,3.35) {$\cdots$};

\draw[flow, draw=black!60] (-3.35,0) -- (15.1,0);
\node[fill=white, inner sep=2pt] at (-1.35,0) {$\cdots$};
\node[fill=white, inner sep=2pt] at (14.25,0) {$\cdots$};

\coordinate (initialStart) at (initial.west |- 0,0);
\coordinate (initialEnd) at (initial.east |- 0,0);
\coordinate (prevStart) at (episodePrev.west |- 0,0);
\coordinate (prevEnd) at (episodePrev.east |- 0,0);
\coordinate (nowStart) at (episodeNow.west |- 0,0);
\coordinate (nowEnd) at (episodeNow.east |- 0,0);
\draw[green!50!black, line width=1.2pt] (initialStart) -- (initialEnd);
\draw[blue!65!black, line width=1.2pt]
    (prevStart) -- node[midway, phase, text=blue!65!black] {Exploration}
    (estimatePrev.center |- 0,0);
\draw[orange!75!black, line width=1.2pt]
    (estimatePrev.center |- 0,0)
    -- node[midway, phase, text=orange!75!black] {Exploitation} (prevEnd);
\draw[blue!65!black, line width=1.2pt]
    (nowStart) -- node[midway, phase, text=blue!65!black] {Exploration}
    (estimateNow.center |- 0,0);
\draw[orange!75!black, line width=1.2pt]
    (estimateNow.center |- 0,0)
    -- node[midway, phase, text=orange!75!black] {Exploitation} (nowEnd);
\draw[period] ([yshift=-7mm]initialStart)
    -- node[midway, below=6pt, font=\small] {Initialization}
    ([yshift=-7mm]initialEnd);
\draw[period] ([yshift=-7mm]prevStart)
    -- node[midway, below=6pt, font=\small] {Episode $k$}
    ([yshift=-7mm]prevEnd);
\draw[period] ([yshift=-7mm]nowStart)
    -- node[midway, below=6pt, font=\small] {Episode $k+1$}
    ([yshift=-7mm]nowEnd);

\end{tikzpicture}%
}
\caption{Personalized exploration and calibrated Exploitation across episodes.}
\label{fig:sale-learning-cycle}
\end{figure}
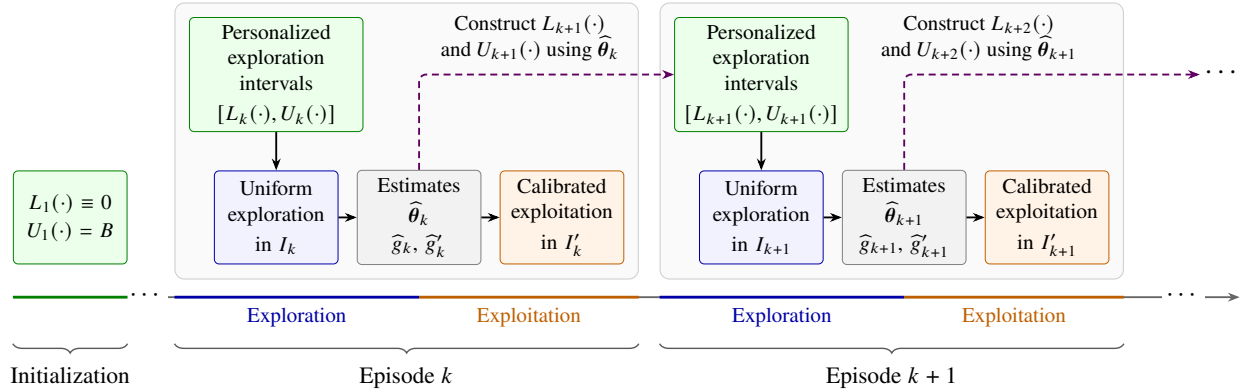

Related dynamic pricing studies  typically  draw exploration prices uniformly from a common fixed interval \citep{FanGuoYu2024,Liu2025,ChenChenGaoLi2026}.
In contrast, SALE-K draws exploration prices uniformly from personalized intervals determined by each buyer's verified features and updated parameter estimates.
This design uses verified information to adapt exploration to buyer heterogeneity while preserving truthful reporting.
Figure~\ref{fig:sale-learning-cycle} visualizes the updates to personalized exploration intervals across episodes and their connection to the exploration and exploitation phases.
The following sections describe these components in detail.
Section~\ref{sec:sale-localized-exploration}  explains the exploration duration, sparse valuation parameter
estimation, and updates to personalized exploration intervals.
Section~\ref{sec:sale-learning} presents the estimation of the pricing function and its
derivative, followed by the calibrated exploitation rule.

\smallskip

\subsection{Personalized Exploration and Valuation Parameter Estimation}
\label{sec:sale-localized-exploration}

\paragraph{Determination of the exploration duration $a_k$.}
SALE-K determines the exploration duration $a_k$ before episode $k$ begins.
A longer exploration stage provides more observations for estimating $\boldsymbol\theta_0$, $g$, and $g'$, but requires randomized pricing over more periods.
A shorter stage leaves more periods for exploitation, with fewer observations available for estimation.
The choice of $a_k$ therefore balances revenue losses during exploration and exploitation.

Personalized exploration intervals affect this balance through their widths.
Before episode $k$ begins, the functions $L_k(\cdot)$ and $U_k(\cdot)$ are available, but the verified features of the arriving buyers have not yet been observed.
SALE-K therefore considers the interval width associated with every feasible verified profile and takes their supremum.
It summarizes this width relative to the length $B$ of the full price range $[0,B]$ by
\[
\omega_k
=
\max\left\{
\omega_{\min},
\frac{1}{B}
\sup_{\mathbf x_{\mathrm V}\in\mathcal B_{\mathrm V}}
\big\{
U_k(\mathbf x_{\mathrm V})-L_k(\mathbf x_{\mathrm V})
\big\}
\right\},
\qquad
\omega_{\min}=\min\{1,2\delta_z/B\},
\]
where $\omega_{\min}$  reflects the noise range of width $2\delta_z$ that the interval construction must accommodate, and $\mathcal B_{\mathrm V}$ is the feasible set of verified features.

The construction of the functions $L_k(\cdot)$ and $U_k(\cdot)$, described below, ensures that $\omega_k\le1$,  which entails that all personalized exploration intervals have lengths no greater than that of the full price range $[0,B]$.
Our regret analysis further shows a smaller $\omega_k$ lowers the bound on the expected revenue loss in each exploration period relative to charging the optimal price, allowing SALE-K to allocate more periods to learning.
Accounting for the revenue losses during both exploration and exploitation,  the analysis yields the following choice of $a_k$:
\begin{equation}
a_k
=
\min\left\{
\ell_k,\,
\left\lceil
c_a(\ell_k\Lambda_{\ell_k})^{\frac{2m+3}{4m+1}}
\omega_k^{-\frac{2m+8}{4m+1}}
\right\rceil
\right\},
\label{eq:width-adaptive-schedule}
\end{equation}
where $c_a>0$ is a fixed constant, $\Lambda_{\ell_k}=1+\log\ell_k+s_0\log(d+1)$, $m$ is the smoothness order of $F$, and the outer minimum ensures that $a_k$ does not exceed $\ell_k$.
As we can see, for fixed $\ell_k$ and $\Lambda_{\ell_k}$, the exploration duration $a_k$ increases or remains unchanged as $\omega_k$ decreases.

\paragraph{Sparse valuation parameter estimation.}
For  \(t \in I_k\), exploration provides observations of the true features
$\mathbf x_t^0=\big((\mathbf x_{t,\mathrm V}^0)^\top,
(\mathbf x_{t,\mathrm R}^0)^\top,1\big)^\top$, prices $p_t$, and binary purchase outcomes $y_t$, while valuations $v_t$ remain unobserved.
Recall that under uniform exploration over $[0,B]$, the response $By_t$ has conditional mean $\boldsymbol\theta_0^\top\mathbf x_t^0$, as in \citet{FanGuoYu2024}.
To retain this target under our personalized exploration, SALE-K adjusts the binary outcome for the interval's location and width:
\[
Y_t
=
L_k(\mathbf x_{t,\mathrm V}^0)
+
W_k(\mathbf x_{t,\mathrm V}^0)y_t,
\]
where
$W_k(\cdot)
=U_k(\cdot)-L_k(\cdot)$  is the function that maps each buyer's verified features to the width of her personalized exploration interval.

When $v_t\in \left[L_k(\mathbf x_{t,\mathrm V}^0),U_k(\mathbf x_{t,\mathrm V}^0)\right]$,  uniform randomization gives
\[
\mathbb P(y_t=1\mid
\mathbf x_t^0,z_t,\mathcal F_{S_k-1})
=
\frac{v_t-L_k(\mathbf x_{t,\mathrm V}^0)}
{W_k(\mathbf x_{t,\mathrm V}^0)},
\]
where $\mathcal F_{S_k-1}$ is the $\sigma$-field generated by the seller's observations and policy randomization through period $S_k-1$.
Using the definition of $Y_t$,  it follows that
\[
\begin{aligned}
\mathbb E[Y_t\mid\mathbf x_t^0,\mathcal F_{S_k-1}]
=
\mathbb E\!\left[
\mathbb E[Y_t\mid\mathbf x_t^0,z_t,\mathcal F_{S_k-1}]
\,\middle|\,
\mathbf x_t^0,\mathcal F_{S_k-1}
\right]
=\boldsymbol\theta_0^\top\mathbf x_t^0.
\end{aligned}
\]
Motivated by this population equation, SALE-K estimates the sparse
valuation parameter using $\ell_1$-regularized least squares:
\begin{equation}
\widehat{\boldsymbol\theta}_k
=
\arg\min_{\boldsymbol\theta\in\Theta}
\left\{
\frac{1}{a_k}\sum_{t\in I_k}
\bigl(Y_t-\boldsymbol\theta^\top\mathbf x_t^0\bigr)^2
+
\lambda_k\|\boldsymbol\theta\|_1
\right\},
\label{eq:least_squares}
\end{equation}
where $\lambda_k=c_\lambda\{(1+\log\ell_k+\log(d+1))/a_k\}^{1/2}$ is the regularization parameter for a constant $c_\lambda>0$.

\begin{remark}
Let $\mathcal G_k$ denote the event that, conditional on
the pre-episode history $\mathcal H_{k-1}:=\mathcal F_{S_k-1}$,
\[
v_t\in
\left[
L_k(\mathbf x_{t,\mathrm V}^0),
U_k(\mathbf x_{t,\mathrm V}^0)
\right]
\quad\text{for all }t\in I_k
\]
holds almost surely.
On \(\mathcal G_k\), uniform price randomization also gives, for every \(t\in I_k\),
\begin{equation*}
\operatorname{Var}(Y_t\mid\mathbf x_t^0,z_t,\mathcal H_{k-1})
=\{v_t-L_k(\mathbf x_{t,\mathrm V}^0)\}
\{U_k(\mathbf x_{t,\mathrm V}^0)-v_t\}
\le \min\left\{\frac{W_k(\mathbf x_{t,\mathrm V}^0)^2}{4},
v_t(B-v_t)\right\}.
\end{equation*}
For comparison, let $y_t^{\mathrm{full}}$ denote the purchase
outcome for the same buyer under uniform exploration over
the full price range $[0,B]$.
Its conditional purchase probability is $v_t/B$, so
\[
\operatorname{Var}
(B y_t^{\mathrm{full}}\mid
\mathbf x_t^0,z_t,\mathcal H_{k-1})
=
B^2\frac{v_t}{B}\left(1-\frac{v_t}{B}\right)
=
v_t(B-v_t).
\]
Consequently, on $\mathcal G_k$, the adjusted response $Y_t$ has conditional variance no greater than that of the canonical response $B y_t^{\mathrm{full}}$.
\end{remark}

\paragraph{Updating personalized exploration intervals.}
SALE-K uses the valuation parameter estimate
$\widehat{\boldsymbol\theta}_k$ obtained from $I_k$ to construct
the next episode's interval functions $L_{k+1}(\cdot)$ and $U_{k+1}(\cdot)$,
which map each buyer's verified features to the endpoints of her
personalized exploration interval. Write
\[
\widehat{\boldsymbol\theta}_k
=
\big(
\widehat{\boldsymbol\beta}_{\mathrm V,k}^{\top},
\widehat{\boldsymbol\beta}_{\mathrm R,k}^{\top},
\widehat\alpha_k
\big)^{\top}.
\]
For each verified profile $\mathbf x_{t,\mathrm V}^0$ with \(t\in I_{k+1}\), the construction considers all feasible true reportable features
$\mathbf q\in\mathcal B_{\mathrm R}$, with the corresponding fitted
mean valuation given by
\[
\widehat{\boldsymbol\beta}_{\mathrm V,k}^{\top}\mathbf x_{t,\mathrm V}^0
+
\widehat{\boldsymbol\beta}_{\mathrm R,k}^{\top}\mathbf q
+
\widehat\alpha_k.
\]
Taking the minimum and maximum of the fitted mean valuation over $\mathcal B_{\mathrm R}$ gives a range for each verified profile.

To cover the buyer's actual valuation, SALE-K must also account for parameter estimation error and market noise.
Under the parameter error bound derived later,
$\|\widehat{\boldsymbol\theta}_k-\boldsymbol\theta_0\|_1
\le4C_\theta s_0\lambda_k$ for a constant $C_\theta>0$,
the fitted mean valuation differs from the true mean valuation
by at most $4C_\theta W_xs_0\lambda_k$,
where $W_x$ bounds the absolute value of each true feature coordinate.
Since the market noise is bounded by $\delta_z$,  SALE-K expands the fitted mean valuation range at both ends by
\[
\varepsilon_k^{\mathrm{loc}}
=
4C_\theta W_xs_0\lambda_k+\delta_z.
\]
Restricting  the expanded valuation range to $[0,B]$ gives the following endpoints:
\begin{equation}
\begin{aligned}
L_{k+1}(\mathbf x_{t,\mathrm V}^0)
&=
\max\left\{0,\,
\widehat{\boldsymbol\beta}_{\mathrm V,k}^{\top}\mathbf x_{t,\mathrm V}^0
+\widehat\alpha_k
+\min_{\mathbf q\in\mathcal B_{\mathrm R}}
\widehat{\boldsymbol\beta}_{\mathrm R,k}^{\top}\mathbf q
-\varepsilon_k^{\mathrm{loc}}
\right\},\\
U_{k+1}(\mathbf x_{t,\mathrm V}^0)
&=
\min\left\{B,\,
\widehat{\boldsymbol\beta}_{\mathrm V,k}^{\top}\mathbf x_{t,\mathrm V}^0
+\widehat\alpha_k
+\max_{\mathbf q\in\mathcal B_{\mathrm R}}
\widehat{\boldsymbol\beta}_{\mathrm R,k}^{\top}\mathbf q
+\varepsilon_k^{\mathrm{loc}}
\right\}.
\end{aligned}
\label{eq:safe-exploration-interval}
\end{equation}
If $L_{k+1}(\mathbf x_{t,\mathrm V}^0)>U_{k+1}(\mathbf x_{t,\mathrm V}^0)$, the endpoints are reset to $0$ and $B$, respectively.

\subsection{Pricing Function Estimation and Calibrated Exploitation}
\label{sec:sale-learning}

\paragraph{Estimation of the pricing function and its derivative.}
To obtain the pricing map and its derivative, SALE-K first estimates \(F\) using the exploration observations of the true features $\mathbf x_t^0$, prices $p_t$, and binary purchase outcomes $y_t$ for $t\in I_k$.
For a candidate parameter $\boldsymbol\theta\in\Theta$,  write the induced residual  $w_t(\boldsymbol\theta)=p_t-\boldsymbol\theta^\top\mathbf x_t^0$.
At the true parameter $\boldsymbol\theta_0$, \(y_t=\mathbb I\{z_t\ge w_t(\boldsymbol\theta_0)\}\), and hence
\[
\mathbb E[y_t\mid w_t(\boldsymbol\theta_0)=u]=1-F(u).
\]
Based on this population equation, SALE-K uses Nadaraya-Watson kernel regression to estimate  \(F\), as in \citet{FanGuoYu2024} and \citet{ChenChenGaoLi2026}.
Specifically, for any $\boldsymbol\theta\in\Theta$, define
\begin{equation*}
\widehat F_k(u,\boldsymbol\theta)
=
1-
\frac{
\sum_{t\in I_k}
K\!\left(
\frac{w_t(\boldsymbol\theta)-u}{b_k}
\right)y_t
}{
\sum_{t\in I_k}
K\!\left(
\frac{w_t(\boldsymbol\theta)-u}{b_k}
\right)
},
\end{equation*}
whenever the denominator is nonzero.   Here $K(\cdot)$ is the kernel and $b_k=(\omega_k/a_k)^{1/(2m+3)}$ is the bandwidth.
When $\boldsymbol\theta=\boldsymbol\theta_0$, $\widehat F_k(u,\boldsymbol\theta_0)$ gives the estimate of $F(u)$.

Since $\boldsymbol\theta_0$ is unknown, SALE-K substitutes the estimate $\widehat{\boldsymbol\theta}_k$ obtained from $I_k$ and estimates \(F\) using
\begin{equation*}
\widehat F_k(u)
=
\widehat F_k(u,\widehat{\boldsymbol\theta}_k).
\end{equation*}
The estimate $\widehat F_k$ is then used to construct the pricing function $\widehat g_k$ and its derivative $\widehat g_k'$ for calibrated exploitation.
Let $\widehat F_k^{(j)}$ denote the $j$th derivative of $\widehat F_k$, for $j=1,2$.
As in \citet{FanGuoYu2024}, substituting $\widehat F_k$ and
$\widehat F_k^{(1)}$ into \eqref{eq:oracle-map-representation}
gives the pricing function estimate.
SALE-K additionally computes $\widehat g_k'$ for strategic correction.
This requires $\widehat F_k^{(2)}$ through $\widehat\phi_k'$,
so SALE-K estimates $F''$ in addition to $F$ and $F'$.
Specifically,
\begin{equation}
\widehat\phi_k(u)
=
u-\frac{1-\widehat F_k(u)}{\widehat F_k^{(1)}(u)},
\qquad
\widehat g_k(u)
=
u+\widehat\phi_k^{-1}(-u),
\qquad
\widehat g_k'(u)
=
\frac{d}{du}\widehat g_k(u).
\label{eq:ghat}
\end{equation}

The numerical procedure for evaluating $\widehat g_k$ and $\widehat g_k'$ on
$I_g:=[\delta_z,B-\delta_z]$, including fallback pricing, is given in
Appendix~\ref{app:pricing-map-evaluation}.

\paragraph{Calibrated exploitation.}
For each $t\in I_k'$, the seller observes the buyer's verified and reported features, collected in $\mathbf x_t$.
Recall from \eqref{eq:population-corrected-valuation} that
\[
\boldsymbol\theta_0^\top\mathbf x_t^0
=
\boldsymbol\theta_0^\top\mathbf x_t
+
\boldsymbol\beta_{\mathrm R,0}^{\top}
\mathbf A^{-1}\boldsymbol\beta_{\mathrm R,0}
g'\bigl(\boldsymbol\theta_0^\top\mathbf x_t\bigr).
\]
The right-hand side expresses the true valuation component in terms of the observed features and a correction for strategic reporting.
SALE-K estimates this component by replacing $\boldsymbol\theta_0$, $\boldsymbol\beta_{\mathrm R,0}$, and $g'$ on the right-hand side with $\widehat{\boldsymbol\theta}_k$, $\widehat{\boldsymbol\beta}_{\mathrm R,k}$, and $\widehat g_k'$, respectively, using the estimates obtained from $I_k$.
Since the optimal pricing function $g$ is unknown to the seller, SALE-K uses its estimate $\widehat g_k$ to compute a price from the corrected valuation component and restricts the resulting
price to $[0,B]$:
\[
p_t
=
\min\left\{B,\,
\max\left\{0,\,
\widehat g_k\left(
\widehat{\boldsymbol\theta}_k^\top\mathbf x_t
+
\widehat{\boldsymbol\beta}_{\mathrm R,k}^{\top}
\mathbf A^{-1}
\widehat{\boldsymbol\beta}_{\mathrm R,k}
\widehat g_k'
\bigl(\widehat{\boldsymbol\theta}_k^\top\mathbf x_t\bigr)
\right)
\right\}
\right\}.
\]

\section{Regret Analysis}\label{sec:Regret}
We first impose some regularity conditions on the model settings.
Under these conditions, we establish estimation guarantees for the sparse valuation parameter and derive uniform error bounds for the estimated noise distribution, pricing function, and its derivative.
We then analyze the policy's self-recovery property and combine these results to establish the regret bound for SALE-K.

\subsection{Regularity Conditions}\label{sec:regularity-conditions}

\begin{assumption}\label{ass:bounded-sparse}
Both \(\mathcal B_{\mathrm V}\) and \(\mathcal B_{\mathrm R}\) are nonempty compact sets contained in the corresponding \(\ell_\infty\)-ball of radius \(W_x\ge1\).
Moreover,   \(\Theta\) is the corresponding $\ell_1$-ball of radius $W_\theta$.
\end{assumption}

As in \citet{Javanmard2019},  Assumption~\ref{ass:bounded-sparse} bounds each feature coordinate and the $\ell_1$-norm of the valuation parameter, so H\"older's inequality gives $|\boldsymbol\theta_0^\top\mathbf x_t^0|\le W_\theta W_x$ uniformly over the feasible feature set.
Similar boundedness assumptions also appear in the dynamic pricing literature; see, for example, \citet{FanGuoYu2024,Liu2025}.

\begin{assumption}\label{ass:covariance}
The conditional Gram matrix \(\boldsymbol\Sigma_t:=\mathbb E[\mathbf x_t^0\mathbf x_t^{0\top}\mid\mathcal F_{t-1}]\) is stable: for fixed constants \(0<\lambda_{\min}\le\lambda_{\max}<\infty\) and \(\eta_\Sigma\ge0\), almost surely,
\begin{equation}
\lambda_{\min}\mathbf I\preceq\boldsymbol\Sigma_t\preceq\lambda_{\max}\mathbf I
\quad\text{and}\quad
\|\boldsymbol\Sigma_t-\boldsymbol\Sigma_s\|_{\mathrm{op}}\le\eta_\Sigma\lambda_{\min}
\qquad\text{for all }s,t\ge1,
\label{eq:conditional-second-moment}
\end{equation}
where \(\eta_\Sigma<\rho_z^{-2}-1\) for the constant \(\rho_z:=\max\{1/2,1-\delta_z/(2W_xW_\theta)\}\).
\end{assumption}

Assumption~\ref{ass:covariance} retains the standard bounded eigenvalue condition used in the i.i.d. feature settings of \citet{Javanmard2019,FanGuoYu2024,Liu2025} to ensure that
the features provide sufficient information for learning, while allowing temporally dependent buyer features whose conditional second moment remains stable.
Intuitively, the buyer population may vary over time, but the overall scale of buyer features and their relationships cannot change too sharply.
In this sense,  this stability condition allows data collected in one episode to remain informative for pricing in the next.

\begin{assumption}\label{ass:noise-regularity}
For some integer \(m\ge3\), \(F\) is \(m\) times continuously differentiable on the interior of its support.
Moreover, \(F^{(m)}\) is Lipschitz on this interior, and the density \(F'\) is bounded away from zero on every compact subinterval of the interior.
\end{assumption}

Similar to \citet{FanGuoYu2024}, Assumption~\ref{ass:noise-regularity} provides the smoothness needed for uniform kernel estimation of $F$ and its derivatives.
We take $m\ge3$ to support uniform estimation of $F''$ and the associated error analysis, as needed for estimating $g'$.
Unlike \citet{FanGuoYu2024}, we do not impose separate smoothness assumptions on the induced residual density $f_{\boldsymbol\theta}(u)$ and purchase-weighted density $h_{\boldsymbol\theta}(u)$,
where $f_{\boldsymbol\theta}(u)$ denotes the density of the induced residual $w_t(\boldsymbol\theta)$ and $h_{\boldsymbol\theta}(u):=f_{\boldsymbol\theta}(u)\mathbb E[y_t\mid w_t(\boldsymbol\theta)=u].$
As shown in the theoretical analysis below, the required smoothness of these densities follows from our exploration design and the smoothness of $F$.
Furthermore,  the Lipschitz condition on $F^{(m)}$ limits how rapidly this derivative can vary and allows uniform control of the kernel smoothing bias.
The lower bound on $F'$ prevents division by a vanishing noise density, while stable evaluation of $g'$ additionally relies on the positive lower bound on $\phi'$ imposed in the
next assumption.

\begin{assumption}
\label{ass:virtual-valuation}
For every \(u\in I_g:=[\delta_z,B-\delta_z]\), \(g(u)-u\in(-\delta_z,\delta_z)\).
There is \(c_\phi>0\) such that \(\phi'(z)\ge c_\phi\) on \((-\delta_z,\delta_z)\).
\end{assumption}

Assumption~\ref{ass:virtual-valuation} imposes two conditions on the optimal pricing function $g$.
The first interior-residual condition ensures that $g(u)\in(0,B)$ for every $u\in I_g$, while Assumption~\ref{ass:noise-regularity} guarantees
$F'(g(u)-u)>0$. The first-order condition therefore gives $\phi(g(u)-u)=-u$.
As in \citet{FanGuoYu2024}, the lower bound on $\phi'$ makes $\phi$ strictly increasing and ensures that the solution to $\phi(z)=-u$ is unique and stable; this bound is implied by,
but does not require, the log-concavity conditions in \citet{Liu2025}.
Consequently, $g(u)=u+\phi^{-1}(-u)$, establishing the pricing
characterization in Remark~\ref{zzjshi}.

\begin{assumption}
\label{ass:manipulation-cost}
For every optimal report $\mathbf x_{t,\mathrm R}$ solving \eqref{eq:buyer-report-problem}, the associated valuation component $\boldsymbol\theta_0^\top\mathbf x_t$ lies in a fixed compact subset of $\operatorname{int}(I_g)$.
\end{assumption}

Assumption~\ref{ass:manipulation-cost} ensures that the valuation component $\boldsymbol\theta_0^\top\mathbf x_t$ induced by every optimal report remains a fixed positive distance from the boundary of $I_g$. Together with Assumptions~\ref{ass:noise-regularity} and~\ref{ass:virtual-valuation}, this ensures that $g$ is differentiable at these valuation components, so every optimal report satisfies the first-order condition in \eqref{eq:manipulation}.
Intuitively, as in \citet{Liu2025}, the positive-definite quadratic cost penalizes every nonzero adjustment, while the off-diagonal entries of $\mathbf A$ capture cost interactions between adjustments to different reportable features. Since a buyer can reduce her price by at most $B$, whereas manipulation costs grow quadratically, any optimal report remains within a bounded distance of the buyer's true reportable features. The interior condition in Assumption~\ref{ass:manipulation-cost} further describes a setting in which buyers may manipulate their reports without optimally pushing the implied valuation components to the extremes of the pricing region.

\begin{assumption}\label{ass:kernel}
The kernel \(K\) satisfies \(\int K(s)\,ds=1\), \(\int s^jK(s)\,ds=0\) for \(j=1,\ldots,m-1\), and \(\int|s^mK(s)|\,ds<\infty\).
Moreover, \(K,K'\), and \(K''\) are Lipschitz and, after a fixed rescaling if necessary, are supported on \([-1,1]\).
\end{assumption}

Assumption~\ref{ass:kernel} follows the compactly supported order-\(m\) kernel condition in \citet{FanGuoYu2024}.
We additionally require $K''$ to be Lipschitz to support uniform estimation of $F''$,  which is needed to estimate $g'$.
Such a kernel can be fixed before pricing begins, and normalizing its support to \([-1,1]\) is without loss of generality.

\subsection{Estimation Guarantees}\label{sec:estimation-guarantees}

Since the interval functions are determined before episode \(k\) begins, we state the following three estimation results conditional on the pre-episode history \(\mathcal H_{k-1}\).

\begin{theorem}
\label{lem:sparse-valuation-main}
 Under Assumptions~\ref{ass:bounded-sparse} and \ref{ass:covariance}, there exists a constant \(B_1>0\) such that, if \(\lambda_k\ge B_1W_x(B+W_xW_\theta)
\{[1+\log\ell_k+\log(d+1)]/a_k\}^{1/2}\) and \(a_k\ge B_1W_x^4s_0^2
\{1+\log\ell_k+\log(d+1)\}/\lambda_{\min}^2\),
 the following bounds hold with probability at least \(1-2\ell_k^{-2}\) on \(\mathcal G_k\):
\begin{equation}
\|\widehat{\boldsymbol\theta}_k-\boldsymbol\theta_0\|_2
\le \frac{3\lambda_k\sqrt{s_0}}{\lambda_{\min}},\qquad
\|\widehat{\boldsymbol\theta}_k-\boldsymbol\theta_0\|_1
\le \frac{12s_0\lambda_k}{\lambda_{\min}}.
\label{eq:main-sparse-estimation}
\end{equation}
\end{theorem}

Recall that SALE-K uses the regularization parameter $\lambda_k=c_\lambda\{(1+\log\ell_k+\log(d+1))/a_k\}^{1/2}$.
Choosing the fixed constant $c_\lambda>0$ sufficiently large ensures that $\lambda_k$ satisfies the lower bound in Theorem~\ref{lem:sparse-valuation-main}.
In addition,  under the exploration schedule in \eqref{eq:width-adaptive-schedule}, the sample-size condition for $a_k$  also holds for all sufficiently large episodes.
Once these requirements are met, Theorem~\ref{lem:sparse-valuation-main} yields, on $\mathcal G_k$, the $\ell_2$ and $\ell_1$ estimation error bounds, $O(\sqrt{s_0}\lambda_k)$ and $O(s_0\lambda_k)$, respectively, without requiring buyer features to be independent over time. Unlike the likelihood-based estimators in \citet{Javanmard2019,WangWangSunCheng2024}, our \(\ell_1\)-regularized least-squares estimator does not require knowledge of $F$.
It also extends the low-dimensional least-squares construction under full-domain exploration in \citet{FanGuoYu2024} to high-dimensional \(\ell_1\)-regularized least-squares estimation with personalized exploration intervals,
while achieving the conventional estimation error bounds.
The proof of Theorem~\ref{lem:sparse-valuation-main} is provided in Appendix~\ref{app:proof-sparse-valuation-main}.

Moreover, we  would like to emphasize that  the estimation error bounds on $\mathcal G_k$ provide a sufficient condition for the next episode's coverage event $\mathcal G_{k+1}$.
To record both bounds, take $C_\theta\ge3/\lambda_{\min}$ and define the confidence set
\begin{equation}
\mathcal C_k=
\left\{
\boldsymbol\theta\in\Theta:
\|\boldsymbol\theta-\widehat{\boldsymbol\theta}_k\|_2
\le C_\theta\lambda_k\sqrt{s_0},
\quad
\|\boldsymbol\theta-\widehat{\boldsymbol\theta}_k\|_1
\le4C_\theta s_0\lambda_k
\right\}.
\label{eq:data-confidence-set}
\end{equation}
Whenever both bounds in Theorem~\ref{lem:sparse-valuation-main} hold, we obtain $\boldsymbol\theta_0\in\mathcal C_k$.
The $\ell_1$ constraint then implies that the fitted mean valuation differs from the true mean valuation by at most $4C_\theta W_xs_0\lambda_k$.
The interval construction in \eqref{eq:safe-exploration-interval} accounts for this difference and the bounded noise by expanding the fitted mean valuation
range at both ends by
$4C_\theta W_xs_0\lambda_k+\delta_z$.
Consequently,
\[
\boldsymbol\theta_0\in\mathcal C_k
\quad\Longrightarrow\quad
\mathcal G_{k+1}.
\]
The $\ell_2$ constraint is used in the subsequent nonparametric analysis to control how parameter estimation error affects the nonparametric estimators.

SALE-K estimates $F$ from the  induced residuals $w_t(\widehat{\boldsymbol\theta}_k)=p_t-\widehat{\boldsymbol\theta}_k^\top\mathbf x_t^0$.
The estimation error in $\widehat{\boldsymbol\theta}_k$ therefore affects the accuracy of the kernel estimator.
In addition, kernel estimation uses observations near each target residual value, and the exploration interval widths affect how many such observations are available on average.
On $\mathcal G_k$, conditional on $\mathcal H_{k-1}$ and the exploration features, each true residual
$w_t(\boldsymbol\theta_0) =p_t-\boldsymbol\theta_0^\top\mathbf x_t^0$ is uniformly distributed over an interval that contains the noise support and has width $W_k(\mathbf x_{t,\mathrm V}^0)$.
For a kernel window $[u-b_k,u+b_k]$ contained in the noise support, the expected number of residual observations in the window is therefore
\[
2b_k\sum_{t\in I_k}
W_k(\mathbf x_{t,\mathrm V}^0)^{-1}
=2a_kb_kq_k,
\]
where
\begin{equation}
q_k
=\frac1{a_k}\sum_{t\in I_k}
W_k(\mathbf x_{t,\mathrm V}^0)^{-1},
\qquad
\frac1{B\omega_k}\le q_k\le\frac1{2\delta_z}.
\label{eq:main-residual-density}
\end{equation}
The bounds follow from $2\delta_z\le W_k(\mathbf x_{t,\mathrm V}^0)\le B\omega_k$ on $\mathcal G_k$.
Consequently, for a fixed sample size and bandwidth, narrower exploration intervals that preserve coverage increase $q_k$ and the expected number of observations available in each local kernel window.
At the same time, averaging purchase outcomes across these windows can introduce smoothing bias when estimating the purchase probability at each window's center.
The following theorem quantifies these effects through uniform error bounds for estimating $F$, $F'$, and $F''$,  with terms corresponding to parameter estimation error, smoothing bias,
and stochastic error, whose magnitude depends on the expected number of observations in each local kernel window.

\begin{theorem}%
\label{lem:noise-function-main}
Under the conditions of Theorem~\ref{lem:sparse-valuation-main} and Assumptions~\ref{ass:noise-regularity} and \ref{ass:kernel},
for any fixed closed interval \(J\) contained in the interior of the noise support,
there exist constants \(B_2,C_2>0\) such that,
if \(a_kb_k^5\ge B_2\Lambda_{\ell_k}\) and \(a_kb_k^2\ge B_2s_0^2
\{1+\log\ell_k+\log(d+1)\}\), the following bounds hold with probability at least \(1-3\ell_k^{-2}\) on \(\mathcal G_k\):
\begin{align}
\sup_{u\in J}|\widehat F_k(u)-F(u)|
&\le C_2\left\{\frac{\lambda_k\sqrt{s_0}}{\lambda_{\min}}+b_k^m
+\sqrt{\frac{\Lambda_{\ell_k}}{a_kb_kq_k}}\right\},
\label{eq:main-F-rate}\\
\sup_{u\in J}|\widehat F_k^{(1)}(u)-F'(u)|
&\le C_2\left\{\frac{\lambda_k\sqrt{s_0}}{\lambda_{\min}}+b_k^{m-1}
+\sqrt{\frac{\Lambda_{\ell_k}}{a_kb_k^3q_k}}\right\},
\label{eq:main-Fprime-rate}\\
\sup_{u\in J}|\widehat F_k^{(2)}(u)-F''(u)|
&\le C_2\left\{\frac{\lambda_k\sqrt{s_0}}{\lambda_{\min}}+b_k^{m-1}
+\sqrt{\frac{\Lambda_{\ell_k}}{a_kb_k^5q_k}}\right\}.
\label{eq:main-Fsecond-rate}
\end{align}
\end{theorem}

On $\mathcal G_k$, Theorem~\ref{lem:noise-function-main} extends the main uniform kernel estimation results for $F$ and $F'$ under full-domain exploration in \citet{FanGuoYu2024} to high-dimensional sparse valuation
models with personalized exploration intervals. The high-dimensional difficulty arises because the residuals depend on a parameter estimated from the same exploration data, requiring uniform control over the relevant parameter
region. Recall that the full-dimensional covering argument used in their main analysis introduces a $\sqrt d$ factor.
However, our analysis instead uses the $\ell_1$ error bound from Theorem~\ref{lem:sparse-valuation-main} to localize the parameter region and applies a contraction argument to control the residual kernel process through
coordinatewise maxima. In the resulting bounds, this replaces the dimensional factor $\sqrt d$ with $\sqrt{s_0\log(d+1)}$.
As shown in this theorem,  personalized exploration provides a further benefit: for a fixed sample size and bandwidth, narrower intervals that preserve coverage increase $q_k$ and reduce the stochastic terms in the error bounds.
The inequality $q_k^{-1}\le B\omega_k$ connects this improvement to the exploration widths used in the regret analysis.
Moreover, our results additionally provide a uniform error bound for $F''$, which is needed to estimate $g'$ for strategic correction.
Together with Assumption~\ref{ass:virtual-valuation}, these estimation guarantees yield the following uniform error bounds for $g$ and $g'$.

\begin{lemma}%
\label{lem:pricing-function-main}
Under the conditions of Theorem~\ref{lem:noise-function-main} and Assumption~\ref{ass:virtual-valuation}, there exist constants \(B_3,C_3>0\) such that, if \(a_kb_k^5\ge B_3\Lambda_{\ell_k}\) and \(a_kb_k^2\ge B_3s_0^2
\{1+\log\ell_k+\log(d+1)\}\),
the following bounds hold with probability at least \(1-3\ell_k^{-2}\) on \(\mathcal G_k\):
\begin{align}
\sup_{u\in I_g}|\widehat g_k(u)-g(u)|
&\le C_3\left\{\frac{\lambda_k\sqrt{s_0}}{\lambda_{\min}}+b_k^{m-1}
+\sqrt{\frac{\Lambda_{\ell_k}}{a_kb_k^3q_k}}\right\},
\label{eq:main-g-rate}\\
\sup_{u\in I_g}|\widehat g_k'(u)-g'(u)|
&\le C_3\left\{\frac{\lambda_k\sqrt{s_0}}{\lambda_{\min}}+b_k^{m-1}
+\sqrt{\frac{\Lambda_{\ell_k}}{a_kb_k^5q_k}}\right\}.
\label{eq:main-gprime-rate}
\end{align}
\end{lemma}

Estimating $g$ requires estimates of $F$ and $F'$, whereas estimating $g'$ additionally requires an estimate of $F''$.
Since $0<\omega_{\min}\le\omega_k\le1$ and $a_k\ge1$, the bandwidth $b_k=(\omega_k/a_k)^{1/(2m+3)}$ satisfies
\[
\left(\frac{\omega_{\min}}{a_k}\right)^{1/(2m+3)}
\le b_k
\le a_k^{-1/(2m+3)}
\le1.
\]
In particular, $0<b_k\le1$, and $b_k\to0$ as $a_k\to\infty$. The additional differentiation makes the stochastic term in the bound for $g'$ equal to $b_k^{-1}$ times the corresponding term for $g$, reflecting the greater
statistical difficulty of derivative estimation. The proofs of Theorem~\ref{lem:noise-function-main} and Lemma~\ref{lem:pricing-function-main} are given in Appendix~\ref{app:pricing-map-estimation}.

\subsection{Self-Recovery and Regret Guarantees}
\label{sec:regret-guarantees}

The preceding estimation guarantees hold with high probability under the coverage event $\mathcal G_k$.
The exploration intervals for episode $k\ge2$, however, are constructed using $\widehat{\boldsymbol\theta}_{k-1}$. An estimation error exceeding the stated bounds in episode $k-1$ may cause coverage failure in episode $k$,
that is, $\mathcal G_k^c$. This means that, conditional on $\mathcal H_{k-1}$, an exploration valuation has a positive probability of falling outside its personalized interval.
The preceding probability bounds cannot be applied to such an episode, so it is necessary to examine whether its potentially biased observations still
allow accurate estimation. To see how this bias arises, uniform price randomization gives
\[
\mathbb E[Y_t\mid\mathbf x_t^0,z_t,\mathcal H_{k-1}]
=
\min\left\{
U_k(\mathbf x_{t,\mathrm V}^0),
\max\left\{L_k(\mathbf x_{t,\mathrm V}^0),v_t\right\}
\right\}.
\]
Thus, valuations inside the interval are recovered without bias, whereas valuations outside it are truncated to the nearest endpoint.
Since the noise has mean zero, the bias in the population regression equation is the average of these truncation errors over the noise.
The interval construction limits this bias by allowing for the full noise range. Specifically, if the fitted mean valuation at a given true feature profile differs from its true mean by at most $r$, an interval miss can occur only when the noise lies within $r$ of a support endpoint, and truncation changes the valuation by at most $r$.
Consequently, when the prediction error is small, both the probability and magnitude of truncation are limited, making the resulting average bias small.
Bounded features and the lower eigenvalue bound connect this bias control to the preceding estimation error, while stability of the conditional second moments allows prediction errors to be compared across consecutive episodes.
The following theorem shows that, when the preceding estimate $\widehat{\boldsymbol\theta}_{k-1}$ is sufficiently accurate, the updated confidence set $\mathcal C_k$ contains $\boldsymbol\theta_0$ with the stated conditional probability, even without assuming $\mathcal G_k$.

\begin{theorem}%
\label{lem:confidence-dynamics-main}
Under Assumptions~\ref{ass:bounded-sparse} and~\ref{ass:covariance}, write $\overline{\boldsymbol\Sigma}_k:=a_k^{-1}\sum_{t\in I_k}\boldsymbol\Sigma_t$
and take $c_\lambda$ sufficiently large.
There exists a constant \(c_4>0\) such that, for every \(k\ge2\), whenever \[\|\widehat{\boldsymbol\theta}_{k-1}-\boldsymbol\theta_0\|_{\overline{\boldsymbol\Sigma}_{k-1}}\le c_4\min\{(d+1)^{-1/2},\sqrt{s_0}\lambda_{k-1}\},\] the following holds with probability at least \(1-2\ell_k^{-2}\) given \(\mathcal H_{k-1}\):
\[
\boldsymbol\theta_0\in\mathcal C_k.
\]
\end{theorem}

Theorem~\ref{lem:confidence-dynamics-main} gives a sufficient condition for restoring $\boldsymbol\theta_0\in\mathcal C_k$, which guarantees coverage in episode $k+1$.
However, this recovery guarantee alone is insufficient to establish the regret bound. The regret analysis must additionally control the estimation errors entering the exploitation prices and the expected total duration of episodes in which coverage or the required estimation bounds fail.
Clearly, coverage ensures that the exploration response has the correct conditional mean, but does not by itself guarantee accurate estimation from a finite sample.
Since the exploitation price depends on $\widehat{\boldsymbol\theta}_k$, $\widehat g_k$, and $\widehat g_k'$, their estimation error bounds are needed to control pricing errors and regret.
Accordingly, let $\mathcal J_k$ denote the event that $\boldsymbol\theta_0\in\mathcal C_{k-1}$ and all estimation error bounds in Theorems~\ref{lem:sparse-valuation-main}--\ref{lem:noise-function-main} and Lemma~\ref{lem:pricing-function-main} hold in episode $k$.
On $\mathcal J_k$, the preceding confidence set ensures current interval coverage, and the current estimation bounds support the analysis of exploitation regret.
For episodes outside $\mathcal J_k$, the following Lemma~\ref{lem:confidence-recovery-main} bounds their expected total duration by $C_5$, from a sufficiently large episode onward.
The proofs of Theorem~\ref{lem:confidence-dynamics-main} and Lemma~\ref{lem:confidence-recovery-main} are given in Appendix~\ref{sec:global-regeneration}.

\begin{lemma}%
\label{lem:confidence-recovery-main}
Suppose that the conditions of Theorems~\ref{lem:sparse-valuation-main}--\ref{lem:confidence-dynamics-main} and Lemma~\ref{lem:pricing-function-main} hold.
Let $k_\star\ge2$ be any deterministic episode index such that the sample-size conditions in Theorems~\ref{lem:sparse-valuation-main}--\ref{lem:noise-function-main} and Lemma~\ref{lem:pricing-function-main}
hold almost surely for every $k\ge k_\star$.
Then there exists a constant $C_5>0$, independent of $k_\star$, such that
\[
\mathbb E\left[
\sum_{k\ge k_\star}
\ell_k\mathbb I\{\mathcal J_k^c\}
\right]\le C_5.
\]
\end{lemma}

Since per-period regret is at most $B$, Lemma~\ref{lem:confidence-recovery-main} implies that episodes $k\ge k_\star$ in which $\mathcal J_k$ fails contribute at most $BC_5$ to expected regret.
For episodes $k\ge k_\star$ in which $\mathcal J_k$ holds, the analysis bounds exploration regret using interval coverage and exploitation regret using the estimation error bounds.
These bounds depend on the exploration widths. Write $\alpha_m=(2m+3)/(4m+1)$ and $\kappa_m=(6m-6)/(4m+1)$ for the horizon and width exponents, respectively.
As shown in the proof, the resulting episode bounds involve the factor $(\ell_k\Lambda_{\ell_k})^{\alpha_m}\omega_k^{\kappa_m}$.
Accordingly, the contribution of the episode widths is summarized by weighting $\omega_k^{\kappa_m}$ with $(\ell_k\Lambda_{\ell_k})^{\alpha_m}$.
Let $K_T$ denote the index of the last episode that starts by time $T$.
For $\pi\in\{\mathrm K,\mathrm U\}$, corresponding to SALE-K and SALE-U, respectively, define the effective exploration width by
\begin{equation}
\omega_{\pi,T}
=
\left[
\frac{
\mathbb E_\pi\!\left[
\sum_{k=1}^{K_T}
(\ell_k\Lambda_{\ell_k})^{\alpha_m}\omega_k^{\kappa_m}
\right]
}{
\sum_{k=1}^{K_T}
(\ell_k\Lambda_{\ell_k})^{\alpha_m}
}
\right]^{1/\kappa_m},
\label{eq:effective-width-factor}
\end{equation}
where $\mathbb E_\pi$ denotes expectation under policy $\pi$.
This definition collects the width-dependent episode terms into a single factor in the aggregate regret bound.
Since $\omega_{\min}\le\omega_k\le1$, it satisfies $\omega_{\min}\le\omega_{\pi,T}\le1$.
The following theorem  gives an unconditional regret bound for SALE-K,  making explicit how the effective exploration width $\omega_{\mathrm K,T}$ affects this bound.

\begin{theorem}%
\label{thm:main_hd}
Under Assumptions~\ref{ass:bounded-sparse}--\ref{ass:kernel}, choose \(c_\lambda\) sufficiently large.
There exist constants \(c_0,C_{\mathrm K}>0\) such that, for  \(T\) satisfying \((T\Lambda_T)^{\alpha_m}\ge c_0\max\left\{
[1+s_0\log(d+1)]^{\frac{2m+3}{2m-2}},
(1+s_0)^{\frac{6m+1}{2m+1}}
[1+\log(d+1)]^{\frac{2m}{2m+1}}
\right\}\), the expected regret of SALE-K satisfies
\begin{equation}
\Reg_{\mathrm K}(T)
\le C_{\mathrm K}\rho_A^4
(\omega_{\mathrm K,T})^{\kappa_m}
(T\Lambda_T)^{\alpha_m},
\label{eq:main-knownA-regret}
\end{equation}
where $\rho_A:=\max\{1,\|\mathbf A^{-1}\|_{\mathrm{op}}\}=\max\{1,\lambda_{\min}(\mathbf A)^{-1}\}$ measures
the sensitivity of strategic reporting.
\end{theorem}

Under the stated horizon conditions, dimension and sparsity enter the displayed factor $(T\Lambda_T)^{\alpha_m}$ through $\Lambda_T=1+\log T+s_0\log(d+1)$.
Using $\omega_{\mathrm K,T}\le1$ gives the worst-case rate
\begin{equation}
\Reg_{\mathrm K}(T)
=
\widetilde O\!\left(
\rho_A^4
\{T[1+s_0\log(d+1)]\}^{\alpha_m}
\right),
\label{eq:main-knownA-worst-case}
\end{equation}
where $\widetilde O$ hides logarithmic factors in $T$.
This worst-case rate suppresses the benefit of personalized exploration by replacing the width factor with one.
To clearly show this benefit, consider Blind ETC-K, which uses the same episode structure, estimation procedures, and exploitation rule as SALE-K,
but explores uniformly over $[0,B]$ and uses $\omega_k=1$ in both its exploration schedule and bandwidth.
Its effective exploration width is therefore one, whereas the sharper SALE-K bound retains the factor $(\omega_{\mathrm K,T})^{\kappa_m}$.
The improvement captured by this factor depends on how much verified features allow the exploration intervals to narrow. %
Moreover, for a fixed width floor $\omega_{\min}>0$, the width factor changes the multiplicative bound without changing the horizon exponent $\alpha_m$.
As $m$ increases, $\alpha_m$ approaches $1/2$, bringing the horizon dependence closer to the classical $\sqrt T$ worst-case benchmark for parametric dynamic pricing
\citep{BroderRusmevichientong2012,FanGuoYu2024}.
The proof of Theorem~\ref{thm:main_hd} is given in Appendix~\ref{app:proof-main-hd}.

\section{Experiments}
\label{sec:Experiments}

\subsection{Experimental Setup}
\label{sec:exp-setup}

We evaluate SALE-K under the known manipulation cost matrix through two synthetic experiments and a real data experiment.
For comparison, we include Blind ETC-K and RMLP-2 \citep{Javanmard2019}. Blind ETC-K uses the same episode structure, estimation procedures, and exploitation rule as SALE-K, but draws exploration prices uniformly from $[0,B]$ and sets $\omega_k=1$ in its exploration schedule and bandwidth.
RMLP-2  is a high-dimensional parametric pricing policy applied to reported features.
Following \citet{FanGuoYu2024}, we implement it using a standard-Gaussian pricing map after monetary normalization.
Following \citet{FanGuoYu2024,Liu2025}, we evaluate performance by cumulative expected regret.
The experiments in this section use $T=819{,}000$ periods and 100 independent replications.
The resulting figures report averages across replications, with shaded regions and error bars indicating one Monte Carlo standard error.

In the synthetic experiments, valuations follow \eqref{eq:valuation-model}, with market noise independent
of buyer features.
The feature and noise distributions are specified in each experiment.
Buyers report truthfully during exploration and follow \eqref{eq:buyer-report-problem} during exploitation.
Reports are restricted to $[-20,20]^{d_{\mathrm R}}$, and all generated optimal reports lie in its interior.
Across all experiments, we use $\ell_0=200$ and specify the manipulation cost matrix using the base matrix
\[
\mathbf A_0=
\begin{pmatrix}
1/4 & 1/8\\
1/8 & 1/4
\end{pmatrix},
\]
following \citet{Liu2025}.
For nonparametric noise estimation, we use the compactly supported kernel
\[
K(u)\propto
(1-11u^2/3)(1-u^2)^3\mathbb I\{|u|\le1\},
\]
normalized to integrate to one.
The estimates of $F$, $F'$, and $F''$ use the same bandwidth, whose specification is given in each experiment.

\subsection{Synthetic Experiments}

\subsubsection{The Value of Personalized Exploration}\label{sec:exp-known-A}

We examine whether the exploration benefit remains when the feature dimension increases while sparsity stays fixed.
We compare SALE-K, Blind ETC-K, and RMLP-2 at $d\in\{20,100,1000\}$, with the first $4d/5$ coordinates verified and the remainder reportable.
True feature coordinates are independently drawn from the Rademacher distribution.
The intercept is $\alpha_0=70$.
The first four verified coefficients are $(-3,3,-3,3)$, and the first reportable coefficient is $-3$; all other coefficients are zero.
Thus $s_0=6$, with verified features accounting for 80\% of the squared coefficient norm.
The noise has the symmetric density
\[
f(z)\propto
\left\{1-(z/\delta_z)^2\right\}^{m}
\mathbb I\{|z|<\delta_z\},
\]
with $m=4$ and $\delta_z=8$.
We set $\mathbf A=3\operatorname{diag}(\mathbf A_0,\ldots,\mathbf A_0)$ and $B=140$.
For SALE-K and Blind ETC-K, we set $c_\lambda=2$, $c_a=0.30$, $C_\theta=4.375$, and $W_\theta=85.5$, and use bandwidth $4(\omega_k/a_k)^{1/(2m+3)}$.

\begin{figure}[htbp]\centering
  \includegraphics[width=\linewidth]{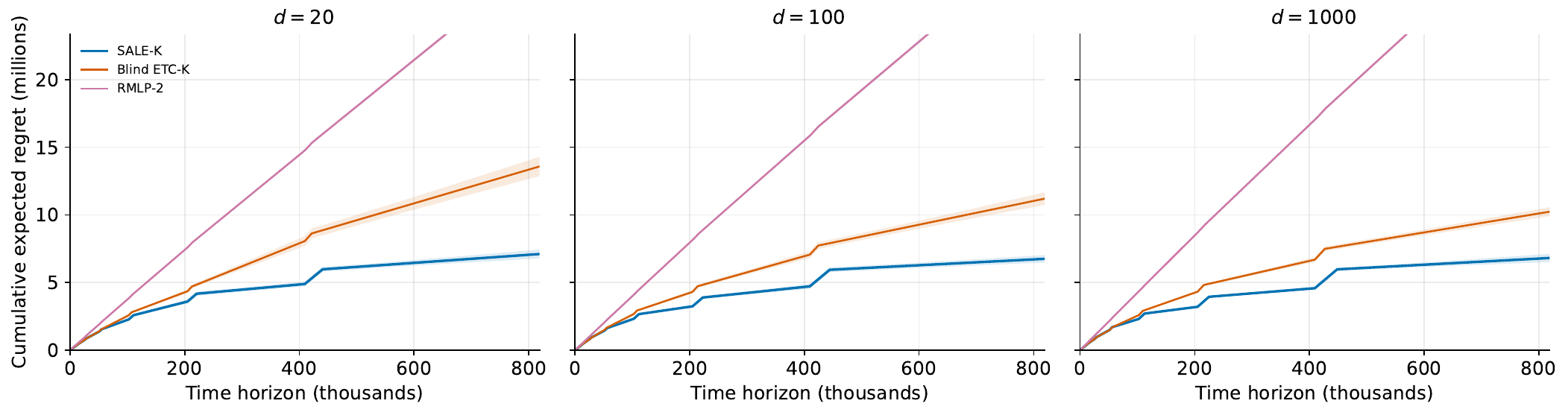}
  \caption{Cumulative expected regret across dimensions with known manipulation cost.}
  \label{fig:highdim-regret}
\end{figure}

Figure~\ref{fig:highdim-regret} shows that SALE-K has lower terminal mean regret than Blind ETC-K in all three dimensions, with reductions of approximately 33--48\%.
Because the two policies use the same estimation and strategic-correction procedures, this comparison demonstrates the benefit of the personalized exploration design, including its width-adaptive schedule, across the dimensions considered.
RMLP-2 has the highest terminal mean regret in all three dimensions, as its Gaussian pricing rule uses reported features without correcting for strategic manipulation.

\FloatBarrier

\subsubsection{The Value of Learning the Noise Distribution}
\label{sec:exp-nongaussian}

We evaluate the value of noise learning by comparing SALE-K with an otherwise identical policy that replaces the learned pricing map and its derivative with their standard-Gaussian counterparts.
There are no verified features and two independent $\operatorname{Unif}[-1,1]$ reportable features, with $\boldsymbol\beta_0=(0.3,0.6)^\top$, $\alpha_0=7.6$, $\mathbf A=\mathbf A_0$, and $B=15$, giving $s_0=3$.
Market noise has the asymmetric density
\[
f(z)\propto
\frac{2}{\sigma}
\varphi\!\left(\frac{z-\mu}{\sigma}\right)
\Phi\!\left(12\frac{z-\mu}{\sigma}\right)
\left\{1-(z/\delta_z)^2\right\}^{m}
\mathbb I\{|z|<\delta_z\},
\qquad
\sigma=\frac{8\delta_z}{27},
\]
where $\varphi$ and $\Phi$ are the standard-Gaussian density and distribution functions, and $\mu$ is chosen to ensure mean-zero noise.
We use $m=3$ and $\delta_z=2.7$. Both pricing maps use the known support $[-\delta_z,\delta_z]$.
For both policies, we set $c_\lambda=0.8$, $c_a=0.30$, $C_\theta=2.50$, and $W_\theta=9$.
SALE-K's noise estimator uses bandwidth $2(\omega_k/a_k)^{1/(2m+3)}$.

\begin{figure}[htbp]\centering
  \includegraphics[width=.94\linewidth]{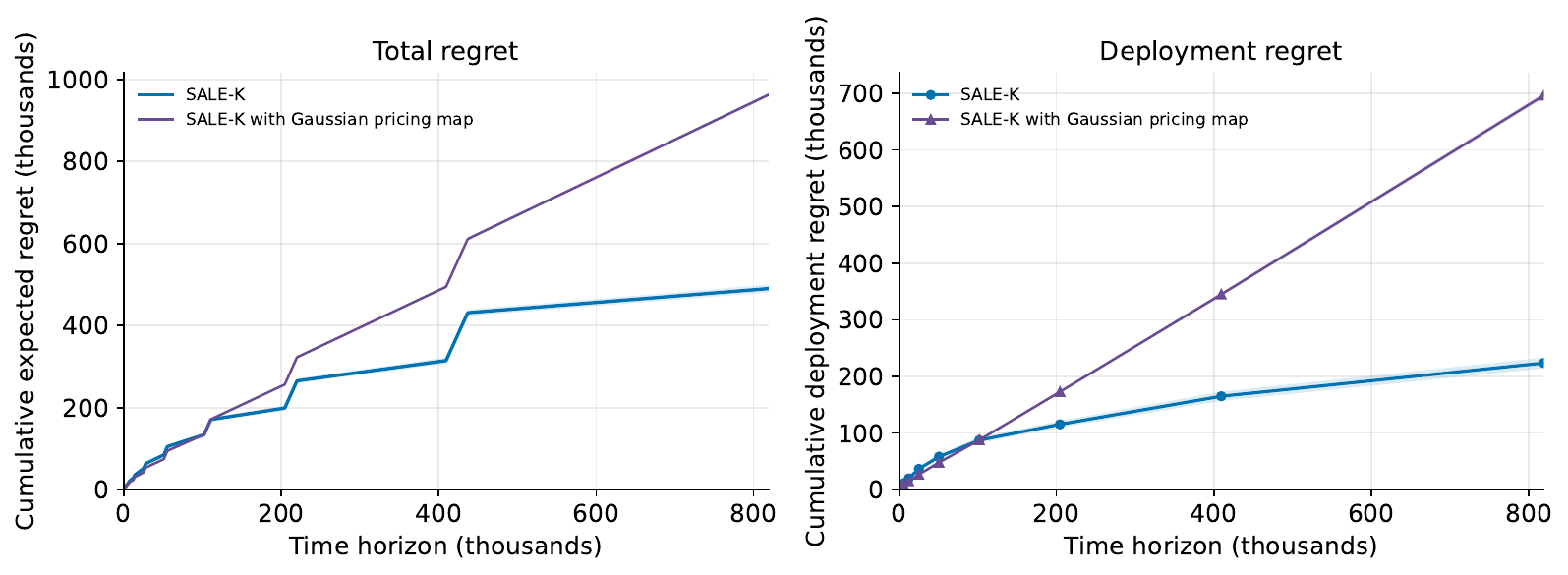}
  \caption{Regret comparison under asymmetric market noise.}
  \label{fig:noise-shape-comparison}
\end{figure}

Using the same exploration rule and random draws, the two policies collect identical exploration data and apply the same estimator to obtain identical valuation estimates.
Their exploration regret is therefore identical in every replication, so their total-regret difference comes entirely from exploitation.
Figure~\ref{fig:noise-shape-comparison} shows that the cumulative exploitation-regret gap widens after the early episodes.
At the terminal horizon, SALE-K's mean total regret is about 49\% lower than that of the Gaussian comparator.
This comparison demonstrates the benefit of learning the noise distribution for exploitation.

\FloatBarrier

\subsection{Real Application}
\label{sec:exp-autoloan}

We compare SALE-K, Blind ETC-K, and RMLP-2 using 208,084 usable AutoLoan application records from Columbia University's Center for Pricing and Revenue Management \citep{PhillipsSimsekVanRyzin2015}.
Historical records do not reveal how buyers would respond to alternative prices.
We therefore follow the calibration approach of \citet{FanGuoYu2024} and \citet{Liu2025} to generate purchase responses under each pricing policy.
We use 70\% of the records to specify valuation coefficients and the remaining 30\% to provide verified features for the pricing experiments, keeping similar historical purchase proportions in both groups.
Following the variable-encoding procedure of \citet{WangWangSunCheng2024}, we construct 64 verified features: six numerical variables and 58 variables indicating category membership by zero or one.
Variables derived from the lender's own quote are excluded.
The numerical variables are rescaled to comparable scales using transformations determined from the first group.

Following the loan-price construction in \citet{FanGuoYu2024} and \citet{WangWangSunCheng2024}, we calculate the historical price as the present value of future repayments minus the approved loan amount.
We discount the repayments using the recorded monthly benchmark interest rate.
Prices are measured in thousands of dollars.
Using the first group, we fit purchase probabilities
with a standard-Gaussian cdf applied to a linear
verified-feature component minus price.
A sparse fit selects five features, whose coefficients
are then re-estimated and used in the simulated
valuation model; the other 59 coefficients are zero.
Each pricing policy subsequently learns its own
valuation estimates from online exploration.

We complete \eqref{eq:valuation-model} with two
independent $\operatorname{Unif}[-1,1]$ true reportable
features, each with coefficient $0.5$, and set
$\mathbf A=\mathbf A_0/2$.
Simulated noise follows the asymmetric density in
Section~\ref{sec:exp-nongaussian}, with $m=4$
and $\delta_z=6$.
Thus, $d=66$ and $s_0=8$, including the intercept.
We set $B=60$, corresponding to $\$60{,}000$,
and adjust the intercept so that valuations satisfy
the model's bounds.
We then set $W_\theta=1.05\|\boldsymbol\theta_0\|_1$.
For online arrivals, we select 40,000 records from
the remaining data and repeatedly draw complete
verified-feature vectors from this pool, allowing
previously drawn vectors to appear again.
Together with simulated reportable features and noise,
these vectors determine $v_t$.
Purchase feedback is generated after each policy
posts its price, as $y_t=\mathbb I\{p_t\le v_t\}$.
We choose numerical constants through preliminary
simulations using 30 repetitions, then fix them
for evaluation.
We use $c_\lambda=5$ and $c_a=0.15$;
for SALE-K and Blind ETC-K, we set $C_\theta=0.60$
and bandwidth $6(\omega_k/a_k)^{1/(2m+3)}$.

\begin{figure}[htbp]\centering
  \includegraphics[width=\linewidth]{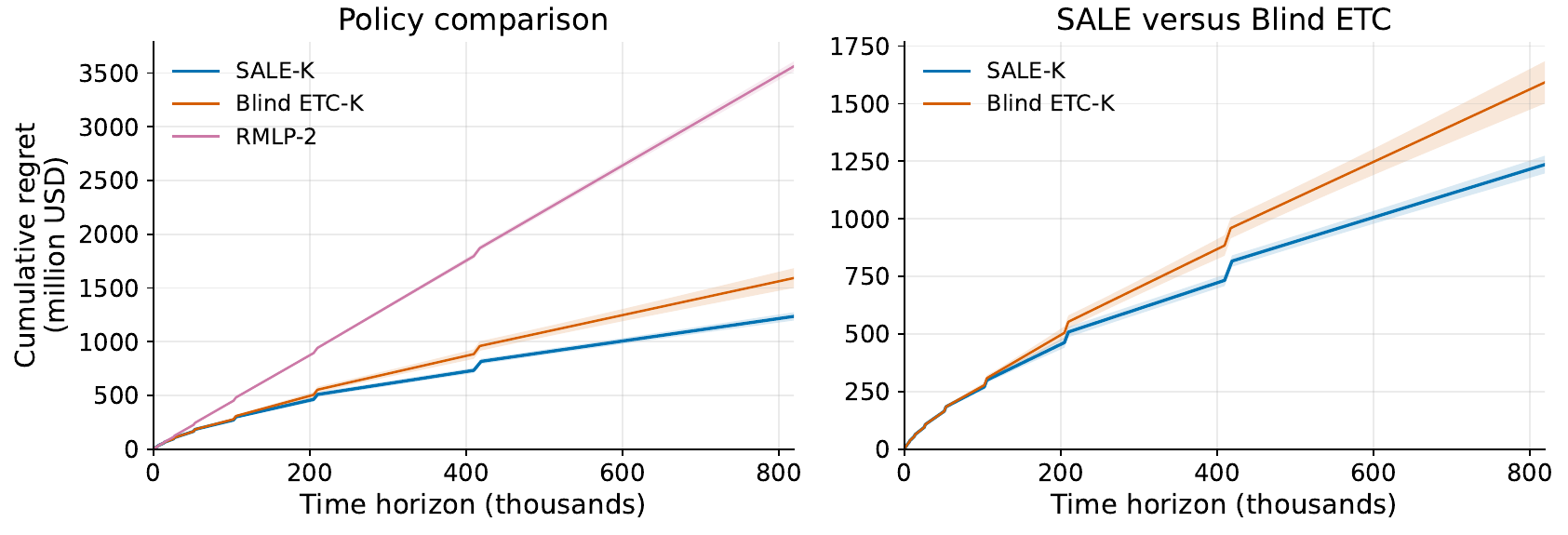}
  \caption{Cumulative expected regret in the AutoLoan study.}
  \label{fig:autoloan}
\end{figure}

Figure~\ref{fig:autoloan} shows that SALE-K and
Blind ETC-K remain close early and separate later.
SALE-K has 22.4\% lower terminal mean regret,
although variability across repetitions remains
substantial.
With their estimation and strategic-correction
procedures unchanged, the comparison supports
personalized exploration and its width-adaptive
schedule when verified features come from real records.
RMLP-2 has the highest terminal mean regret, consistent with a pricing rule that does not learn the asymmetric noise distribution or adjust for manipulated reports.

\FloatBarrier

\section{Conclusion and Discussion}\label{sec:Discussion}

This paper develops a source-aware approach to high-dimensional contextual dynamic pricing with strategic buyers and an unknown market-noise distribution.
The proposed SALE policy uses verified features to personalize randomized exploration and estimates the sparse valuation parameter and noise distribution from the resulting observations for calibrated pricing during exploitation.
The regret analysis quantifies how narrower exploration intervals improve the regret upper bound.
Numerical experiments further illustrate the advantages of the proposed method over the benchmark policies.

We discuss several directions for future work.
First, to reduce the risk of model misspecification, the high-dimensional semiparametric model in this paper could be extended to a high-dimensional doubly nonparametric framework, with both the valuation function and the noise distribution modeled nonparametrically.
Second, verification costs could be incorporated into the model, allowing sellers to jointly decide which features to verify and how to experiment with prices while weighing the benefits of verified information against the required investment.
Last but not the least, the manipulation cost matrix could be allowed to vary across buyers or over time, and pricing rules could be developed to adapt to the resulting variation in buyers' reporting behavior.

\clearpage
\begingroup
\small
\renewcommand{\theHequation}{EC.\arabic{equation}}
\renewcommand{\theHtheorem}{EC.\arabic{theorem}}
\renewcommand{\theHlemma}{EC.\arabic{lemma}}
\renewcommand{\theHremark}{EC.\arabic{remark}}
\renewcommand{\theHfigure}{EC.\arabic{figure}}
\renewcommand{\theHtable}{EC.\arabic{table}}
\section*{Supplementary Material for ``SALE: Source-Aware Dynamic Pricing with Strategic Buyers''}
\appendix
\setcounter{equation}{0}
\renewcommand{\theequation}{EC.\arabic{equation}}
\setcounter{figure}{0}
\renewcommand{\thefigure}{EC.\arabic{figure}}
\setcounter{table}{0}
\renewcommand{\thetable}{EC.\arabic{table}}
\setcounter{theorem}{0}
\renewcommand{\thetheorem}{EC.\arabic{theorem}}
\setcounter{lemma}{0}
\renewcommand{\thelemma}{EC.\arabic{lemma}}
\setcounter{proposition}{0}
\renewcommand{\theproposition}{EC.\arabic{proposition}}
\setcounter{corollary}{0}
\renewcommand{\thecorollary}{EC.\arabic{corollary}}
\setcounter{claim}{0}
\renewcommand{\theclaim}{EC.\arabic{claim}}
\setcounter{conjecture}{0}
\renewcommand{\theconjecture}{EC.\arabic{conjecture}}
\setcounter{hypothesis}{0}
\renewcommand{\thehypothesis}{EC.\arabic{hypothesis}}
\setcounter{assumption}{0}
\renewcommand{\theassumption}{EC.\arabic{assumption}}
\setcounter{remark}{0}
\renewcommand{\theremark}{EC.\arabic{remark}}
\setcounter{example}{0}
\renewcommand{\theexample}{EC.\arabic{example}}
\setcounter{problem}{0}
\renewcommand{\theproblem}{EC.\arabic{problem}}
\setcounter{definition}{0}
\renewcommand{\thedefinition}{EC.\arabic{definition}}
\setcounter{question}{0}
\renewcommand{\thequestion}{EC.\arabic{question}}
\setcounter{answer}{0}
\renewcommand{\theanswer}{EC.\arabic{answer}}
\setcounter{exercise}{0}
\renewcommand{\theexercise}{EC.\arabic{exercise}}
\renewcommand{\theHsection}{appendix.\arabic{section}}
\renewcommand{\theHsubsection}{appendix.\arabic{section}.\arabic{subsection}}
\renewcommand{\theHsubsubsection}{appendix.\arabic{section}.\arabic{subsection}.\arabic{subsubsection}}

\section{Extensions and Implementation}

\subsection{Pricing with Unknown Manipulation Costs}
\label{sec:sale-unknown-a}

\subsubsection{SALE-U Policy}
To deal with the unknown manipulation cost matrix \(\mathbf A\), SALE-U estimates the strategic adjustment
\(\boldsymbol\gamma_0=-\mathbf A^{-1}\boldsymbol\beta_{\mathrm R,0}\) by matching repeat buyers across the two phases,
while retaining SALE-K's episode structure, personalized exploration, and statistical updates.
Let \(\mathrm{id}_t\) denote the buyer's identifier, and suppose that visits
with the same ID share the same true features. The ID-indexed archives
\(\mathcal E_1\) and \(\mathcal E_2\) store each buyer's first exploration and
exploitation records, containing the true and reported features, respectively.
The archive \(\mathcal M\) stores one true--reported feature pair per buyer
once both records are available, regardless of which record was collected first.

\begin{definition}[Effective repeat-buyer rate]
\label{def:repeat-buyer-rate}
For an episodic pricing policy, let \(N_k=\sum_{j\le k}a_j\) denote the
cumulative number of exploration records collected by the end of the
exploration phase in episode \(k\). For \(t\in I_k'\), let \(n_t\) denote the
number of distinct buyers whose exploration and exploitation records have
been matched by time \(t\). The effective repeat-buyer rate is the largest
constant \(\tau\in[0,1]\) such that, almost surely,
\begin{equation}
\max\{1,n_t\}\ge\tau N_k
\qquad\text{for all }k\text{ and }t\in I_k'.
\label{eq:effective-repeat-rate}
\end{equation}
\end{definition}

\begin{remark}
\label{rem:liu-repeat-rate}
Our matching requirement is strictly weaker than the cumulative-time
matched-count conditions used in the proof of Lemma~2 of \citet{Liu2025}.
Their proof uses \(n\ge\tau_{\mathrm L}\sum_{j<k}\ell_j\), where \(n\) is the
matched sample size, \(\tau_{\mathrm L}\) is their repeat-buyer rate, and
\(\ell_j\) is the length of episode \(j\). When episode lengths double from one
episode to the next,
\(N_k\le\sum_{j\le k}\ell_j\le3\sum_{j<k}\ell_j\) for \(k\ge2\), so their
bound guarantees a positive effective repeat-buyer rate under our definition,
with the initial episode covered by \(\max\{1,n_t\}\).
Our condition does not imply their lower bound. For example, take
\(a_k=\lceil\sqrt{\ell_k}\rceil\) new exploration buyers per episode and have
each return during that episode's exploitation phase, starting from a
sufficiently long initial episode. Then \(N_{k-1}\le n_t\le N_k\) for
\(k\ge2\) and \(t\in I_k'\), while
\(N_{k-1}\asymp N_k=o(\sum_{j<k}\ell_j)\). Thus the effective rate is positive,
but their lower bound fails for every fixed \(\tau_{\mathrm L}>0\).
The expected-regret bound in Theorem~\ref{thm:main_hd_unknownA} also holds if,
for some constant \(\tau\in(0,1]\), \eqref{eq:effective-repeat-rate} is replaced by
\(\mathbb E\!\left[
\frac{N_k}{\max\{1,n_t\}}
\,\middle|\,
\mathcal F_{S_k-1}
\right]\le\frac1\tau\),
almost surely for all \(k\) and \(t\in I_k'\).
The argument is given \hyperref[app:proof-unknownA-moment]{at the end of the theorem's proof}.
\end{remark}

For a matched buyer \(i\), Equation~\eqref{eq:manipulation} gives
\begin{equation}\label{eq:delta}
\boldsymbol\delta_i
:=\mathbf x_{i,\mathrm R}-\mathbf x_{i,\mathrm R}^0
=\boldsymbol\gamma_0 g_i',
\qquad
g_i':=g'(\boldsymbol\theta_0^\top\mathbf x_i),
\end{equation}
where \(\mathbf x_i\) is the reported feature vector from the matched
exploitation visit. For each matched buyer, the difference between her reported and true features equals
\(\boldsymbol\gamma_0\) multiplied by \(g'(\boldsymbol\theta_0^\top\mathbf x_i)\).
SALE-U therefore estimates this common vector rather than the full cost matrix \(\mathbf A\).

At time \(t\in I_k'\), SALE-U evaluates
\(\widehat g_i':=\widehat g_k'(\widehat{\boldsymbol\theta}_k^\top\mathbf x_i)\)
for each available pair using the estimates from episode \(k\). It then forms
the vector least-squares estimator
\begin{equation}\label{eq:gamma_est}
\widehat{\boldsymbol\gamma}_t
=
\frac{\sum_{i\in\mathcal M}\widehat g_i'\boldsymbol\delta_i}
{\sum_{i\in\mathcal M}(\widehat g_i')^2}.
\end{equation}
The estimator is used when the required derivative estimates exist and the
denominator is positive. SALE-U prices buyers with stored true features
directly from those features. For other buyers, it uses
\(\widehat{\boldsymbol\gamma}_t\) to correct the reported valuation as in
\eqref{eq:population-corrected-valuation}. When the estimator is unavailable,
it temporarily uses the uncorrected reported valuation. If \(\widehat g_k\)
itself is unavailable, the fallback price is \(B/2\), as in SALE-K.
Algorithm~\ref{alg:unknownA} summarizes the SALE-U policy.
\par\medskip
\begin{breakablealgorithm}
\caption{SALE-U (Unknown $\mathbf A$)}
\label{alg:unknownA}
\begin{algorithmic}[1]
    \STATE \textbf{Input:} Price bound $B$; initial episode length $\ell_0$.
    \STATE Initialize the personalized exploration interval $[L_1(\cdot), U_1(\cdot)]$, where constant functions $L_1(\cdot)\equiv0$ and $U_1(\cdot)\equiv B$.
    \STATE Initialize $\mathcal E_1,\mathcal E_2,\mathcal M\gets\emptyset$.

    \FOR{each episode $k=1,2,\ldots$}
        \STATE Set $\ell_k=2^{k-1}\ell_0$ and
        $S_k=1+\sum_{j<k}\ell_j$.

        \STATE Set the exploration length $a_k$ by
        \eqref{eq:width-adaptive-schedule}.

        \STATE Set $I_k=\{S_k,\ldots,S_k+a_k-1\}$ and
        $I_k'=\{S_k+a_k,\ldots,S_k+\ell_k-1\}$.

        \STATE \textbf{Exploration phase:}
        \FOR{$t\in I_k$}
            \STATE Observe $\mathbf x_{t,\mathrm V}^0$, the truthful
            report
            $\mathbf x_{t,\mathrm R}=\mathbf x_{t,\mathrm R}^0$, and $\mathrm{id}_t$.

            \STATE Draw
            $p_t\sim\mathrm{Unif}\!\left(
            L_k(\mathbf x_{t,\mathrm V}^0),
            U_k(\mathbf x_{t,\mathrm V}^0)\right)$
            and observe $y_t$.

            \STATE If $\mathrm{id}_t\notin\mathcal E_1$, store
            $(\mathrm{id}_t,\mathbf x_t^0)$ in $\mathcal E_1$.
            \STATE If $\mathrm{id}_t\in\mathcal E_2$ and
            $\mathrm{id}_t\notin\mathcal M$, add the corresponding
            true--reported pair to $\mathcal M$.
        \ENDFOR
        \STATE Using observations from
        $I_k$, compute $\widehat{\boldsymbol\theta}_k$ by
        \eqref{eq:least_squares}.

        \STATE Using observations from $I_k$, compute $\widehat g_k$ and $\widehat g_k'$
        by \eqref{eq:ghat}.

        \STATE Set the personalized exploration interval $[L_{k+1}(\cdot),U_{k+1}(\cdot)]$ for episode $k+1$, where the functions $L_{k+1}(\cdot)$ and $U_{k+1}(\cdot)$ are constructed using $\widehat{\boldsymbol\theta}_k$
        according to \eqref{eq:safe-exploration-interval}.

        \STATE \textbf{Exploitation phase:}
        \FOR{$t\in I_k'$}
            \STATE Observe $\mathbf x_{t,\mathrm V}^0$,
            $\mathbf x_{t,\mathrm R}$, and $\mathrm{id}_t$.
            \STATE If $\mathrm{id}_t\notin\mathcal E_2$, store
            $(\mathrm{id}_t,\mathbf x_t)$ in $\mathcal E_2$.
            \STATE If $\mathrm{id}_t\in\mathcal E_1$ and
            $\mathrm{id}_t\notin\mathcal M$, add the corresponding
            true--reported pair to $\mathcal M$.
            \IF{$\mathrm{id}_t\in\mathcal M$}
                \STATE Retrieve $\mathbf x_t^0$ from $\mathcal E_1$ and post
                $p_t=\min\{B,\max\{0,\widehat g_k(
                \widehat{\boldsymbol\theta}_k^\top\mathbf x_t^0)\}\}$.
            \ELSIF{the estimator in \eqref{eq:gamma_est} is available}
                \STATE Compute $\widehat{\boldsymbol\gamma}_t$ by
                \eqref{eq:gamma_est} and post
                \begin{equation*}
                p_t=\min\!\left\{B,\max\!\left\{0,
                \widehat g_k\!\left(
                \widehat{\boldsymbol\theta}_k^\top\mathbf x_t
                -\widehat{\boldsymbol\beta}_{\mathrm R,k}^\top
                \widehat{\boldsymbol\gamma}_t
                \widehat g_k'\!\left(
                \widehat{\boldsymbol\theta}_k^\top\mathbf x_t
                \right)
                \right)\right\}\right\}.
                \end{equation*}
            \ELSE
                \STATE Post $p_t=\min\{B,\max\{0,\widehat g_k(
                \widehat{\boldsymbol\theta}_k^\top\mathbf x_t)\}\}$.
            \ENDIF
            \STATE Observe $y_t$.
        \ENDFOR
    \ENDFOR
\end{algorithmic}
\end{breakablealgorithm}
\smallskip

\subsubsection{Regret Guarantee}

To establish the regret guarantee for SALE-U, we first bound the error in estimating
\(\boldsymbol\gamma_0\) from matched records.
We assume that \(|g'(\boldsymbol\theta_0^\top\mathbf x_i)|\ge c_g>0\) for every matched record \(i\),
so that \eqref{eq:delta} identifies \(\boldsymbol\gamma_0\).
The following lemma bounds this estimation error for any positive lower bound \(\tau\)
on the effective repeat-buyer rate in Definition~\ref{def:repeat-buyer-rate}.

\begin{lemma}
\label{lem:gamma-main}
Fix an episode \(k\).
Under the conditions of Lemma~\ref{lem:pricing-function-main} and Assumption~\ref{ass:manipulation-cost}, suppose \(\tau>0\) and the derivative condition above holds.
There exist constants \(B_6\ge B_3\) and \(C_6>0\), with \(B_6\) independent of \(\tau\) and \(C_6\) independent of \(\rho_A\) and \(\tau\),
such that, if \(a_kb_k^5\ge B_6\Lambda_{\ell_k}\) and \(a_kb_k^2\ge B_6 s_0^2
\{1+\log\ell_k+\log(d+1)\}\), then, except on an event of probability at most \(4\ell_k^{-2}\), the following bound holds on \(\mathcal G_k\) simultaneously for every \(t\in I_k'\) at which the matched-pair estimator is available:
\begin{equation}
\|\widehat{\boldsymbol\gamma}_t-\boldsymbol\gamma_0\|_2^2
\le\frac{C_6\rho_A^4}{\tau}\left\{
\frac{s_0\lambda_k^2}{\lambda_{\min}^2}
+b_k^{2m-2}
+\frac{\Lambda_{\ell_k}}{a_kb_k^5q_k}
\right\}.
\label{eq:main-gamma-rate}
\end{equation}
\end{lemma}

The three terms correspond to sparse estimation error, kernel bias, and stochastic error in estimating \(g'\).
The factor \(1/\tau\) is the price of using the smaller matched sample, while \(\rho_A^4\) records the sensitivity of reports to the pricing rule.
Hence moving from known to unknown manipulation cost adds a matched-sample factor but no new dependence on the ambient dimension or smoothness order.
The proof is in Appendix~\ref{app:proof-gamma}.
We then use this estimation bound in the regret analysis to obtain the following guarantee for SALE-U.

\begin{theorem}
\label{thm:main_hd_unknownA}
Under the conditions of Theorem~\ref{thm:main_hd}, consider a policy--environment pair with effective repeat-buyer rate \(\tau>0\) satisfying the derivative condition in Lemma~\ref{lem:gamma-main}.
There exist constants \(c_0>0\), independent of \(\tau\), and \(C_{\mathrm U,0},C_{\mathrm U,1}>0\), independent of \(\rho_A\) and \(\tau\),
such that, for every horizon \(T\) satisfying \((T\Lambda_T)^{\alpha_m}\ge c_0\max\left\{
[1+s_0\log(d+1)]^{\frac{2m+3}{2m-2}},
(1+s_0)^{\frac{6m+1}{2m+1}}
[1+\log(d+1)]^{\frac{2m}{2m+1}}
\right\}\), the expected regret of SALE-U satisfies
\begin{equation}
\Reg_{\mathrm U}(T)
\le \rho_A^4\left(C_{\mathrm U,0}+\frac{C_{\mathrm U,1}}{\tau}\right)
(\omega_{\mathrm U,T})^{\kappa_m}
(T\Lambda_T)^{\alpha_m}.
\label{eq:main-unknownA-regret}
\end{equation}
\end{theorem}

The decomposition \(C_{\mathrm U,0}+C_{\mathrm U,1}/\tau\) separates the learning cost shared with SALE-K from the additional cost of learning the strategic adjustment from matched records.
Only the latter is amplified by a smaller effective repeat-buyer rate.
Thus, for every fixed \(\tau>0\), SALE-U preserves the horizon and width exponents, with the worst-case rate
\begin{equation}
\Reg_{\mathrm U}(T)
=\widetilde O\left(
\frac{\rho_A^4}{\tau}
\{T[1+s_0\log(d+1)]\}^{\frac{2m+3}{4m+1}}\right).
\label{eq:main-unknownA-worst-case}
\end{equation}

Blind ETC-U is the full-domain counterpart of SALE-U, using the same episode structure, estimators, deployment rule, and matched-record procedure,
but with acquisition prices uniform on \([0,B]\) and \(\omega_k=1\).
Its effective width is one, and the width comparison uses a common positive lower bound on the two policies' effective matching rates.
The proof is given in Appendix~\ref{app:regret-proofs}.

\subsubsection{Numerical Experiment}
\label{sec:exp-unknown-response}

We compare SALE-U with Blind ETC-U and RMLP-2 under unknown manipulation cost, using the simulation setup in Section~\ref{sec:exp-setup}.
Because SALE-U learns the strategic adjustment $\boldsymbol\gamma_0=-\mathbf A^{-1}\boldsymbol\beta_{\mathrm R,0}$ from repeat buyers,
we examine how their availability affects the benefit of personalized exploration.
We use the feature distribution, intercept, noise model, tuning constants, and bandwidth specified at the start of Appendix~\ref{app:supplementary-experiments}.
We set $d=50$, $d_{\mathrm V}=d_{\mathrm R}=25$, $s_0=6$, and $B=100$.
The first two verified coefficients are $(-1/\sqrt2,1/\sqrt2)$ and the first three reportable coefficients are
$(-\sqrt{4/3},\sqrt{4/3},-\sqrt{4/3})$; the remaining coefficients are zero.
The cost matrix has twelve $\mathbf A_0$ blocks and a final diagonal entry $1/4$, with $\mathbf A_0$ defined in Section~\ref{sec:exp-setup}.
The buyer-arrival processes satisfy the expected matching condition in Remark~\ref{rem:liu-repeat-rate}.
We compare target return-visit fractions $\tau\in\{0.01\%,0.5\%\}$.
Visits with the same ID retain the same true features, while market noise is independently drawn at every visit.
All three policies face the same pre-generated buyer paths, market noise, and price-randomization streams;
SALE-U and Blind ETC-U match records across phases in either order.
We use $T=204{,}600$ and 100 replications.

Figure~\ref{fig:strategic-regret} compares the three policies under unknown manipulation cost.
At the two target return-visit fractions, SALE-U's terminal mean regret is 2.67 and 2.42 million monetary units,
compared with 3.22 and 3.06 million for Blind ETC-U and 5.18 and 5.17 million for RMLP-2.
The reductions relative to Blind ETC-U are 17.0\% and 20.6\%.
SALE-U also collects more matched records: its terminal mean counts are 4.30 and 228.49,
compared with 3.27 and 173.23 for Blind ETC-U.
Thus personalized exploration retains its regret benefit when the strategic adjustment is learned from repeat buyers, with the larger relative reduction occurring at $\tau=0.5\%$.

\begin{figure}[htbp]\centering
  \includegraphics[width=\linewidth]{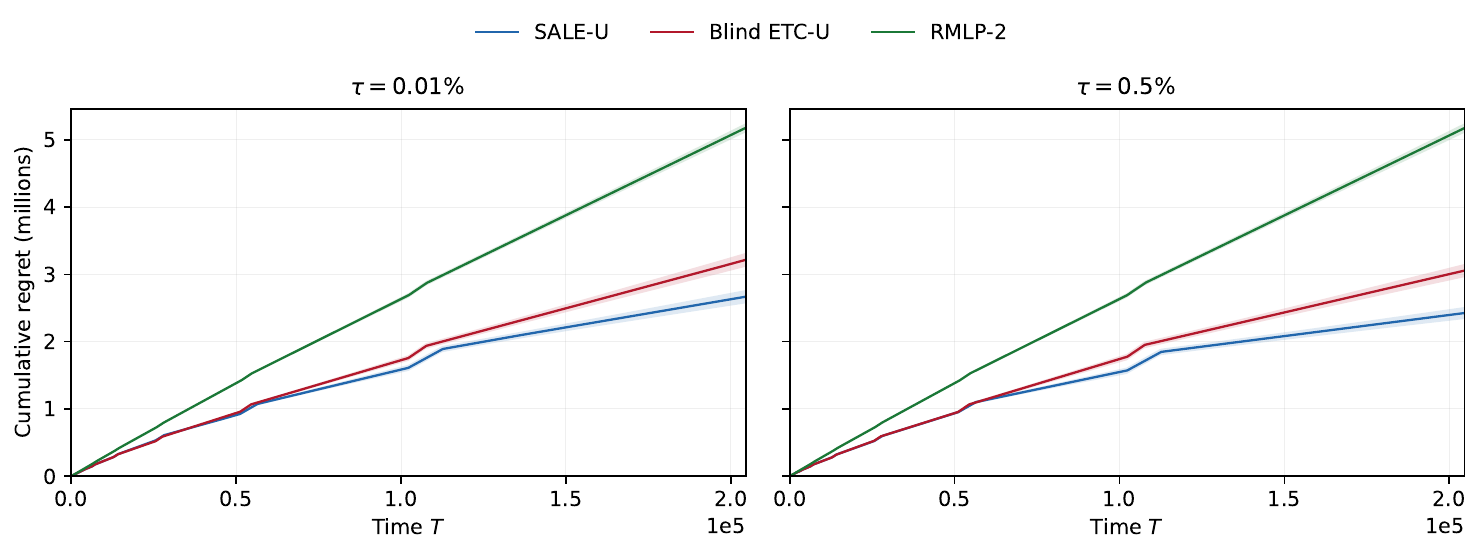}
  \caption{Cumulative expected regret with unknown manipulation cost.}
  \label{fig:strategic-regret}
\end{figure}
\FloatBarrier

\subsection{Computing the Inverse Function}
\label{app:pricing-map-evaluation}

To compute $\widehat g_k(u)$ for a valuation component
$u\in I_g=[\delta_z,B-\delta_z]$, SALE-K must evaluate
$\widehat\phi_k^{-1}(-u)$.
By the definition of the inverse function, this means finding
a solution $z$ to
\[
\widehat\phi_k(z)=-u
\]
and setting $\widehat g_k(u)=u+z$.
Since the estimated function $\widehat\phi_k$ need not be monotone,
the equation may have no solution or multiple solutions.
SALE-K therefore identifies a search interval on which the
solution exists and is unique for every $u\in I_g$.
For episode $k$, it considers the nested intervals
\[
J^q=
\left[
-\delta_z+\frac{\delta_z}{q+1},
\delta_z-\frac{\delta_z}{q+1}
\right],
\qquad q=1,\ldots,k,
\]
and selects the smallest $q$ such that $\widehat\phi_k'>0$
throughout $J^q$ and the range $\widehat\phi_k(J^q)$ contains
$-I_g=[\delta_z-B,-\delta_z]$.
The first condition makes $\widehat\phi_k$ strictly increasing
on $J^q$, ensuring at most one solution.
The second ensures that every target value $-u$, for $u\in I_g$,
is attained, guaranteeing existence.
SALE-K then computes the unique solution by root solving
on the selected interval.
It posts the fallback price $B/2$ if no candidate interval
satisfies these conditions, if an argument of $\widehat g_k$
or $\widehat g_k'$ lies outside $I_g$, or if a required estimate
or derivative is undefined.
The subsequent theoretical analysis uses the estimates $\widehat g_k$ and $\widehat g_k'$ obtained by this procedure.

\subsection{Data-Driven Smoothness and Sparsity Selection}
\label{sec:cvm}

\subsubsection{Smoothness Selection}

The kernel order \(m\) controls the bias--variance trade-off of the nonparametric estimates and, through \(a_k\) and \(b_k\), the exploration schedule.
As in \citet{FanGuoYu2024}, the regret theorems treat \(m\) as known.
When it is unknown, the following cross-validation rule selects it before each episode.
The rule uses only the preceding truthful exploration block.
It is an optional implementation device rather than part of the policies analyzed in the regret theorems.
The experiments use the prespecified smoothness values reported in Section~\ref{sec:Experiments}.
Deriving regret guarantees after smoothness selection is outside the scope of this paper.

At the end of exploration block \(I_{k-1}\), we form the residual \(w_t=p_t-\widehat{\boldsymbol\theta}_{k-1}^\top\mathbf x_t^0\) and retain the purchase indicators \(y_t\).
At the true parameter, \(\mathbb E[y_t\mid w_t(\boldsymbol\theta_0)]=1-F\{w_t(\boldsymbol\theta_0)\}\); for the generated residual, \eqref{eq:derived-conditional-response-stability} controls the change in the population purchase probability.
Partition \(I_{k-1}\) into \(V\) folds \(\mathcal I_1,\ldots,\mathcal I_V\).
For each candidate order \(m\in\mathcal S_m\) (for example, \(\mathcal S_m=\{3,4,6\}\)), we estimate the conditional purchase probability \(\widehat\mu_m(w)\) by Nadaraya--Watson regression with an order-\(m\) kernel and candidate bandwidth \(b_{k-1,m}=(\omega_{k-1}/a_{k-1})^{1/(2m+3)}\),
and score it by \(V\)-fold cross-validated squared error,
\[
   \mathrm{CV}(m)
   =
   \sum_{v=1}^{V}\sum_{t\in\mathcal I_v}
      \{y_t-\widehat\mu_m^{(-v)}(w_t)\}^2 ,
\]
where \(\widehat\mu_m^{(-v)}\) is fitted on the data outside fold \(\mathcal I_v\).
Before episode \(k\), we set \(\widehat m_k=\arg\min_{m\in\mathcal S_m}\mathrm{CV}(m)\) and use \(\widehat m_k\) to choose that episode's exploration length, bandwidth, and kernel order.
Episode 1 uses the conservative candidate \(m_{\min}=\min\mathcal S_m\).

\par\medskip
\begin{breakablealgorithm}
\caption{Data-driven smoothness selection (per episode)}
\label{alg:cvm}
\begin{algorithmic}[1]
\REQUIRE Exploration observations from \(I_{k-1}\), candidate set
\(\mathcal S_m\), and \(V\) folds.
\STATE Form \(w_t=p_t-\widehat{\boldsymbol\theta}_{k-1}^{\top}\mathbf x_t^0\)
and partition \(I_{k-1}\) into \(\mathcal I_1,\ldots,\mathcal I_V\).
\FOR{\(m\in\mathcal S_m\)}
  \STATE Set \(b_{k-1,m}=(\omega_{k-1}/a_{k-1})^{1/(2m+3)}\).
  \FOR{\(v=1,\ldots,V\)}
    \STATE Fit \(\widehat\mu_m^{(-v)}\) by order-\(m\) Nadaraya--Watson
    regression of \(y\) on \(w\) outside fold \(v\).
    \STATE Accumulate
    \(\sum_{t\in\mathcal I_v}\{y_t-\widehat\mu_m^{(-v)}(w_t)\}^2\).
  \ENDFOR
\ENDFOR
\STATE \textbf{return}
\(\widehat m_k\in\arg\min_{m\in\mathcal S_m}\mathrm{CV}(m)\).
\end{algorithmic}
\end{breakablealgorithm}
\smallskip

\subsubsection{Sparsity Selection}

The main algorithms and regret theorems use the sparsity level \(s_0=\|\boldsymbol\theta_0\|_0\), just as the main semiparametric analysis uses the smoothness order \(m\).
When \(s_0\) is not supplied, it can be selected from a finite candidate set using only a completed truthful exploration block.
We retain the notation \(s_0\) in the main theory and use \(\widehat s_k\) only for the practical value selected before episode \(k\).
This selector is likewise optional and is not used in the reported experiments.
The regret theorems remain stated for the model sparsity \(s_0\).
An adaptive confidence guarantee for \(\widehat s_k\) would require additional model separation conditions, which we do not impose.

Let \(\mathcal S_s\subseteq\{1,\ldots,d+1\}\) be a finite grid of candidate sparsity levels.
For each training fold and each \(s\in\mathcal S_s\), first fit the constrained Lasso in \eqref{eq:least_squares}, retain the \(s\) coordinates with largest coefficient magnitude,
and refit the squared-loss criterion on that support subject to the same \(\ell_1\) parameter bound.
Let \(\widehat{\boldsymbol\theta}_s^{(-v)}\) denote the resulting fold-specific estimator.
Score it on the held-out exploration observations by
\[
\mathrm{CV}(s)
=
\sum_{v=1}^{V}\sum_{t\in\mathcal I_v}
\left\{Y_t-(\mathbf x_t^0)^\top
\widehat{\boldsymbol\theta}_s^{(-v)}\right\}^2.
\]
Write \(s_{\min}\in\arg\min_{s\in\mathcal S_s}\mathrm{CV}(s)\), and let \(\mathrm{SE}(s_{\min})\) be the fold-level standard error of its validation score.
To avoid choosing an unnecessarily small model when several candidates predict similarly, use the largest candidate within one standard error of the minimum:
\begin{equation}
\widehat s_k
:=
\max\left\{
s\in\mathcal S_s:
\mathrm{CV}(s)
\le
\mathrm{CV}(s_{\min})+\mathrm{SE}(s_{\min})
\right\}.
\label{eq:data-driven-sparsity}
\end{equation}
Episode 1 uses \(\max\mathcal S_s\).
Subsequent episodes replace \(s_0\) by \(\widehat s_k\) in \(\Lambda_{\ell_k}\), the exploration schedule, the confidence radii, and the exploration interval.

\par\medskip
\begin{breakablealgorithm}
\caption{Data-driven sparsity selection (per episode)}
\label{alg:cvs}
\begin{algorithmic}[1]
\REQUIRE Exploration observations from \(I_{k-1}\), candidate set
\(\mathcal S_s\), and \(V\) folds.
\STATE Partition \(I_{k-1}\) into \(\mathcal I_1,\ldots,\mathcal I_V\).
\FOR{\(s\in\mathcal S_s\)}
  \FOR{\(v=1,\ldots,V\)}
    \STATE Fit the constrained Lasso outside fold \(v\), retain its \(s\)
    largest coordinates, and refit on that support.
    \STATE Accumulate the held-out squared error in \(\mathrm{CV}(s)\).
  \ENDFOR
  \STATE Record \(\mathrm{CV}(s)\) and its fold-level standard error.
\ENDFOR
\STATE \textbf{return} \(\widehat s_k\) from
\eqref{eq:data-driven-sparsity}.
\end{algorithmic}
\end{breakablealgorithm}
\smallskip

\subsection{Supplementary Experiments}
\label{app:supplementary-experiments}

We compare SALE-K with Blind ETC-K in two supplementary experiments using the implementation and evaluation procedures in Section~\ref{sec:exp-setup}.
Both experiments use $T=204{,}600$ periods and 100 independent replications.
New buyers' feature coordinates are independently drawn from a Rademacher distribution.
The intercept is $\alpha_0=50$, and market noise follows the symmetric density in Section~\ref{sec:exp-known-A}, with $m=4$ and $\delta_z=5$.
For both policies, we set $c_\lambda=2$, $c_a=0.30$, $C_\theta=2.50$, and $W_\theta=55.5$, and use bandwidth $b_k=(\omega_k/a_k)^{1/(2m+3)}$.

\subsubsection{Impact of Feature-Source Composition}
\label{sec:exp-source-composition}

We examine how feature-source composition affects the benefit of personalized exploration across four designs with $d=50$, $s_0=6$, and $B=100$.
We combine two candidate-feature allocations, $(d_{\mathrm V},d_{\mathrm R})=(10,40)$ and $(40,10)$, with two allocations of the five nonzero feature coefficients, $(s_{\mathrm V},s_{\mathrm R})=(1,4)$ and $(4,1)$.
The first $s_{\mathrm V}$ verified and $s_{\mathrm R}$ reportable coefficients have magnitudes $\sqrt{5/(2s_{\mathrm V})}$ and $\sqrt{5/(2s_{\mathrm R})}$, respectively.
Their signs alternate from negative across the five active coordinates, ordered with verified coordinates first; all remaining feature coefficients are zero.
Each source therefore has a total squared coefficient norm of $5/2$.
We set $\mathbf A=(5/32)\mathbf I_{d_{\mathrm R}}$, fixing the manipulation strength $\boldsymbol\beta_{\mathrm R,0}^{\top}\mathbf A^{-1}\boldsymbol\beta_{\mathrm R,0}=16$.

Figure~\ref{fig:source-composition} shows that SALE-K has lower terminal mean regret than Blind ETC-K in all four designs.
The reductions are approximately 15--17\% when 10 candidate features are verified and 31--33\% when 40 are verified.
This pattern holds under both allocations of the active coefficients.
With each source's squared coefficient norm and the manipulation strength held fixed, these comparisons show that the allocation of candidate features across sources also affects the benefit of personalized exploration.

\begin{figure}[htbp]\centering
  \includegraphics[width=\linewidth]{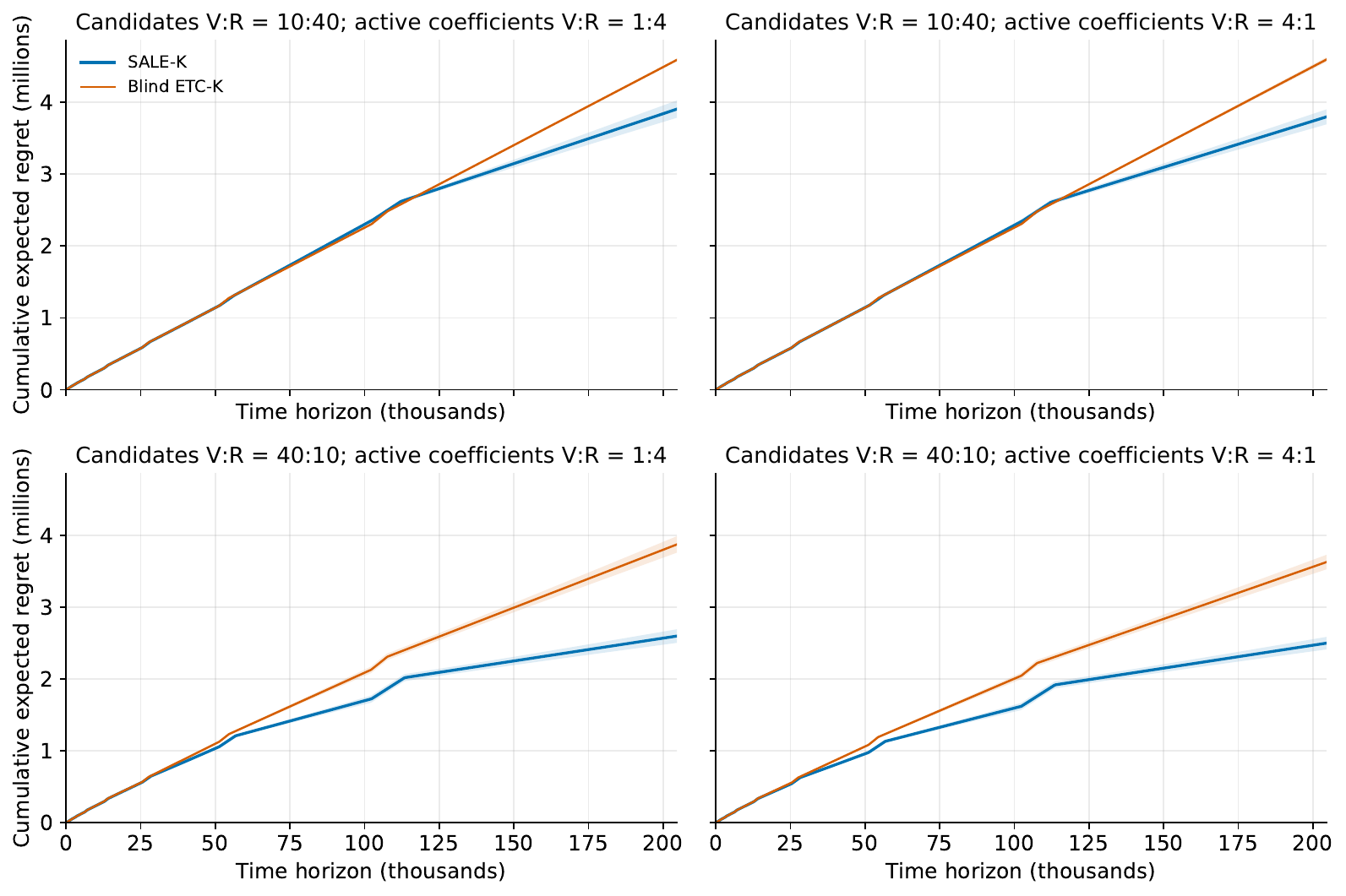}
  \caption{Cumulative expected regret across feature-source compositions.}
  \label{fig:source-composition}
\end{figure}
\FloatBarrier

\subsubsection{Impact of the Price Cap}
\label{sec:exp-price-cap}

We examine how personalized exploration changes the cost of learning as the price domain expands.
We compare SALE-K with Blind ETC-K at $B\in\{65,80,100,140\}$, holding the valuation distribution and manipulation costs fixed.
Here $d=100$, $d_{\mathrm V}=d_{\mathrm R}=50$, and $s_0=6$.
The first three verified coefficients are $(-1,1,-1)$, and the first two reportable coefficients are $(1,-1)$; all other feature coefficients are zero.
Valuations lie in $[40,60]$, inside every price domain considered.
We set $\mathbf A=\operatorname{diag}(\mathbf A_0,\ldots,\mathbf A_0)$, using the base matrix defined in Section~\ref{sec:exp-setup}.
We report the final normalized exploration width, the ratio of SALE-K's cumulative exploration count to Blind ETC-K's, and exploration regret per observation.

Figure~\ref{fig:price-range-mechanism} shows that, as $B$ increases from 65 to 140, SALE-K's mean final normalized exploration width decreases from about 0.88 to 0.30, while its mean exploration-count ratio relative to Blind ETC-K increases from about 1.05 to 2.32.
At every price cap, SALE-K also incurs lower exploration regret per observation than Blind ETC-K.
The width-adaptive schedule thus uses smaller normalized widths to allocate more periods to learning, while personalized pricing keeps exploration regret per observation below that of full-domain exploration.

\begin{figure}[htbp]\centering
  \includegraphics[width=\linewidth]{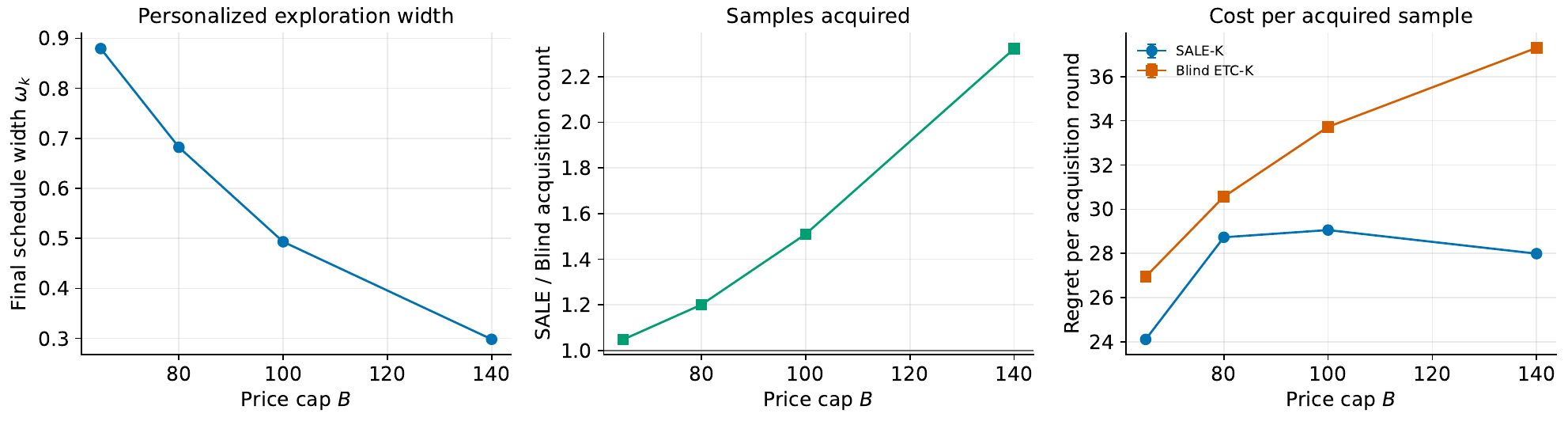}
  \caption{Effects of the price cap on personalized exploration.}
  \label{fig:price-range-mechanism}
\end{figure}
\FloatBarrier

\section{Proofs of Estimation Guarantees}
\label{app:intermediate-proofs}

\subsection{Proof of Theorem~\ref{lem:sparse-valuation-main}}
\label{app:proof-sparse-valuation-main}

Throughout the proofs, unsubscripted \(C,c\) denote positive constants that may change from line to line; named constants remain fixed with the dependence stated when they are introduced.
Fix an episode \(k\). The coverage event introduced in Section~\ref{sec:sale-localized-exploration} can be written as
\begin{equation}
\mathcal G_k:=
\left\{
\mathbb P\!\left(
v_t\in[L_k(\mathbf x_{t,\mathrm V}^0),U_k(\mathbf x_{t,\mathrm V}^0)]
\text{ for all }t\in I_k
\,\middle|\,\mathcal H_{k-1}
\right)=1
\right\}.
\label{eq:coverage-event}
\end{equation}
The event \(\mathcal G_k\) is \(\mathcal H_{k-1}\)-measurable.
On this event, coverage holds almost surely throughout exploration.
Independent price randomization and valuation noise therefore give, for \(t\in I_k\),
\[
\varepsilon_t:=Y_t-(\mathbf x_t^0)^\top\boldsymbol\theta_0,
\qquad
\mathbb E[\varepsilon_t\mid\mathbf x_t^0,\mathcal F_{t-1}]=0.
\]
Since \(0\le Y_t\le B\) and \(|(\mathbf x_t^0)^\top\boldsymbol\theta_0|\le W_xW_\theta\), \(|\varepsilon_t|\le B+W_xW_\theta\).
Consequently, \(\varepsilon_t\mathbf x_t^0\) is a bounded martingale difference.
The proof combines concentration of this score and the sample Gram matrix with the usual Lasso basic inequality.

For the remainder of this proof, write \(n=|I_k|\).
Let \(\mathbf X\in\mathbb R^{n\times(d+1)}\) have rows \((\mathbf{x}_t^0)^\top\) and let \(\mathbf Y_k=(Y_t)_{t\in I_k}\).
Define the empirical loss by
\[
\mathcal L_k(\boldsymbol{\theta})
=
n^{-1}\|\mathbf Y_k-\mathbf X\boldsymbol{\theta}\|_2^2.
\]
Write \(\widehat{\boldsymbol\Sigma}_k=n^{-1}\mathbf X^\top\mathbf X\), and recall the episode's predictable Gram matrix \(\overline{\boldsymbol\Sigma}_k\) defined in Section~\ref{sec:regret-guarantees}.
The event
\begin{equation*}
\mathcal E_{k,\mathrm{RE}}
:=\left\{\|\widehat{\boldsymbol\Sigma}_k-\overline{\boldsymbol\Sigma}_k\|_{\max}
\le\frac{\lambda_{\min}}{32s_0}\right\}
\end{equation*}
provides restricted curvature on the Lasso cone.
Under Assumptions~\ref{ass:bounded-sparse} and \ref{ass:covariance}, let \(S=\supp(\boldsymbol\theta_0)\), fix \(\delta\in(0,1/2)\), and suppose \(\lambda_k\ge8W_x(B+W_xW_\theta)
\{[\log(d+1)+\log(1/\delta)]/n\}^{1/2}\) and \(n\ge2048W_x^4s_0^2
\{2\log(d+1)+\log(2/\delta)\}/\lambda_{\min}^2\).

Let \(\boldsymbol\Delta_k=\widehat{\boldsymbol\theta}_k-\boldsymbol\theta_0\).
For \(j,l\in[d+1]\), \(\boldsymbol\Sigma_t\) is \(\mathcal F_{t-1}\)-measurable and bounded features make \(x_{tj}^0x_{tl}^0-(\boldsymbol\Sigma_t)_{jl}\) a bounded martingale difference.
Conditional Hoeffding--Azuma and a union bound over the \((d+1)^2\) entries give
\begin{equation}
\mathbb P(\mathcal E_{k,\mathrm{RE}}^c\mid\mathcal H_{k-1})
\le2(d+1)^2\exp\left\{-\frac{n\lambda_{\min}^2}
{2048W_x^4s_0^2}\right\}
=:\varepsilon_{\mathrm{RE}}(n).
\label{eq:RE-remainder}
\end{equation}

For later use, write \(\mathcal C(S,3):=\{\mathbf v:\|\mathbf v_{S^c}\|_1\le3\|\mathbf v_S\|_1\}\) for the Lasso cone.
If \(\mathbf v\in\mathcal C(S,3)\), then \(\|\mathbf v\|_1\le4\sqrt{s_0}\|\mathbf v\|_2\).
Hence, on \(\mathcal E_{k,\mathrm{RE}}\),
\[
\mathbf v^\top\widehat{\boldsymbol\Sigma}_k\mathbf v
\ge \mathbf v^\top\overline{\boldsymbol\Sigma}_k\mathbf v-
\|\widehat{\boldsymbol\Sigma}_k-\overline{\boldsymbol\Sigma}_k\|_{\max}\|\mathbf v\|_1^2
\ge\frac{\lambda_{\min}}2\|\mathbf v\|_2^2,
\]
where the last step uses \(\lambda_{\min}(\overline{\boldsymbol\Sigma}_k)\ge\lambda_{\min}\).

For the score, \(\nabla\mathcal L_k(\boldsymbol\theta_0)
=-2n^{-1}\sum_{t\in I_k}\varepsilon_t\mathbf x_t^0\).
The conditional centering and boundedness established above, followed by a conditional Hoeffding--Azuma inequality and a union bound, give
\[
\mathbb P\left(
\|\nabla\mathcal L_k(\boldsymbol\theta_0)\|_\infty>\lambda_k/2
\,\middle|\,\mathcal H_{k-1}\right)
\le2(d+1)\exp\left\{-\frac{n\lambda_k^2}
{32W_x^2(B+W_xW_\theta)^2}\right\}\le\delta.
\]

On \(\mathcal E_{k,\mathrm{RE}}\) and the score bound above, optimality and \((\boldsymbol\theta_0)_{S^c}=0\) give
\[
\boldsymbol\Delta_k^\top\widehat{\boldsymbol\Sigma}_k\boldsymbol\Delta_k
+\frac{\lambda_k}{2}\|(\boldsymbol\Delta_k)_{S^c}\|_1
\le\frac{3\lambda_k}{2}\|(\boldsymbol\Delta_k)_S\|_1.
\]
Thus \(\boldsymbol\Delta_k\) satisfies the cone inequality, and the preceding curvature bound gives
\[
\frac{\lambda_{\min}}2\|\boldsymbol\Delta_k\|_2^2
\le\frac{3\lambda_k}{2}\sqrt{s_0}\|\boldsymbol\Delta_k\|_2.
\]
Division when \(\boldsymbol\Delta_k\ne0\) gives the \(\ell_2\) bound, and the cone inequality gives \(\|\boldsymbol\Delta_k\|_1\le4\sqrt{s_0}\|\boldsymbol\Delta_k\|_2\).
Thus the restricted-eigenvalue and score events imply
\begin{equation}
\begin{gathered}
\|\widehat{\boldsymbol\theta}_k-\boldsymbol\theta_0\|_2
\le \frac{3\lambda_k\sqrt{s_0}}{\lambda_{\min}},\qquad
\|\widehat{\boldsymbol\theta}_k-\boldsymbol\theta_0\|_1
\le \frac{12s_0\lambda_k}{\lambda_{\min}},\\
\|(\widehat{\boldsymbol\theta}_k-\boldsymbol\theta_0)_{S^c}\|_1
\le3\|(\widehat{\boldsymbol\theta}_k-\boldsymbol\theta_0)_S\|_1.
\end{gathered}
\label{eq:raw-lasso-rate}
\end{equation}
The sample-size condition ensures \(\varepsilon_{\mathrm{RE}}(n)\le\delta\).
A union bound therefore gives the stated conclusions with conditional probability at least \(1-2\delta\); independence of the Gram and score bounds is not required.

For the episode-level application, let \(c_\delta>0\) be a fixed exponent controlling the decay of the failure probability with dimension, and set \(\delta_k:=(d+1)^{-c_\delta}\ell_k^{-2}\).
We choose this exponent sufficiently large to make the expected recovery cost summable; it is used only in the analysis and is not an input to the policy.
Take \(n=a_k\) and \(\delta=\delta_k\) in the preceding bounds.
Since \(\log(1/\delta_k)\le C\{1+\log\ell_k+\log(d+1)\}\), the two conditions in Theorem~\ref{lem:sparse-valuation-main}, after enlarging \(B_1\), imply the finite-sample penalty and sample-size requirements.
The bounds in \eqref{eq:main-sparse-estimation} therefore hold with conditional probability at least \(1-2\delta_k\ge1-2\ell_k^{-2}\).
This proves Theorem~\ref{lem:sparse-valuation-main}.\hfill\(\square\)

\subsection{Proofs of Theorem~\ref{lem:noise-function-main} and Lemma~\ref{lem:pricing-function-main}}
\label{app:pricing-map-estimation}
\label{app:proof-noise-function}
\label{app:proof-pricing-function}

The main difficulty is the empirical-process term obtained after substituting the same-sample Lasso estimate into the residual kernel.
The residual \(w_t(\boldsymbol\theta)=p_t-\mathbf x_t^{0\top}\boldsymbol\theta\) is indexed by a high-dimensional parameter estimated from the same observations.
A full-dimensional Euclidean covering, as in \citet{FanGuoYu2024}, introduces a \(\sqrt d\) term and is therefore unsuitable when \(d\) exceeds the exploration sample size.
Here the Lasso error lies in a local cone but need not be exactly sparse.
Local contraction over its \(\ell_1\) radius reduces the parameter complexity to coordinatewise maxima, so dimension enters through \(s_0\log(d+1)\) rather than \(d\).

Source-aware exploration also produces independent, nonidentically distributed residuals after the realized context path is fixed, because exploration widths may differ across buyers.
Their mixture has constant local density equal to the average inverse width \(q_k\) on the region used for smoothing.
A narrower interval therefore increases the effective local sample size.
The resulting uniform bounds yield Theorem~\ref{lem:noise-function-main}; the deterministic inverse-map argument below then gives Lemma~\ref{lem:pricing-function-main}.

\subsubsection{Proof of Theorem~\ref{lem:noise-function-main}}

Fix an episode \(k\) and a closed interval \(J\) contained in the interior of the noise support.
Let \(S=\supp(\boldsymbol\theta_0)\), and write
\[
R_{2,k}:=\frac{3\lambda_k\sqrt{s_0}}{\lambda_{\min}},
\qquad
R_{1,k}:=4\sqrt{s_0}\,R_{2,k},
\]
for the \(\ell_2\)- and \(\ell_1\)-radii supplied by the Lasso bound.
The corresponding local parameter set is
\[
\Theta_k
:=\left\{\boldsymbol\theta\in\Theta:
\boldsymbol\theta-\boldsymbol\theta_0\in\mathcal C(S,3),\quad
\|\boldsymbol\theta-\boldsymbol\theta_0\|_2\le R_{2,k}\right\}.
\]
Every \(\boldsymbol\theta\in\Theta_k\) satisfies \(\|\boldsymbol\theta-\boldsymbol\theta_0\|_1\le R_{1,k}\).
Equation \eqref{eq:raw-lasso-rate} places \(\widehat{\boldsymbol\theta}_k\) in this set, so a bound proved uniformly over \(\Theta_k\) may be evaluated at the random Lasso estimator without sample splitting.
No support-recovery claim is used.

Let \(n=|I_k|\) and \(I_z=[-\delta_z,\delta_z]\).
Work on \(\mathcal G_k\) and condition on \(\mathcal H_{k-1}\) and the realized exploration contexts \(\{\mathbf x_t^0:t\in I_k\}\).
The conditional arguments below are for almost every such context path.
Until the probability statement is restored at the end of the proof, all expectations and probabilities are under this conditional law.
Under the exogenous arrival model in Section~\ref{sec:Setting}, the fresh exploration draws and valuation noises retain their original distributions and remain mutually independent under this conditioning.
The exploration observations are therefore independent, though not identically distributed.

To identify the population targets of the kernel estimators, draw \(t\) uniformly from \(I_k\).
Let \(f_{\boldsymbol\theta}\) be the density of the generated residual \(w_t(\boldsymbol\theta)\), and define
\[
h_{\boldsymbol\theta}(u)
:=f_{\boldsymbol\theta}(u)\,\mathbb P\{y_t=1\mid w_t(\boldsymbol\theta)=u\},
\qquad
r_{\boldsymbol\theta}(u):=\frac{h_{\boldsymbol\theta}(u)}{f_{\boldsymbol\theta}(u)}.
\]
Thus, \(f_{\boldsymbol\theta}\) is the residual density, \(h_{\boldsymbol\theta}\) is its purchase-weighted counterpart, and \(r_{\boldsymbol\theta}\) is the conditional purchase probability.
They are the population counterparts of the kernel denominator, numerator, and ratio, respectively.

On \(\mathcal G_k\), \eqref{eq:coverage-event} and the tower property give conditional almost-sure coverage for almost every exploration-context path.
The noises retain distribution \(F\) under this conditioning.
Since \(-\delta_z\) and \(\delta_z\) are the endpoints of the noise support specified in Section~\ref{sec:Setting}, this implies, simultaneously for \(t\in I_k\),
\begin{equation}
L_k(\mathbf x_{t,\mathrm V}^0)
\le\boldsymbol\theta_0^\top\mathbf x_t^0-\delta_z,
\qquad
U_k(\mathbf x_{t,\mathrm V}^0)
\ge\boldsymbol\theta_0^\top\mathbf x_t^0+\delta_z.
\label{eq:coverage-residual-margin}
\end{equation}
If either inequality failed, the interval would exclude a noise tail of positive \(F\)-probability, contradicting conditional almost-sure coverage.

Work on \(\mathcal G_k\cap\mathcal E_{k,\mathrm{RE}}\) under \(b_k+W_xR_{1,k}\le\operatorname{dist}(J,\partial\supp(z_1))/2\), and let \(q_k=\frac1n\sum_{t\in I_k}W_k(\mathbf x_{t,\mathrm V}^0)^{-1}\).
For \(\boldsymbol\theta\in\Theta_k\), \(|\mathbf x_t^{0\top}(\boldsymbol\theta-\boldsymbol\theta_0)|\le W_xR_{1,k}\).
For each \(t\in I_k\), the variable \(w_t(\boldsymbol\theta)=p_t-\mathbf x_t^{0\top}\boldsymbol\theta\) is uniform on \([L_k(\mathbf x_{t,\mathrm V}^0)-\mathbf x_t^{0\top}\boldsymbol\theta,
U_k(\mathbf x_{t,\mathrm V}^0)-\mathbf x_t^{0\top}\boldsymbol\theta]\).
Together with \eqref{eq:coverage-residual-margin}, the stated margin condition therefore gives, whenever \(\operatorname{dist}(u,J)\le b_k\), \(u+\mathbf x_t^{0\top}(\boldsymbol\theta-\boldsymbol\theta_0)\in I_z\).
Thus the support of \(w_t(\boldsymbol\theta)\) contains every kernel window centered in \(J\), and the residual has conditional density \(W_k(\mathbf x_{t,\mathrm V}^0)^{-1}\) on these windows.
On \(\mathcal G_k\), \(2\delta_z\le W_k(\mathbf x_{t,\mathrm V}^0)\le B\omega_k\) on each exploration period, where the upper bound is \eqref{eq:width-adaptive-schedule}.
Averaging over \(I_k\) gives
\begin{equation}
f_{\boldsymbol\theta}(u)
=q_k,\qquad (B\omega_k)^{-1}\le q_k\le(2\delta_z)^{-1}.
\label{eq:local-residual-density-identity}
\end{equation}

If \(w_t(\boldsymbol\theta)=u\), then \(p_t-\boldsymbol\theta_0^\top\mathbf x_t^0
=u+\mathbf x_t^{0\top}(\boldsymbol\theta-\boldsymbol\theta_0)\).
Conditional purchase probability and averaging give, on the same neighborhood,
\begin{equation}
h_{\boldsymbol\theta}(u)
=\frac1n\sum_{t\in I_k}
\frac{1-F\{u+\mathbf x_t^{0\top}(\boldsymbol\theta-\boldsymbol\theta_0)\}}
{W_k(\mathbf x_{t,\mathrm V}^0)}.
\label{eq:local-generated-response-identity}
\end{equation}
Dividing this identity by \(q_k\) shows that \(r_{\boldsymbol\theta}(u)\) is a weighted average of \(1-F\{u+\mathbf x_t^{0\top}(\boldsymbol\theta-\boldsymbol\theta_0)\}\).
The same weights and Cauchy--Schwarz give
\[
 \frac1{nq_k}\sum_{t\in I_k}
 \frac{|\mathbf x_t^{0\top}(\boldsymbol\theta-\boldsymbol\theta_0)|}
 {W_k(\mathbf x_{t,\mathrm V}^0)}
 \le\left\{\frac{B\omega_k}{2\delta_z}
 (\boldsymbol\theta-\boldsymbol\theta_0)^\top\widehat{\boldsymbol\Sigma}_k
(\boldsymbol\theta-\boldsymbol\theta_0)\right\}^{1/2}
\le C\|\boldsymbol\theta-\boldsymbol\theta_0\|_2.
\]
The last inequality uses \(\mathcal E_{k,\mathrm{RE}}\) and the cone inequality in the definition of \(\Theta_k\), which give
\[
(\boldsymbol\theta-\boldsymbol\theta_0)^\top\widehat{\boldsymbol\Sigma}_k
(\boldsymbol\theta-\boldsymbol\theta_0)
\le(\lambda_{\max}+\lambda_{\min}/2)
\|\boldsymbol\theta-\boldsymbol\theta_0\|_2^2.
\]

Equation~\eqref{eq:local-generated-response-identity}, the weighted bound, and the mean-value theorem applied to \(F,F'\), and \(F''\) give, for a common finite constant \(C_{\mathrm{des}}\),
\begin{equation}
\begin{gathered}
r_{\boldsymbol\theta_0}(u)=1-F(u),\qquad u\in J,\\
\max_{0\le j\le2}\sup_{u\in J}
|r_{\boldsymbol\theta}^{(j)}(u)-r_{\boldsymbol\theta_0}^{(j)}(u)|
\le C_{\mathrm{des}}\|\boldsymbol\theta-\boldsymbol\theta_0\|_2.
\end{gathered}
\label{eq:derived-conditional-response-stability}
\end{equation}
Finally, \(h_{\boldsymbol\theta}=q_kr_{\boldsymbol\theta}\) on the \(b_k\)-neighborhood of \(J\).
Differentiating this identity and using the uniformly bounded derivatives in Assumption~\ref{ass:noise-regularity} give
\[
|h_{\boldsymbol\theta}^{(j)}(u)|\le C_{\mathrm{des}}q_k,\qquad 0\le j\le m,
\]
and \(h_{\boldsymbol\theta}^{(m)}\) is \(C_{\mathrm{des}}q_k\)-Lipschitz on the same neighborhood.
Every argument lies in a fixed compact subset of the noise-support interior, so \(C_{\mathrm{des}}\) can be chosen to depend only on the fixed feature, covariance, width, and noise-regularity bounds, uniformly over the conditioned histories and \(\boldsymbol\theta\in\Theta_k\).

Let \(b_k\) be the SALE bandwidth specified in Section~\ref{sec:sale-learning}.
Define the kernel estimators
\[
\begin{aligned}
h_k(u,\boldsymbol\theta)
&=\frac{1}{n b_k}\sum_{t\in I_k}
K\!\left(\frac{w_t(\boldsymbol\theta)-u}{b_k}\right)y_t,\\
f_k(u,\boldsymbol\theta)
&=\frac{1}{n b_k}\sum_{t\in I_k}
K\!\left(\frac{w_t(\boldsymbol\theta)-u}{b_k}\right).
\end{aligned}
\]
For \(j=0,1,2\), write \(K_j=K^{(j)}\), set \(h_k^{(0)}=h_k\) and \(f_k^{(0)}=f_k\), and define
\[
h_k^{(j)}(u,\boldsymbol\theta)=\frac{(-1)^j}{nb_k^{j+1}}
\sum_{t\in I_k}K_j\!\left(\frac{w_t(\boldsymbol\theta)-u}{b_k}\right)y_t,
\quad
f_k^{(j)}(u,\boldsymbol\theta)=\frac{(-1)^j}{nb_k^{j+1}}
\sum_{t\in I_k}K_j\!\left(\frac{w_t(\boldsymbol\theta)-u}{b_k}\right).
\]

We next control the deterministic smoothing error.
The argument is local because \eqref{eq:local-residual-density-identity} holds only on the stated neighborhood of \(J\).
Under these coverage and margin conditions and Assumptions~\ref{ass:noise-regularity} and \ref{ass:kernel}, there is a constant \(C_{\mathrm{bias}}>0\), depending only on \(C_{\mathrm{des}}\) and the fixed kernel and smoothness bounds, such that, uniformly over \(J\times\Theta_k\),
\begin{equation*}
\begin{aligned}
\sup_{J\times\Theta_k}|\mathbb Eh_k-h_{\boldsymbol\theta}|
&\le C_{\mathrm{bias}}q_kb_k^m,\\
\sup_{J\times\Theta_k}|\mathbb Eh_k^{(j)}-h_{\boldsymbol\theta}^{(j)}|
&\le C_{\mathrm{bias}}q_kb_k^{m-1},\qquad j=1,2,\\
\sup_{J\times\Theta_k}|\mathbb Ef_k^{(j)}-f_{\boldsymbol\theta}^{(j)}|
&=0,\qquad j=0,1,2.
\end{aligned}
\end{equation*}
To see this, work under the conditional law fixed above.
For \(j=0,1,2\), integration by parts \(j\) times, followed by the change of variables \(w=u+b_ks\), gives
\[
\mathbb Eh_k^{(j)}(u,\boldsymbol\theta)
=\int K(s)h_{\boldsymbol\theta}^{(j)}(u+b_ks)\,ds.
\]
The margin condition used in \eqref{eq:local-residual-density-identity} keeps every Taylor segment in the same neighborhood.
For \(j=0\), expansion through order \(m-1\), the kernel moment conditions, and the bound on \(h_{\boldsymbol\theta}^{(m)}\) yield bias \(Cq_kb_k^m\).
For \(j=1\), expansion of \(h_{\boldsymbol\theta}'\) through order \(m-2\) leaves a remainder bounded by \(Cq_kb_k^{m-1}|s|^{m-1}\).

For \(j=2\), expand \(h_{\boldsymbol\theta}''\) through order \(m-2\).
The term involving \(h_{\boldsymbol\theta}^{(m)}(u)(b_ks)^{m-2}\) vanishes because \(\int s^{m-2}K(s)\,ds=0\), while the Lipschitz continuity of \(h_{\boldsymbol\theta}^{(m)}\) bounds the remainder by \(Cq_kb_k^{m-1}|s|^{m-1}\).
Integrability of the corresponding kernel moments proves the numerator bounds.
Finally, \eqref{eq:local-residual-density-identity} makes \(f_{\boldsymbol\theta}\) constant on every kernel window centered in \(J\), which proves the three density identities.
The preceding remainder bounds depend only on \(C_{\mathrm{des}}\) and the fixed kernel and smoothness bounds; choosing \(C_{\mathrm{bias}}\) to dominate them proves the displayed bounds.

We next control the stochastic error uniformly over \(\Theta_k\) using the following lemma, whose proof is given in Appendix~\ref{app:proof-kernel-stochastic}.
For \(\delta\in(0,1/2)\), set
\[
H_k(\delta):=\sqrt{\log a_k}+\sqrt{s_0\log(d+1)}
+\sqrt{\log(1/\delta)}.
\]

\begin{lemma}
\label{lem:kernel-stochastic}
There are constants \(B_{\mathrm{emp}},C_{\mathrm{emp}}>0\), depending only on the primitive regularity bounds and the kernel, such that the following holds.
On \(\mathcal G_k\), under Assumption~\ref{ass:kernel}, let \(\delta\in(0,1/2)\), \(b_k\le1\), \(W_xR_{1,k}\le b_k\), and \(nb_kq_k\ge B_{\mathrm{emp}}\{\log n+\log(1/b_k)+
\log\{2(d+1)\}+\log(8/\delta)\}\).
For conditionally almost every exploration-context path, the following bound holds with probability at least \(1-\delta\) under the conditional law above, simultaneously for \(j=0,1,2\),
\begin{equation}
\max_{v\in\{h_k^{(j)},f_k^{(j)}\}}
\sup_{u\in I_z,\,\boldsymbol\theta\in\Theta_k}
|v(u,\boldsymbol\theta)-\mathbb Ev(u,\boldsymbol\theta)|
\le C_{\mathrm{emp}}\sqrt{\frac{q_k}{nb_k^{2j+1}}}H_k(\delta).
\label{eq:kernel-stochastic-unified}
\end{equation}
\end{lemma}

\begin{remark}
The supremum in Lemma~\ref{lem:kernel-stochastic} is taken over the deterministic region \(\Theta_k\) before the random estimator \(\widehat{\boldsymbol\theta}_k\) is substituted.
The same exploration block can therefore be used for the Lasso and kernel estimators without an independence argument.
Exact sparsity of the Lasso error is not required: the proof uses only its \(\ell_1\) bound, coordinatewise bounded features, and \(W_xR_{1,k}\le b_k\), which keeps the generated residuals within one kernel bandwidth.
\end{remark}

We now bound the kernel ratio and its derivatives.
For \(n=|I_k|\) and \(\delta\in(0,1/2)\), let
\[
E_{0,k}:=b_k^m+\frac{H_k(\delta)}{\sqrt{nb_kq_k}}+R_{2,k},\qquad
E_{j,k}:=b_k^{m-1}+\frac{H_k(\delta)}{\sqrt{nb_k^{2j+1}q_k}}+R_{2,k},
\quad j=1,2.
\]
Suppose that
\begin{equation}
\begin{gathered}
n\ge2,\quad 0<b_k\le1,\quad
b_k+W_xR_{1,k}\le\operatorname{dist}(J,\partial\supp(z_1))/2,\\
W_xR_{1,k}\le b_k,\quad
\lambda_k\ge8W_x(B+W_xW_\theta)
\sqrt{\frac{\log(d+1)+\log(1/\delta)}n}
\end{gathered}
\label{eq:finite-sample-base-a}
\end{equation}
Let \(B_F\ge1\) be a fixed constant depending only on \(C_{\mathrm{des}},C_{\mathrm{bias}},C_{\mathrm{emp}}\) and the fixed density and kernel bounds.
The corresponding sample-size and denominator conditions are
\begin{align}
nb_kq_k&\ge B_{\mathrm{emp}}
\left\{\log n+\log(1/b_k)+\log\{2(d+1)\}+\log(8/\delta)\right\},\notag\\
n&\ge\frac{2048W_x^4s_0^2}{\lambda_{\min}^2}
\{2\log(d+1)+\log(2/\delta)\},
\label{eq:finite-sample-base-b}\\
B_F\frac{H_k(\delta)}{\sqrt{nb_kq_k}}&\le\frac12,\qquad
B_F\left\{b_k^{m-1}+
\frac{H_k(\delta)}{\sqrt{nb_k^3q_k}}\right\}\le1.
\label{eq:finite-sample-denominator}
\end{align}
The first line is the effective-sample condition, the second controls the restricted-eigenvalue event, and the final line keeps the denominator and the quotient map in a fixed regular region.
Write \(\widehat F_k(u,\boldsymbol\theta)=1-h_k(u,\boldsymbol\theta)/f_k(u,\boldsymbol\theta)\), and let \(\widehat F_k^{(j)}(u,\boldsymbol\theta)\) denote its \(j\)th derivative in \(u\).

Under Assumptions~\ref{ass:bounded-sparse}, \ref{ass:covariance}, \ref{ass:noise-regularity}, and \ref{ass:kernel}, suppose \eqref{eq:finite-sample-base-a}--\eqref{eq:finite-sample-denominator} hold.
Retain the conditional law above.
On the intersection of \(\mathcal E_{k,\mathrm{RE}}\) and the common kernel event, the bias bounds above and Lemma~\ref{lem:kernel-stochastic} give, uniformly on \(J\times\Theta_k\),
\begin{equation*}
\left|\frac{h_k^{(j)}-h_{\boldsymbol\theta}^{(j)}}{q_k}\right|
+\left|\frac{f_k^{(j)}-f_{\boldsymbol\theta}^{(j)}}{q_k}\right|
\le CE_{j,k},\qquad j=0,1,2.
\end{equation*}
Here \(f_{\boldsymbol\theta}=q_k\) and \(f_{\boldsymbol\theta}^{(j)}=0\) for \(j=1,2\) by \eqref{eq:local-residual-density-identity}.
The denominator condition therefore gives \(f_k/q_k\ge1/2\).

Write \(\widehat r_k(u,\boldsymbol\theta)
=h_k(u,\boldsymbol\theta)/f_k(u,\boldsymbol\theta)\).
For \(j=0,1\), use the usual quotient identities.
For the second derivative, since \(f_{\boldsymbol\theta}=q_k\) and \(f_{\boldsymbol\theta}'=f_{\boldsymbol\theta}''=0\),
\[
\widehat r_k''-r_{\boldsymbol\theta}''
=
\frac{h_k''-h_{\boldsymbol\theta}''}{f_k}
+h_{\boldsymbol\theta}''
\left(\frac1{f_k}-\frac1{q_k}\right)
-\frac{h_kf_k''+2h_k'f_k'}{f_k^2}
+\frac{2h_k(f_k')^2}{f_k^3}.
\]
The denominator conditions and the preceding bias and stochastic bounds give, uniformly on \(J\times\Theta_k\), \(f_k/q_k\ge1/2\) and \((|h_k|+|h_k'|+|h_{\boldsymbol\theta}''|+|f_k'|)/q_k\le C\).
In particular, \((f_k'/q_k)^2\le C|f_k'|/q_k\).
Applying the preceding bounds to the quotient identities and using \(E_{0,k}\le E_{1,k}\le E_{2,k}\) therefore gives
\[
\sup_{J\times\Theta_k}
|\widehat r_k^{(j)}-r_{\boldsymbol\theta}^{(j)}|
\le CE_{j,k},
\qquad j=0,1,2.
\]

It remains to replace the localized population target by the true distribution.
Equation~\eqref{eq:derived-conditional-response-stability} and \(\sup_{\boldsymbol\theta\in\Theta_k}\|\boldsymbol\theta-\boldsymbol\theta_0\|_2\le R_{2,k}\) give
\[
\sup_{J\times\Theta_k}
|r_{\boldsymbol\theta}^{(j)}-r_{\boldsymbol\theta_0}^{(j)}|
\le CR_{2,k},\qquad j=0,1,2.
\]
Since \(r_{\boldsymbol\theta_0}=1-F\), substitution into the quotient bounds gives, for a constant \(C_F>0\) depending only on \(C_{\mathrm{des}},C_{\mathrm{bias}},C_{\mathrm{emp}}\) and the fixed density and kernel bounds,
\begin{align}
\sup_{(u,\boldsymbol\theta)\in J\times\Theta_k}
|\widehat F_k(u,\boldsymbol\theta)-F(u)|
&\le C_F E_{0,k},
\label{eq:proof-5-6-prime}\\
\sup_{(u,\boldsymbol\theta)\in J\times\Theta_k}
|\widehat F_k^{(1)}(u,\boldsymbol\theta)-F'(u)|
&\le C_F E_{1,k},
\label{eq:proof-5-7-prime}\\
\sup_{(u,\boldsymbol\theta)\in J\times\Theta_k}
|\widehat F_k^{(2)}(u,\boldsymbol\theta)-F''(u)|
&\le C_F E_{2,k}.
\label{eq:Fsecond-uniform-rate}
\end{align}
After integrating over the exploration-context path, the restricted-eigenvalue and kernel events hold jointly with probability at least \(1-\varepsilon_{\mathrm{RE}}(n)-\delta\), conditional on \(\mathcal H_{k-1}\), by a union bound;
no independence between these events is required.

Equation~\eqref{eq:main-residual-density} follows directly from conditional uniform exploration and the width bounds.
Use \eqref{eq:raw-lasso-rate} and \eqref{eq:proof-5-6-prime}--\eqref{eq:Fsecond-uniform-rate} with \((n,\delta)=(a_k,\delta_k)\) and the fixed interval \(J\).
The bandwidth identity \(a_kb_k^{2m+3}=\omega_k\), the bounds \(\omega_{\min}\le\omega_k\le1\), and \(q_k\ge(B\omega_k)^{-1}\) give
\begin{equation}
 b_k^{m-1}=\left(\frac{\omega_k}{a_kb_k^5}\right)^{1/2},
 \qquad
 a_kb_k^{2j+1}q_k\ge\frac{a_kb_k^5}{B},
 \quad j=0,1,2.
\label{eq:main-to-kernel-conditions}
\end{equation}
Also, \(\log a_k+\log(1/b_k)+\log(1/\delta_k)\le C\Lambda_{\ell_k}\).
Choose \(B_2\) sufficiently large, depending also on the fixed interval \(J\).
The condition \(a_kb_k^5\ge B_2\Lambda_{\ell_k}\) verifies the effective-sample and denominator requirements in \eqref{eq:finite-sample-base-b} and \eqref{eq:finite-sample-denominator}.
Since \(b_k^{2m-2}=\omega_k/(a_kb_k^5)\) and \(\omega_k\le1\), it also ensures \(b_k\le\operatorname{dist}(J,\partial\supp(z_1))/4\).

It remains to localize the generated residuals and verify restricted curvature.
The penalty choice and the definition of \(R_{1,k}\) imply
\begin{equation}
 \frac{W_xR_{1,k}}{b_k}
 \le C s_0
 \left\{\frac{1+\log\ell_k+\log(d+1)}{a_kb_k^2}\right\}^{1/2}.
\label{eq:main-to-residual-localization}
\end{equation}
Thus \(a_kb_k^2\ge B_2s_0^2
\{1+\log\ell_k+\log(d+1)\}\) ensures \(W_xR_{1,k}\le b_k\).
Together with the preceding bandwidth bound, this gives \(b_k+W_xR_{1,k}\le\operatorname{dist}(J,\partial\supp(z_1))/2\), as required in \eqref{eq:finite-sample-base-a}.
Since \(b_k\le1\), the same condition gives the sample size in \eqref{eq:finite-sample-base-b} and hence \(\varepsilon_{\mathrm{RE}}(a_k)\le\delta_k\).
The assumed penalty condition supplies the remaining requirement in \eqref{eq:finite-sample-base-a}.
Finally,
\[
R_{2,k}=\frac{3\lambda_k\sqrt{s_0}}{\lambda_{\min}},
\qquad
H_k(\delta_k)^2
\le C\{1+\log\ell_k+s_0\log(d+1)\}=C\Lambda_{\ell_k}.
\]
On the Lasso event, \eqref{eq:raw-lasso-rate} gives \(\widehat{\boldsymbol\theta}_k\in\Theta_k\).
We may therefore evaluate \eqref{eq:proof-5-6-prime}--\eqref{eq:Fsecond-uniform-rate} at \(\widehat{\boldsymbol\theta}_k\).
Taking \(C_2\) to dominate the fixed multiplier \(C_F\) and the substitution constants, and substituting into \eqref{eq:proof-5-6-prime}, \eqref{eq:proof-5-7-prime}, and \eqref{eq:Fsecond-uniform-rate} gives \eqref{eq:main-F-rate}--\eqref{eq:main-Fsecond-rate}.
The restricted-eigenvalue, score, and kernel events fail with total probability at most \(\varepsilon_{\mathrm{RE}}(a_k)+2\delta_k\le3\delta_k\le
3\ell_k^{-2}\).
This proves Theorem~\ref{lem:noise-function-main}.\hfill\(\square\)

\subsubsection{Proof of Lemma~\ref{lem:pricing-function-main}}

We first identify the population root recovered by the construction in Appendix~\ref{app:pricing-map-evaluation}.
Continuity of \(p\mapsto p\{1-F(p-u)\}\) and compactness of \([0,B]\) make the argmax in \eqref{eq:global-oracle-map} nonempty.
On the interior of the noise support, its derivative is
\[
-F'(p-u)\{u+\phi(p-u)\}.
\]
Because \(\phi\) is strictly increasing, the derivative changes sign at most once.
If \(u>-\delta_z\), some positive price has positive revenue, so the maximizer is unique.
If \(u\le-\delta_z\), all prices yield zero revenue and the zero-revenue convention sets \(g(u)=0\).
The maximum theorem gives continuity for \(u>-\delta_z\).
Moreover, as \(u\downarrow-\delta_z\), any positive-revenue price is smaller than \(u+\delta_z\), and hence \(g(u)\to0\).
Thus \(g:\mathbb R\to[0,B]\) is continuous.

For every \(u\in I_g\), Assumption~\ref{ass:virtual-valuation} places the maximizer in the interior, and its first-order condition gives
\[
g(u)=u+\phi^{-1}(-u).
\]
This is the representation in \eqref{eq:oracle-map-representation}.
Since \(\phi'\ge c_\phi\), the root \(\phi^{-1}(-u)\) is continuous in \(u\).
Its image over the compact interval \(I_g\) is therefore a compact subset of \((-\delta_z,\delta_z)\).
Choose a fixed \(q_F\) such that \(J=J^{q_F}\) contains this image in its interior.
Since all population oracle residuals lie in its interior, \(\phi(\inf J)<\delta_z-B\) and \(\phi(\sup J)>-\delta_z\).
Define
\[
c_{F'}:=\inf_{z\in J}F'(z)>0,\qquad
\Delta_F:=\min\{\delta_z-B-\phi(\inf J),\,
\phi(\sup J)+\delta_z\}>0.
\]
These fixed margins are used only to verify the plug-in inverse.
Let \(B_{\mathrm{inv}}>0\) be a fixed inverse-regularity constant depending only on the margins \(c_{F'},\Delta_F,c_\phi\) and the regularity bounds on \(J\).
The inverse-map conditions are
\begin{align}
B_{\mathrm{inv}}E_{1,k}
&\le\min\left\{\frac{c_{F'}}2,\frac{\Delta_F}2\right\},
\label{eq:finite-sample-inverse}\\
B_{\mathrm{inv}}E_{2,k}
&\le\frac{c_\phi}{2},
\label{eq:finite-sample-derivative}
\end{align}

Under Assumptions~\ref{ass:bounded-sparse}, \ref{ass:covariance}, \ref{ass:noise-regularity}, \ref{ass:virtual-valuation}, and \ref{ass:kernel}, suppose \eqref{eq:finite-sample-base-a}--\eqref{eq:finite-sample-denominator}, \(k\ge q_F\), and \eqref{eq:finite-sample-inverse}--\eqref{eq:finite-sample-derivative} hold.
On \(\mathcal G_k\), work on the common restricted-eigenvalue, score, and kernel events used in \eqref{eq:raw-lasso-rate} and \eqref{eq:proof-5-6-prime}--\eqref{eq:Fsecond-uniform-rate}.
The restricted-eigenvalue and score bounds place \(\widehat{\boldsymbol\theta}_k\) in \(\Theta_k\), so the uniform distribution bounds and the two margin conditions imply
\begin{equation}
\begin{gathered}
\inf_{J\times\Theta_k}f_k\ge q_k/2\ge(2B\omega_k)^{-1},
\qquad
\inf_{J}\widehat F_k^{(1)}\ge c_{F'}/2,
\qquad
\inf_{J}\widehat\phi_k'\ge c_\phi/2,\\
\widehat\phi_k(\inf J)\le\delta_z-B,
\qquad
\widehat\phi_k(\sup J)\ge-\delta_z.
\end{gathered}
\label{eq:joint-plugin-regularity}
\end{equation}
Thus \(J\) satisfies the safeguard whenever \(k\ge q_F\).
The selected interval is then some \(J^q\subseteq J\), so every target \(-u\), \(u\in I_g\), has a unique selected root \(\widehat z(u)=\widehat\phi_k^{-1}(-u)\) in \(J\).

On the region in \eqref{eq:joint-plugin-regularity}, the maps \((a,b)\mapsto(1-a)/b\) and \((a,b,c)\mapsto2+(1-a)c/b^2\) are Lipschitz.
Equations~\eqref{eq:proof-5-6-prime}--\eqref{eq:Fsecond-uniform-rate} therefore give
\[
\sup_{J}|\widehat\phi_k-\phi|\le CE_{1,k},
\qquad
\sup_{J}|\widehat\phi_k'-\phi'|\le CE_{2,k}.
\]
Let \(z(u)=\phi^{-1}(-u)\).
Both \(z(u)\) and \(\widehat z(u)\) lie in \(J\).
Since \(\phi'\ge c_\phi\) there, strong monotonicity yields
\[
c_\phi|\widehat z(u)-z(u)|
\le|\phi\{\widehat z(u)\}-\phi\{z(u)\}|
=|\phi\{\widehat z(u)\}-\widehat\phi_k\{\widehat z(u)\}|
\le CE_{1,k}.
\]
Taking the supremum over \(u\in I_g\) controls the inverse error and, since \(\widehat g_k(u)-g(u)=\widehat z(u)-z(u)\), the pricing-map error.

For the derivative, \(\phi'\) is Lipschitz on \(J\), and \(E_{1,k}\le CE_{2,k}\).
Hence
\[
\sup_{u\in I_g}
|\widehat\phi_k'\{\widehat z(u)\}-\phi'\{z(u)\}|
\le CE_{2,k}.
\]
Finally, \(x\mapsto1-1/x\) is Lipschitz on \([c_\phi/2,\infty)\).
The identities
\[
\widehat g_k'(u)=1-\frac1{\widehat\phi_k'\{\widehat z(u)\}},
\qquad
g'(u)=1-\frac1{\phi'\{z(u)\}}
\]
together with the preceding bounds give
\begin{align}
\sup_{u\in I_g}|\widehat g_k(u)-g(u)|
&\le C_{\mathrm{inv}}E_{1,k},
\label{eq:proof-5-8-prime}\\
\sup_{u\in I_g}|\widehat g_k'(u)-g'(u)|
&\le C_{\mathrm{inv}}E_{2,k},
\label{eq:gprime-lemma-rate}
\end{align}
for a constant \(C_{\mathrm{inv}}>0\) depending only on \(C_F\) and the fixed inverse-regularity margins.
The score and kernel bounds each fail with conditional probability at most \(\delta\), while their shared restricted-eigenvalue event fails with probability at most \(\varepsilon_{\mathrm{RE}}(n)\).
Thus the two bounds hold simultaneously with conditional probability at least \(1-\varepsilon_{\mathrm{RE}}(n)-2\delta\).

We now specialize this transfer to Lemma~\ref{lem:pricing-function-main}.
As in the preceding proof, \(R_{2,k}=3\lambda_k\sqrt{s_0}/\lambda_{\min}\) and \(H_k(\delta_k)^2\le C\Lambda_{\ell_k}\).
Equations \eqref{eq:main-to-kernel-conditions} and \eqref{eq:main-to-residual-localization} verify \eqref{eq:finite-sample-base-a}--\eqref{eq:finite-sample-denominator} and \(\varepsilon_{\mathrm{RE}}(a_k)\le\delta_k\).
They also show that \(E_{2,k}\le C/\sqrt{B_3}\).
Choosing \(B_3\) once so that this upper bound is smaller than the fixed inverse and derivative margins verifies \eqref{eq:finite-sample-inverse}--\eqref{eq:finite-sample-derivative}.

It remains to check that the safeguarded search has reached \(J\).
The index \(q_F\) is fixed, while \(a_kb_k^5\le a_k\le\ell_k\).
By increasing \(B_3\) once more, the sample-size condition fails for each of the finitely many episodes with \(k<q_F\).
Hence the sample-size conditions imply \(k\ge q_F\).
On the Lasso event, \(\widehat{\boldsymbol\theta}_k\in\Theta_k\).
Taking \((n,\delta)=(a_k,\delta_k)\) in \eqref{eq:proof-5-8-prime}--\eqref{eq:gprime-lemma-rate} gives \eqref{eq:main-g-rate} and \eqref{eq:main-gprime-rate}, with \(C_3\) chosen to dominate \(C_{\mathrm{inv}}\).
The restricted-eigenvalue, score, and kernel events hold jointly with conditional probability at least \(1-\varepsilon_{\mathrm{RE}}(a_k)-2\delta_k
\ge1-3\delta_k\ge1-3\ell_k^{-2}\).
This proves Lemma~\ref{lem:pricing-function-main}.\hfill\(\square\)

\section{Proofs of Self-Recovery and Regret Guarantees}
\label{app:main-proofs}
\label{app:regret-proofs}

\subsection{Proofs of Theorem~\ref{lem:confidence-dynamics-main} and Lemma~\ref{lem:confidence-recovery-main}}
\label{sec:global-regeneration}
\label{app:proof-confidence-dynamics-main}
\label{app:proof-confidence-recovery-main}

The estimation results above require \(\mathcal G_k\), under which all exploration valuations lie in their respective intervals almost surely conditional on the pre-episode history.
This requirement creates feedback: a failed update can make the next interval miss some valuations, in which case the constructed response acquires a systematic bias and no longer satisfies the estimating equation used on \(\mathcal G_k\).
The nominal failure probabilities therefore cannot be summed directly.
The proof of Theorem~\ref{lem:confidence-dynamics-main} turns this directional bias into prediction contraction and, once the error is local, confidence re-entry.
Lemma~\ref{lem:confidence-recovery-main} assigns each activated failure a deterministic recovery window and sums its conditional probability against the window length.

\subsubsection{Proof of Theorem~\ref{lem:confidence-dynamics-main}}

Write \(e_k=\|\widehat{\boldsymbol\theta}_k-\boldsymbol\theta_0\|_{\overline{\boldsymbol\Sigma}_k}\) for \(k\ge1\), and set \(e_0=0\).
On \(\mathcal G_j\), the constructed response is unbiased, so Lemma~\ref{lem:sparse-confidence-reentry} implies that the confidence set \(\mathcal C_j\) in \eqref{eq:data-confidence-set} contains \(\boldsymbol\theta_0\) with conditional probability at least \(1-\delta_j\).
On \(\boldsymbol\theta_0\in\mathcal C_j\), the extrema over \(\mathcal B_{\mathrm R}\) and the confidence radius in \eqref{eq:safe-exploration-interval} cover the valuations of buyers in \(I_{j+1}\) before truncation.
Their valuations lie in \([\delta_v,B-\delta_v]\) almost surely, so intersection with \([0,B]\) preserves coverage.
Hence \(\boldsymbol\theta_0\in\mathcal C_j\) implies \(\mathcal G_{j+1}\).

When \(\mathcal G_k\) fails, the constructed response \(Y_t\) instead has conditional mean
\[
\mathbb E_z\!\left[
\min\!\left\{U_k(\mathbf x_{t,\mathrm V}^0),
\max\!\left\{L_k(\mathbf x_{t,\mathrm V}^0),
\boldsymbol\theta_0^\top\mathbf x_t^0+z\right\}\right\}
\right],
\]
where \(\mathbb E_z\) integrates over \(z\sim F\).
Thus an interval miss replaces a valuation outside the interval by the nearest endpoint in the conditional mean of \(Y_t\).

Recall the episode norm \(\|\mathbf u\|_{\overline{\boldsymbol\Sigma}_k}
=(\mathbf u^\top\overline{\boldsymbol\Sigma}_k\mathbf u)^{1/2}\) and the clipping constant
\begin{equation*}
\rho_z=\max\left\{\frac12,
1-\frac{\delta_z}{2W_xW_\theta}\right\}\in[1/2,1)
\end{equation*}
from Assumption~\ref{ass:covariance}, and set \(\rho_{\mathrm g}:=\rho_z(1+\eta_\Sigma)^{1/2}\), which lies in \((0,1)\) because \(\eta_\Sigma<\rho_z^{-2}-1\).
These constants are used only in the recovery argument; \(\rho_{\mathrm g}\) absorbs the change of episode norm \(\|\mathbf u\|_{\overline{\boldsymbol\Sigma}_k}
\le(1+\eta_\Sigma)^{1/2}\|\mathbf u\|_{\overline{\boldsymbol\Sigma}_{k-1}}\) between consecutive episodes, by \eqref{eq:conditional-second-moment}.

Fix any history before episode \(k\).
For an adapted sequence \(Z_t\), write
\[
\overline P_k Z:=\frac1{|I_k|}\sum_{t\in I_k}
\mathbb E[Z_t\mid\mathcal F_{t-1}].
\]
Define the missed-interval bias \(\beta_k(\mathbf x)\) through
\[
\mathbb E[Y_t\mid \mathbf x_t^0=\mathbf x,\mathcal F_{t-1}]
=\mathbf x^\top\boldsymbol\theta_0+\beta_k(\mathbf x),
\]
and define its conditional prediction norm and the distance from the truth to the preceding confidence set by
\[
\sigma_k^2:=\overline P_k\{\beta_k(\mathbf x_t^0)^2\},
\qquad
d_{\Sigma,k-1}=\inf_{\boldsymbol\vartheta\in\mathcal C_{k-1}}
\|\boldsymbol\vartheta-\boldsymbol\theta_0\|_{\overline{\boldsymbol\Sigma}_k} .
\]
For \(r>0\), let \(\rho_F(r)=\max\{1-F(\delta_z-r),F(-\delta_z+r)\}\).
This is the larger of the two noise-tail probabilities within distance \(r\) of a support endpoint, and \(\rho_F(r)\downarrow0\) as \(r\downarrow0\).
The biased conditional mean need not have \(\boldsymbol\theta_0\) as its least-squares target.
We therefore introduce its conditional population projection onto the linear predictor class:
\[
M_k(\boldsymbol\theta):=\overline P_k
\{Y_t-\mathbf x_t^{0\top}\boldsymbol\theta\}^2,
\qquad
\boldsymbol\theta_k^{\mathrm{proj}}
:=\arg\min_{\boldsymbol\theta\in\Theta}M_k(\boldsymbol\theta).
\]
Positive definiteness of \(\overline{\boldsymbol\Sigma}_k\) makes this projection unique.
In what follows, statements about true feature and noise realizations are understood almost surely under their conditional law.
The clipping argument applies to every pre-episode history, without assuming \(\mathcal G_k\).

Choose \(\boldsymbol\vartheta\in\mathcal C_{k-1}\) attaining \(d_{\Sigma,k-1}\), put \(\mathbf h=\boldsymbol\theta_0-\boldsymbol\vartheta\) and \(a(\mathbf x)=\mathbf x^\top\mathbf h\), and fix a realization \((\mathbf x,z)\) of the true features and noise.
With \(\mathbf x=(\mathbf x_{\mathrm V}^\top,(\mathbf x_{\mathrm R}^0)^\top,1)^\top\), let \(v=\boldsymbol\theta_0^\top\mathbf x+z\) and define
\[
r_{\mathbf h}(\mathbf x,z):=\min\{U_k(\mathbf x_{\mathrm V}),
\max\{L_k(\mathbf x_{\mathrm V}),v\}\}-v.
\]
If the global interval is used at \(\mathbf x_{\mathrm V}\), then \(v\in[0,B]\) and \(r_{\mathbf h}(\mathbf x,z)=0\).
Otherwise, put \(a=\mathbf x^\top\mathbf h\).
Before intersection with \([0,B]\), the interval determined by the extrema over \(\mathcal B_{\mathrm R}\) and the confidence radius in \eqref{eq:safe-exploration-interval} contains \(\boldsymbol\vartheta^\top\mathbf x+z'\) for every \(z'\in[-\delta_z,\delta_z]\).
Because \(v\in[\delta_v,B-\delta_v]\), intersecting this interval with \([0,B]\) does not itself exclude \(v\).
If \(a\ge0\), only the upper endpoint can bind, and
\begin{equation*}
0\ge r_{\mathbf h}(\mathbf x,z)\ge-\max\{0,a+z-\delta_z\}.
\end{equation*}
For the convex function \(\psi_a(z)=\max\{0,a+z-\delta_z\}\), the chord joining its values at the support endpoints gives
\[
\psi_a(z)\le
\frac{\delta_z-z}{2\delta_z}\max\{0,a-2\delta_z\}
+\frac{\delta_z+z}{2\delta_z}a.
\]
Since \(\mathbb E_z z=0\),
\[
\mathbb E_z\psi_a(z)
\le\frac{a+\max\{0,a-2\delta_z\}}{2}
\le\rho_z a.
\]
The preceding upper bound equals \(a/2\) when \(a\le2\delta_z\), and \(a-\delta_z\) otherwise.
In the latter case, \(|a|\le W_x\|\mathbf h\|_1\le2W_xW_\theta\) gives \(a-\delta_z\le\{1-\delta_z/(2W_xW_\theta)\}a\).

For the local contraction, if \(a\le r\), then
\[
\psi_a(z)\le a\,\mathbb I\{z>\delta_z-a\}
\le a\,\mathbb I\{z>\delta_z-r\}.
\]
The case \(a<0\) is symmetric, with the lower support tail.
Conditional uniform pricing gives \(\mathbb E[Y_t\mid \mathbf x,z,\mathcal F_{t-1}]
=\min\{U_k(\mathbf x_{\mathrm V}),\max\{L_k(\mathbf x_{\mathrm V}),v\}\}\).
Using \(\mathbb E_z z=0\), the definition above gives \(\beta_k(\mathbf x)=\mathbb E_z[r_{\mathbf h}(\mathbf x,z)\mid \mathbf x,\mathcal F_{t-1}]\).
The preceding bounds therefore give
\begin{equation}
a(\mathbf x)\beta_k(\mathbf x)\le0,\qquad |\beta_k(\mathbf x)|\le\rho_z|a(\mathbf x)|.
\label{eq:pointwise-clipping-contraction}
\end{equation}
If \(|a(\mathbf x)|\le r\), the second bound sharpens to
\begin{equation}
|\beta_k(\mathbf x)|\le\rho_F(r)|a(\mathbf x)|.
\label{eq:local-tail-contraction}
\end{equation}
Moreover, \(\rho_F(r)\downarrow0\) as \(r\downarrow0\) because the noise is supported on \((-\delta_z,\delta_z)\).
During global exploration, bounded support gives \(r_{\mathbf h}\equiv0\), and hence \(\beta_k\equiv0\).

For a \(\overline{\boldsymbol\Sigma}_k\)-nearest point \(\boldsymbol\vartheta\in\mathcal C_{k-1}\), taking conditional \(L_2\) norms in \eqref{eq:pointwise-clipping-contraction} and \eqref{eq:local-tail-contraction}, and using
\[
\overline P_k\{a(\mathbf x_t^0)^2\}
=(\boldsymbol\theta_0-\boldsymbol\vartheta)^\top\overline{\boldsymbol\Sigma}_k
(\boldsymbol\theta_0-\boldsymbol\vartheta),
\]
gives the first bound below.
If \(|a(\mathbf x_t^0)|\le r\) almost surely for every \(t\in I_k\), the same argument gives the second:
\begin{equation}
\sigma_k\le\rho_zd_{\Sigma,k-1},\qquad
\sigma_k\le\rho_F(r)d_{\Sigma,k-1}.
\label{eq:clipping-L2-contraction}
\end{equation}

First-order optimality over the convex set \(\Theta\), using \(\boldsymbol\theta_0\in\Theta\), and Cauchy--Schwarz give
\[
\begin{aligned}
\|\boldsymbol\theta_k^{\mathrm{proj}}-\boldsymbol\theta_0\|_{\overline{\boldsymbol\Sigma}_k}^2
&\le
\left\{\overline P_k\!\left[
\mathbf x_t^0\beta_k(\mathbf x_t^0)
\right]\right\}^{\top}
(\boldsymbol\theta_k^{\mathrm{proj}}-\boldsymbol\theta_0)\\
&\le\sigma_k
\|\boldsymbol\theta_k^{\mathrm{proj}}-\boldsymbol\theta_0\|_{\overline{\boldsymbol\Sigma}_k}.
\end{aligned}
\]
Together with \eqref{eq:clipping-L2-contraction}, \(\widehat{\boldsymbol\theta}_{k-1}\in\mathcal C_{k-1}\), and Assumption~\ref{ass:covariance}, this yields
\begin{equation}
\|\boldsymbol\theta_k^{\mathrm{proj}}-\boldsymbol\theta_0\|_{\overline{\boldsymbol\Sigma}_k}
\le\sigma_k\le\rho_z\|\widehat{\boldsymbol\theta}_{k-1}-\boldsymbol\theta_0\|_{\overline{\boldsymbol\Sigma}_k}
\le\rho_{\mathrm g}e_{k-1}.
\label{eq:population-global-contraction}
\end{equation}
Under Assumptions~\ref{ass:bounded-sparse} and \ref{ass:covariance} and the penalty lower bound in Theorem~\ref{lem:sparse-valuation-main}, write \(n=a_k\) and define the empirical loss by
\[
\widehat M_k(\boldsymbol\theta):=\frac1n\sum_{t\in I_k}
(Y_t-\mathbf x_t^{0\top}\boldsymbol\theta)^2.
\]
Set
\[
\begin{aligned}
D_k&:=\|\widehat{\boldsymbol\Sigma}_k-\overline{\boldsymbol\Sigma}_k\|_{\max},\\
Z_k&:=\left\|
\frac1n\sum_{t\in I_k}
\left[\mathbf x_t^0Y_t-
\mathbb E(\mathbf x_t^0Y_t\mid\mathcal F_{t-1})\right]
\right\|_\infty.
\end{aligned}
\]
Here \(D_k\) controls the empirical quadratic term and \(Z_k\) controls the linear response term.
Their coordinate summands are bounded martingale differences: \(\boldsymbol\Sigma_t\) and \(\mathbb E(\mathbf x_t^0Y_t\mid\mathcal F_{t-1})\) are \(\mathcal F_{t-1}\)-measurable,
and features and responses are bounded by Assumption~\ref{ass:bounded-sparse}.
Coordinatewise conditional Hoeffding--Azuma inequalities and a union bound over at most \((d+1)^2\) coordinates give, with conditional probability at least \(1-\delta_k\),
\begin{equation}
D_k+Z_k
\le C\left\{\frac{2\log(d+1)+\log(4/\delta_k)}{a_k}\right\}^{1/2}.
\label{eq:global-uniform-deviation}
\end{equation}
Uniformly over \(\boldsymbol\theta\in\Theta\),
\begin{equation}
\left|
[\widehat M_k(\boldsymbol\theta)-\widehat M_k(\boldsymbol\theta_k^{\mathrm{proj}})]
-[M_k(\boldsymbol\theta)-M_k(\boldsymbol\theta_k^{\mathrm{proj}})]
\right|
\le4W_\theta^2D_k+4W_\theta Z_k.
\label{eq:global-loss-oscillation}
\end{equation}
Because \(\Theta\subseteq\{\boldsymbol\theta:\|\boldsymbol\theta\|_1\le W_\theta\}\), the penalty difference between \(\widehat{\boldsymbol\theta}_k\) and \(\boldsymbol\theta_k^{\mathrm{proj}}\) is at most \(2W_\theta\lambda_k\).
Empirical penalized optimality and \eqref{eq:global-uniform-deviation}--\eqref{eq:global-loss-oscillation} therefore give
\[
 M_k(\widehat{\boldsymbol\theta}_k)-M_k(\boldsymbol\theta_k^{\mathrm{proj}})
\le C\left[
\lambda_k+
\left\{\frac{2\log(d+1)+\log(4/\delta_k)}{a_k}\right\}^{1/2}
\right]
\le C\lambda_k.
\]
The last inequality is exactly where the penalty lower bound is used.
The first-order optimality condition for the predictable minimizer over the convex set \(\Theta\), together with \(\overline P_k(\mathbf x_t^0\mathbf x_t^{0\top})=\overline{\boldsymbol\Sigma}_k\), gives
\[
M_k(\widehat{\boldsymbol\theta}_k)-M_k(\boldsymbol\theta_k^{\mathrm{proj}})
\ge\|\widehat{\boldsymbol\theta}_k-\boldsymbol\theta_k^{\mathrm{proj}}\|_{\overline{\boldsymbol\Sigma}_k}^2.
\]
Taking square roots, then applying \eqref{eq:population-global-contraction} and the triangle inequality, gives a constant \(C_{\mathrm{hist}}>0\) such that, for every episode \(k\) and every history,
\begin{equation}
\mathbb P(e_k\le\rho_{\mathrm g} e_{k-1}+C_{\mathrm{hist}}\sqrt{\lambda_k}
\mid\mathcal H_{k-1})\ge1-\delta_k.
\label{eq:history-uniform-contraction}
\end{equation}
The term \(\rho_{\mathrm g} e_{k-1}\) is absent during global exploration.
The constant \(C_{\mathrm{hist}}\) absorbs only the fixed concentration and penalty multipliers.

\begin{lemma}
\label{lem:sparse-confidence-reentry}
Under Assumptions~\ref{ass:bounded-sparse} and \ref{ass:covariance}, suppose the SALE penalty satisfies the lower bound in Theorem~\ref{lem:sparse-valuation-main} and \(C_\theta\ge3/\lambda_{\min}\).
There are fixed constants \(C_\Sigma,C_{\mathrm{mis}},c_{\mathrm{ent}}>0\) such that, for every episode \(k\) and every history,
with conditional probability at least \(1-\delta_k\),
\begin{align}
e_k&\le2\sigma_k+C_\Sigma\sqrt{s_0}\lambda_k,
\label{eq:misspecified-prediction-bound}\\
\|\widehat{\boldsymbol\theta}_k-\boldsymbol\theta_0\|_1
&\le C_{\mathrm{mis}}\left(s_0\lambda_k+\frac{\sigma_k^2}{\lambda_k}\right).\notag
\end{align}
Moreover, on the same event, whenever \(\sigma_k\le c_{\mathrm{ent}}\sqrt{s_0}\lambda_k\),
\begin{equation}
\boldsymbol\theta_0\in\mathcal C_k.
\label{eq:confidence-reentry-condition}
\end{equation}
These conclusions do not require \(\mathcal G_k\).
\end{lemma}

Using Lemma~\ref{lem:sparse-confidence-reentry}, whose proof is given in Appendix~\ref{app:proof-sparse-confidence-reentry}, we calibrate the local part of the confidence dynamics.
For this proof and the arguments that follow, write \(\alpha_m=(2m+3)/(4m+1)\), and define the exploration length under the global-width benchmark by
\begin{equation}
a_k^{\mathrm{ref}}:=\min\left\{\ell_k,
\left\lceil c_a(\ell_k\Lambda_{\ell_k})^{\alpha_m}\right\rceil\right\}.
\label{eq:reference-exploration-schedule}
\end{equation}
Since \(\omega_k\in[\omega_{\min},1]\), the realized exploration length satisfies
\begin{equation}
a_k^{\mathrm{ref}}\le a_k\le C_\omega a_k^{\mathrm{ref}},
\qquad
C_\omega:=1+\omega_{\min}^{-(2m+8)/(4m+1)},
\label{eq:sale-schedule-comparability}
\end{equation}
The doubling schedule also gives \(C_{\mathrm{gr}}:=\sup_{k\ge2}a_k^{\mathrm{ref}}/
a_{k-1}^{\mathrm{ref}}<\infty\).

By \eqref{eq:sale-schedule-comparability}, the deterministic reference sequence has growth factor \(C_{\mathrm{gr}}\), and hence, pathwise,
\begin{equation}
\lambda_{i+1}\ge r_\lambda\lambda_i,
\qquad
r_\lambda:=(C_\omega C_{\mathrm{gr}})^{-1/2}>0.
\label{eq:lambda-ratio-lower-bound}
\end{equation}
Indeed, \(1+\log\ell_i+\log(d+1)\) is nondecreasing, whereas \(a_{i+1}/a_i\le C_\omega C_{\mathrm{gr}}\).
Put
\[
\rho_{\mathrm l}:=\frac{r_\lambda}{4},
\qquad
U_0:=1+\frac{4C_\Sigma}{3}.
\]
Because \(\rho_F(r)\downarrow0\), choose \(r_F>0\) such that
\begin{equation*}
2(1+\eta_\Sigma)^{1/2}\rho_F(r_F)\le \rho_{\mathrm l},
\qquad
\frac{(1+\eta_\Sigma)^{1/2}\rho_F(r_F)U_0}{r_\lambda}\le c_{\mathrm{ent}},
\end{equation*}
and define
\[
d_0:=\frac{r_F\sqrt{\lambda_{\min}}}{(1+\eta_\Sigma)^{1/2}W_x\sqrt{d+1}}.
\]
Thus \(d_0=c_d(d+1)^{-1/2}\) for a fixed \(c_d>0\).

If \(e_{i-1}\le d_0\), a \(\overline{\boldsymbol\Sigma}_i\)-nearest \(\boldsymbol\vartheta\in\mathcal C_{i-1}\) satisfies \(\|\boldsymbol\theta_0-\boldsymbol\vartheta\|_{\overline{\boldsymbol\Sigma}_i}
\le\|\boldsymbol\theta_0-\widehat{\boldsymbol\theta}_{i-1}\|_{\overline{\boldsymbol\Sigma}_i}
\le(1+\eta_\Sigma)^{1/2}e_{i-1}\), and hence, almost surely for every \(t\in I_i\),
\[
|\mathbf x_t^{0\top}(\boldsymbol\theta_0-\boldsymbol\vartheta)|
\le\frac{(1+\eta_\Sigma)^{1/2}W_x\sqrt{d+1}}{\sqrt{\lambda_{\min}}}e_{i-1}
\le r_F.
\]
Equations~\eqref{eq:clipping-L2-contraction} and \eqref{eq:misspecified-prediction-bound}, with \(\sigma_i\le\rho_F(r_F)(1+\eta_\Sigma)^{1/2}e_{i-1}\), therefore give
\begin{equation}
e_i\le \rho_{\mathrm l}e_{i-1}+C_\Sigma\sqrt{s_0}\lambda_i.
\label{eq:local-recovery-recurrence}
\end{equation}
On the same misspecified-Lasso event, if also \(e_{i-1}\le U_0\sqrt{s_0}\lambda_{i-1}\), then
\[
\sigma_i\le(1+\eta_\Sigma)^{1/2}\rho_F(r_F)e_{i-1}
\le(1+\eta_\Sigma)^{1/2}\rho_F(r_F)U_0\sqrt{s_0}\lambda_{i-1}
\le c_{\mathrm{ent}}\sqrt{s_0}\lambda_i,
\]
so \eqref{eq:confidence-reentry-condition} gives \(\boldsymbol\theta_0\in\mathcal C_i\).
We have proved the implication
\begin{equation}
e_{i-1}\le d_0,\qquad
e_{i-1}\le U_0\sqrt{s_0}\lambda_{i-1}
\quad\Longrightarrow\quad
\boldsymbol\theta_0\in\mathcal C_i
\label{eq:local-confidence-reentry}
\end{equation}
on that event.

Set \(c_4=\min\{c_d,U_0\}\).
Since \(d_0=c_d(d+1)^{-1/2}\), the stated bound on \(e_{k-1}\) implies both premises of \eqref{eq:local-confidence-reentry}, and hence \(\boldsymbol\theta_0\in\mathcal C_k\) on the misspecified-Lasso event.
The constant \(c_4\) depends only on the constants in the assumptions, the fixed tuning, and the endpoint-tail behavior of the fixed \(F\), not on \(k,d,s_0\).
The history-uniform and misspecified-Lasso events hold jointly, conditional on the pre-episode history, with probability at least \(1-2\delta_k\ge1-2\ell_k^{-2}\).
This proves Theorem~\ref{lem:confidence-dynamics-main}.\hfill\(\square\)

\subsubsection{Proof of Lemma~\ref{lem:confidence-recovery-main}}

Retain the reference schedule and the constants \(C_\omega,C_{\mathrm{gr}},r_\lambda,\rho_{\mathrm l},U_0\), and \(d_0\) constructed in the preceding proof.
The joint event defined in Section~\ref{sec:regret-guarantees} uses the noise-estimation bounds on the fixed interval \(J=J^{q_F}\) chosen in the proof of Lemma~\ref{lem:pricing-function-main} and is
\begin{equation}
\mathcal J_k:=\left\{
\begin{array}{l}
\boldsymbol\theta_0\in\mathcal C_{k-1},\ \text{and the bounds in}\\
\eqref{eq:main-sparse-estimation},
\eqref{eq:main-F-rate}\text{--}\eqref{eq:main-Fsecond-rate},
\text{ and }\eqref{eq:main-g-rate}\text{--}\eqref{eq:main-gprime-rate}
\text{ hold in episode }k
\end{array}
\right\}.
\label{eq:main-joint-good-event}
\end{equation}

Let \(k_{\mathrm{init}}\) be the first episode \(k\ge2\) for which
\begin{equation}
4C_\theta W_xs_0c_\lambda
\left\{
\frac{1+\log\ell_{k-1}+\log(d+1)}{a_{k-1}^{\mathrm{ref}}}
\right\}^{1/2}+\delta_z<B+W_\theta W_x.
\label{eq:implicit-global-startup}
\end{equation}
This is the localization cutoff: episode \(k_{\mathrm{init}}-1\) is the last episode for which the analysis guarantees the full exploration interval \([0,B]\).
This index is used only in the proof, not by the policy.

Fix a deterministic \(k_\star\) satisfying the conditions of Lemma~\ref{lem:confidence-recovery-main}.
Enlarge \(B_3\), if necessary, so that
\[
4C_\theta W_xc_\lambda
\sqrt{\frac{C_\omega C_{\mathrm{gr}}}{B_3}}
<B+W_\theta W_x-\delta_z.
\]
By \eqref{eq:sale-schedule-comparability}, the definition of \(C_{\mathrm{gr}}\), and \(b_{k_\star}\le1\),
\[
s_0\left\{
\frac{1+\log\ell_{k_\star-1}+\log(d+1)}
{a_{k_\star-1}^{\mathrm{ref}}}
\right\}^{1/2}
\le
\left[
\frac{C_\omega C_{\mathrm{gr}}s_0^2
\{1+\log\ell_{k_\star}+\log(d+1)\}}
{a_{k_\star}b_{k_\star}^2}
\right]^{1/2}
\le\sqrt{\frac{C_\omega C_{\mathrm{gr}}}{B_3}},
\]
where the last inequality uses the sample-size condition at \(k_\star\).
Hence \eqref{eq:implicit-global-startup} holds at \(k_\star\), and therefore \(k_{\mathrm{init}}\le k_\star\).

To verify the interpretation of \(k_{\mathrm{init}}\), note that every \(\widehat{\boldsymbol\theta}_i\in\Theta\) places the fitted extrema in \([-W_\theta W_x,W_\theta W_x]\).
Episode \(1\) uses the full interval.
Inductively, if episode \(i\) uses that interval, then \(\omega_i=1\) and \(a_i=a_i^{\mathrm{ref}}\).
For \(i+1<k_{\mathrm{init}}\), the reverse of \eqref{eq:implicit-global-startup} then makes the actual localization allowance large enough to force \([L_{i+1},U_{i+1}]=[0,B]\).
Hence episodes \(1,\ldots,k_{\mathrm{init}}-1\) all use the full interval.
In particular, \(a_{k_{\mathrm{init}}-1}=a_{k_{\mathrm{init}}-1}^{\mathrm{ref}}\), and the missed-interval bias vanishes in that episode.
This initializes the confidence-valid state used below.

We now calibrate the length of a recovery period.
Let
\[
E_\Theta:=2W_\theta\sqrt{\lambda_{\max}},
\qquad
\kappa_z:=\frac{\log2}{2\log(1/\rho_{\mathrm g})}.
\]
The first constant bounds every \(e_i\).
Define the deterministic penalty level
\[
\lambda_i^{\mathrm{ref}}:=c_\lambda
\left\{\frac{1+\log\ell_i+\log(d+1)}{a_i^{\mathrm{ref}}}\right\}^{1/2}.
\]
Since \(a_i\ge a_i^{\mathrm{ref}}\), one has \(\lambda_i\le\lambda_i^{\mathrm{ref}}\).
Let \(k_c\) be the first episode such that, for every \(i\ge k_c\), the reference schedule is uncapped, \(\lambda_i^{\mathrm{ref}}\) is nonincreasing, and
\begin{equation}
C_{\mathrm{hist}}\sqrt{\lambda_i^{\mathrm{ref}}}
\le\frac{(1-\rho_{\mathrm g})d_0}8,
\qquad
C_\Sigma\sqrt{s_0}\lambda_i^{\mathrm{ref}}
\le\frac{(1-\rho_{\mathrm l})d_0}4
\label{eq:two-stage-cutoff-conditions}
\end{equation}
The two inequalities control the additive terms in the global and local recurrences, respectively.

In the uncapped regime,
\[
\lambda_i^{\mathrm{ref}}
\lesssim \ell_i^{-\alpha_m/2}
\Lambda_{\ell_i}^{(1-\alpha_m)/2}.
\]
Let \(D=d+1\).
Since \(s_0\le D\) and \(\alpha_m>1/2\), at a polynomial threshold \(\ell_i\asymp D^{q_{\mathrm{rec}}}\), where \(q_{\mathrm{rec}}=2/\alpha_m+1\), one has \(\Lambda_{\ell_i}\lesssim D\{1+\log D\}\), and the preceding bound gives
\[
\sqrt{\lambda_i^{\mathrm{ref}}}
\lesssim D^{-(1+2\alpha_m)/4}
\{1+\log D\}^{(1-\alpha_m)/4},
\qquad
\sqrt{s_0}\lambda_i^{\mathrm{ref}}
\lesssim D^{-\alpha_m}
\{1+\log D\}^{(1-\alpha_m)/2}.
\]
Both quantities are a fixed polynomial factor smaller than \(d_0=c_dD^{-1/2}\).
Beyond this threshold, the reference schedule is uncapped and \(\lambda_i^{\mathrm{ref}}\) is nonincreasing.
Increasing its fixed multiplier therefore makes the two inequalities in \eqref{eq:two-stage-cutoff-conditions} hold.
By doubling, there is a fixed recovery constant \(C_{\mathrm{cal}}>0\), depending only on the fixed distribution and tuning constants and independent of \(T,d,s_0\), such that
\begin{equation}
\ell_{k_c}\le C_{\mathrm{cal}}(d+1)^{q_{\mathrm{rec}}},
\qquad q_{\mathrm{rec}}=\frac{2}{\alpha_m}+1.
\label{eq:dimension-recovery-cutoff}
\end{equation}
Set
\[
h_c:=\max\left\{1,\left\lceil
\frac{\log(8E_\Theta/d_0)}{\log(1/\rho_{\mathrm g})}
\right\rceil\right\}.
\]
After increasing \(C_{\mathrm{cal}}\) by a fixed factor if necessary, iterating \eqref{eq:history-uniform-contraction} over \(h_c\) successful episodes beginning at or after \(k_c\),
and using the first inequality in \eqref{eq:two-stage-cutoff-conditions}, gives
\begin{equation}
e_i\le d_0/4\quad\text{at the end of the global stage},
\qquad
2^{h_c}\le C_{\mathrm{cal}}(d+1)^{\kappa_z}.
\label{eq:recovery-episode-multiplier}
\end{equation}

The second inequality in \eqref{eq:two-stage-cutoff-conditions} keeps the subsequent errors inside the region \(e_i\le d_0\): if \(e_{i-1}\le d_0\), then \eqref{eq:local-recovery-recurrence} gives
\[
e_i\le\left\{\rho_{\mathrm l}+\frac{1-\rho_{\mathrm l}}4\right\}d_0<d_0.
\]
Within that region, put \(\bar e_i=e_i/(\sqrt{s_0}\lambda_i)\).
Dividing \eqref{eq:local-recovery-recurrence} by \(\sqrt{s_0}\lambda_i\), and using \eqref{eq:lambda-ratio-lower-bound} and \(\rho_{\mathrm l}=r_\lambda/4\), yields
\begin{equation}
\bar e_i\le\frac14\bar e_{i-1}+C_\Sigma.
\label{eq:normalized-local-recurrence}
\end{equation}
Starting from any local episode \(i_0\), this recursion reaches \(\bar e_{i-1}\le U_0\) after at most \(\lceil\log(\max\{1,\bar e_{i_0}\})/\log4\rceil+1\) successful local episodes.
The next successful episode restores \(\boldsymbol\theta_0\in\mathcal C_i\) by \eqref{eq:local-confidence-reentry}.

Throughout this proof we use the stronger conditional probability calibration established above, with \(\delta_j=(d+1)^{-c_\delta}\ell_j^{-2}\): the two one-episode recovery events fail with total probability at most \(2\delta_j\),
while, from a valid state, the shared restricted-eigenvalue, score, and kernel events fail with total probability at most \(3\delta_j\).
We next introduce the events needed to propagate confidence validity.
For every \(j\ge k_{\mathrm{init}}-1\), define the recovery-success event
\[
\begin{aligned}
\mathcal R_j:=
&\{e_j\le\rho_{\mathrm g} e_{j-1}+C_{\mathrm{hist}}\sqrt{\lambda_j}\}\\
&\cap\left\{e_j\le2\sigma_j+C_\Sigma\sqrt{s_0}\lambda_j,
\ \|\widehat{\boldsymbol\theta}_j-\boldsymbol\theta_0\|_1
\le C_{\mathrm{mis}}\left(s_0\lambda_j+\frac{\sigma_j^2}{\lambda_j}\right)
\right\}\\
&\cap\left(\{\sigma_j>c_{\mathrm{ent}}\sqrt{s_0}\lambda_j\}
\cup\{\boldsymbol\theta_0\in\mathcal C_j\}\right).
\end{aligned}
\]
The first two lines are the history-uniform and misspecified-Lasso bounds; the last line records the confidence re-entry implication.
Thus \(\mathcal R_j\) records the contraction bounds and confidence re-entry established in the proof of Theorem~\ref{lem:confidence-dynamics-main}.
If \(j=1\), the contraction term in the first line is read as zero, as prescribed for global exploration.
Equation~\eqref{eq:history-uniform-contraction} and Lemma~\ref{lem:sparse-confidence-reentry} give, for every pre-episode history,
\begin{equation}
\mathbb P(\mathcal R_j^c\mid\mathcal H_{j-1})\le2\delta_j.
\label{eq:recovery-event-probability}
\end{equation}

For \(j\ge k_{\mathrm{init}}\), define
\begin{equation*}
\mathcal V_j:=\{\boldsymbol\theta_0\in\mathcal C_{j-1}\}.
\end{equation*}
This event means that the preceding confidence set certifies the exploration interval used in episode \(j\).
It is \(\mathcal H_{j-1}\)-measurable and \(\mathcal V_j\subseteq\mathcal G_j\).
In the full-interval episode \(k_{\mathrm{init}}-1\), \(\sigma_{k_{\mathrm{init}}-1}=0\); hence \(\mathcal R_{k_{\mathrm{init}}-1}\) implies \(\boldsymbol\theta_0\in\mathcal C_{k_{\mathrm{init}}-1}\),
or equivalently \(\mathcal V_{k_{\mathrm{init}}}\).
This initializes the valid state without a restricted-eigenvalue event.

For \(j\ge k_\star\), recall that \(\mathcal J_j\) is the joint event on which \(\mathcal V_j\) and all conclusions of Theorems~\ref{lem:sparse-valuation-main}--\ref{lem:noise-function-main} and Lemma~\ref{lem:pricing-function-main} hold.
Since \(\mathcal V_j\subseteq\mathcal G_j\), the assumed penalty and sample-size conditions and the common restricted-eigenvalue, score, and kernel events used in the proofs of those results give
\begin{equation}
\mathbb P(\mathcal V_j\cap\mathcal J_j^c\mid\mathcal H_{j-1})
\le3\delta_j.
\label{eq:valid-estimation-failure}
\end{equation}
The intersection with \(\mathcal V_j\) is essential: it invokes the conditional estimation bounds only when their coverage premise is valid.

We now combine the failures that can interrupt the confidence cycle.
Define the trigger event by
\begin{equation}
\mathcal T_j:=
\begin{cases}
\mathcal R_j^c,&k_{\mathrm{init}}-1\le j<k_\star,\\
\mathcal R_j^c\cup(\mathcal V_j\cap\mathcal J_j^c),&j\ge k_\star.
\end{cases}
\label{eq:gated-trigger-definition}
\end{equation}
Thus \(\mathcal T_j\) records whether episode \(j\) interrupts the confidence-valid cycle and must initiate a recovery period.
Before \(k_\star\), only failure of the recovery-success event is relevant.
From \(k_\star\) onward, a failed statistical update is also a trigger, but only from a valid state.
Equations \eqref{eq:recovery-event-probability} and \eqref{eq:valid-estimation-failure} imply
\begin{equation}
\mathbb P(\mathcal T_j\mid\mathcal H_{j-1})\le5\delta_j
\qquad\text{a.s. for every }j\ge k_{\mathrm{init}}-1.
\label{eq:trigger-conditional-probability}
\end{equation}
Having isolated valid trigger events, we assign a deterministic recovery window to a failure in episode \(j\).
Let \(r_j\) be the first post-trigger episode no earlier than the coarse cutoff, let \(t_j\) be the end of its \(h_c\)-episode coarse stage, and use \(\lambda_i^-\) as a deterministic lower bound on the realized penalty:
\[
\begin{aligned}
r_j&:=\max\{j+1,k_c\},&
t_j&:=r_j+h_c-1,\\
\lambda_i^-
&:=c_\lambda\sqrt{\frac{\log4}{\ell_i}}
\le\lambda_i.
\end{aligned}
\]
The inequality follows from \(a_i\le\ell_i\) and \(1+\log\ell_i+\log(d+1)\ge\log4\).
The number of local episodes needed after \(t_j\) is bounded by
\[
h_j^{\mathrm{loc}}:=
\left\lceil
\frac{\max\{0,\log[d_0/(\sqrt{s_0}\,\lambda_{t_j}^-)]\}}
{\log4}
\right\rceil+2.
\]
The corresponding recovery period is
\[
\mathcal W_j:=\{j,j+1,\ldots,t_j+h_j^{\mathrm{loc}}+2\}.
\]
This deterministic set covers the wait to the coarse cutoff, both recovery stages, confidence re-entry, and the subsequent transition to the joint event.
The covering argument uses valid-state propagation and post-trigger recovery.
For valid-state propagation, suppose \(i\ge k_{\mathrm{init}}\) and \(\mathcal V_i\cap\mathcal T_i^c\) holds.
Because \(\mathcal V_i\subseteq\mathcal G_i\), the missed-interval bias is zero.
The recovery event therefore gives \(\boldsymbol\theta_0\in\mathcal C_i\), and hence \(\mathcal V_{i+1}\).
When \(i\ge k_\star\), the second line of \eqref{eq:gated-trigger-definition} also gives \(\mathcal J_i\).
Thus a nontriggering episode propagates confidence validity and, from \(k_\star\) onward, yields the joint event used in the regret analysis.

For post-trigger recovery, suppose \(\mathcal T_j\) occurs and no later trigger occurs through the end of \(\mathcal W_j\).
Then \(\mathcal R_i\) holds in every subsequent episode of the assigned recovery window.
The coarse calculation \eqref{eq:recovery-episode-multiplier} gives \(e_{t_j}\le d_0/4\).
Since \(\lambda_{t_j}\ge\lambda_{t_j}^-\) and \(e_{t_j}\le d_0/4\), the recurrence \eqref{eq:normalized-local-recurrence}, with the chosen \(h_j^{\mathrm{loc}}\), reduces the normalized error to at most \(U_0\) within the assigned local stage.
Let \(i\) be the first episode in that stage for which \(\bar e_{i-1}\le U_0\).
Since no later trigger occurs, \(\mathcal R_i\) holds, and \eqref{eq:local-confidence-reentry} gives \(\boldsymbol\theta_0\in\mathcal C_i\).
Thus \(\mathcal V_{i+1}\) holds.
Applying valid-state propagation from episode \(i+1\) onward preserves confidence-set coverage and, once the episode index reaches \(k_\star\), yields the joint event.
This propagation continues until the next failure.
The two final episodes in \(\mathcal W_j\) conservatively cover these state transitions.

To obtain the covering, fix \(k\ge k_\star\) outside the union of all realized recovery periods.
If no trigger occurs from episode \(k_{\mathrm{init}}-1\) through \(k\), the full-interval initialization and valid-state propagation give \(\mathcal J_k\).
Otherwise, let \(j\le k\) be the last episode in which \(\mathcal T_j\) occurs.
Since \(k\notin\mathcal W_j\), episode \(k\) follows the recovery sequence verified above.
No later trigger occurs, so valid-state propagation again gives \(\mathcal J_k\).
Hence, pathwise,
\begin{equation}
\mathbb I(\mathcal J_k^c)
\le
\sum_{j\ge k_{\mathrm{init}}-1}
\mathbb I(\mathcal T_j)\mathbb I\{k\in\mathcal W_j\},
\qquad k\ge k_\star.
\label{eq:bad-event-window-cover}
\end{equation}
It remains to bound their expected cost.
Define the total episode length associated with this recovery period by
\[
L_j^{\mathrm{rec}}:=\sum_{k\in\mathcal W_j}\ell_k.
\]
The geometric episode schedule first gives
\begin{equation}
\begin{aligned}
\ell_{r_j}
&\le 2\ell_j+\ell_{k_c}
\le C_{\mathrm{cal}}\{\ell_j+(d+1)^{q_{\mathrm{rec}}}\},\\
\ell_{t_j}
&=2^{h_c-1}\ell_{r_j}
\le C_{\mathrm{cal}}(d+1)^{\kappa_z}
\{\ell_j+(d+1)^{q_{\mathrm{rec}}}\},
\end{aligned}
\label{eq:trigger-coarse-length}
\end{equation}
where the second inequalities use \eqref{eq:dimension-recovery-cutoff} and \eqref{eq:recovery-episode-multiplier}.
Next, by the definition of \(h_j^{\mathrm{loc}}\), the identity \(\lambda_{t_j}^-=c_\lambda\sqrt{\log4/\ell_{t_j}}\), and \(1\le s_0\le d+1\),
\begin{equation}
2^{h_j^{\mathrm{loc}}}
\le8\max\left\{1,
\frac{d_0}{\sqrt{s_0}\,\lambda_{t_j}^-}
\right\}^{1/2}
\le C_{\mathrm{cal}}\ell_{t_j}^{1/4}.
\label{eq:trigger-local-length}
\end{equation}
Finally, the sum of the geometrically increasing episode lengths in \(\mathcal W_j\) is at most twice its last term.
Equations \eqref{eq:trigger-coarse-length}--\eqref{eq:trigger-local-length} therefore give
\begin{equation}
L_j^{\mathrm{rec}}
\le C_{\mathrm{cal}}\left[(d+1)^{\kappa_z}
\{\ell_j+(d+1)^{q_{\mathrm{rec}}}\}\right]^{5/4}.
\label{eq:trigger-recovery-cost}
\end{equation}
Multiplying \eqref{eq:bad-event-window-cover} by \(\ell_k\), summing over \(k\), taking expectations, and using \eqref{eq:trigger-conditional-probability} yields
\[
\begin{aligned}
\mathbb E\left[
\sum_{k\ge k_\star}\ell_k\mathbb I(\mathcal J_k^c)\right]
&\le\sum_{j\ge k_{\mathrm{init}}-1}
\mathbb P(\mathcal T_j)L_j^{\mathrm{rec}}\\
&\le C_{\mathrm{cal}}(d+1)^{5\kappa_z/4-c_\delta}
\sum_{j\ge k_{\mathrm{init}}-1}
\left\{\ell_j^{-3/4}
+(d+1)^{5q_{\mathrm{rec}}/4}\ell_j^{-2}\right\}
\le C_5.
\end{aligned}
\]
Both series are geometric, and \(c_\delta>5(\kappa_z+q_{\mathrm{rec}})/4\).
Hence the conclusion of Lemma~\ref{lem:confidence-recovery-main} holds.
The constant \(C_5\) absorbs the fixed recovery constant \(C_{\mathrm{cal}}\) and is independent of \(k_\star,T,d\), and \(s_0\).
\hfill\(\square\)

\begin{remark}
For the more general choice \(\delta_k=(d+1)^{-c_\delta}\ell_k^{-p}\), \eqref{eq:trigger-recovery-cost} is summable whenever \(p>5/4\) and the dimension exponent is sufficiently large.
Our choice \(p=2\) requires only \(c_\delta>5(\kappa_z+q_{\mathrm{rec}})/4\).
Moreover, \(\log(1/\delta_k)=c_\delta\log(d+1)+2\log\ell_k\lesssim\Lambda_{\ell_k}\), so this choice changes only fixed constants in the estimation bounds and the regret rate.
\end{remark}

To apply the estimation results uniformly over the remaining episodes, let \(k_0\) be the first episode satisfying
\begin{equation}
\ell_{k_0}\ge C_{\mathrm{cut}}\max\left\{
[1+s_0\log(d+1)]^{\frac{2m+3}{2m-2}},
(1+s_0)^{\frac{6m+1}{2m+1}}
[1+\log(d+1)]^{\frac{2m}{2m+1}}
\right\}.
\label{eq:appendix-cutoff-definition}
\end{equation}
This cutoff is used only in the proof.
For every subsequent episode, the required sparse and nonparametric sample-size conditions hold as follows.
The proof of Lemma~\ref{lem:finite-localization} is given in Appendix~\ref{app:proof-finite-localization}.

\begin{lemma}
\label{lem:finite-localization}
Fix \(M\ge1\).
By choosing \(C_{\mathrm{cut}}\) in \eqref{eq:appendix-cutoff-definition} sufficiently large, depending on \(M\) and the fixed model, regularity, and tuning constants but not on \(k,T,d\), or \(s_0\), one has, for every \(k\ge k_0\),
\[
a_kb_k^5\ge M\Lambda_{\ell_k},
\qquad
a_kb_k^2\ge M s_0^2\{1+\log\ell_k+\log(d+1)\}.
\]
For \(M\) sufficiently large relative to the fixed constants in Theorems~\ref{lem:sparse-valuation-main}--\ref{lem:noise-function-main} and Lemma~\ref{lem:pricing-function-main} and the reference-schedule restricted-eigenvalue bound,
their finite-sample conditions hold with \((n,\delta)=(a_k,\delta_k)\), and \(\varepsilon_{\mathrm{RE}}(a_k^{\mathrm{ref}})\le\delta_k\), for every \(k\ge k_0\).
The same choice can ensure \(k_0\ge q_F\).
Moreover,
\begin{equation}
\ell_{k_0}\le C_{\mathrm{loc}}\max\left\{
[1+s_0\log(d+1)]^{\frac{2m+3}{2m-2}},
(1+s_0)^{\frac{6m+1}{2m+1}}
[1+\log(d+1)]^{\frac{2m}{2m+1}}
\right\},
\label{eq:cutoff-size}
\end{equation}
where \(C_{\mathrm{loc}}\) may depend on \(M\) and the fixed model, regularity, and tuning constants, but not on \(T,d\), or \(s_0\).
\end{lemma}

\subsection{Proof of Theorem~\ref{thm:main_hd}}
\label{app:proof-main-hd}

The regret bounds for SALE-K and SALE-U use the estimation guarantees in Appendix~\ref{app:intermediate-proofs} and the recovery results in Appendix~\ref{sec:global-regeneration}.
We first record the deterministic schedule calculation that converts a valid episode envelope into the final horizon rate.
For the regret and schedule calculations below, write
\[
\alpha_m=\frac{2m+3}{4m+1},\qquad
r_m=\frac{2m-2}{2m+3},\qquad
\nu_m=\frac{2m+8}{4m+1},\qquad
\kappa_m=\frac{6m-6}{4m+1}.
\]
Here \(\alpha_m\) is the horizon exponent, \(r_m\) is the balanced nonparametric-error exponent, \(\nu_m\) is the width adjustment in the SALE schedule, and \(\kappa_m\) is the resulting width-gain exponent.

The factor \(\Gamma_k=1+\log(d+1)+\log\ell_k\) is the logarithmic factor in the sparse-estimation error, whereas \(\Lambda_{\ell_k}\) also records the sparsity-weighted dimensional complexity used by the SALE schedule.
Given \(C_{\mathrm{env}}>0\), consider a possibly random nonnegative sequence \((r_k)_{k\ge k_0}\) satisfying the episode envelope
\begin{equation}
r_k\le C_{\mathrm{env}}\rho_A^4\left[
\omega_k^2a_k
+\mathbb I\{a_k<\ell_k\}\ell_k
\left\{\Lambda_{\ell_k}\omega_k^{r_m}a_k^{-r_m}
+\frac{s_0\Gamma_k}{a_k}\right\}\right].
\label{eq:technical-episode-envelope}
\end{equation}

First consider an uncapped episode.
The schedule and its reference version give \(a_k\asymp a_k^{\mathrm{ref}}\omega_k^{-\nu_m}\asymp
(\ell_k\Lambda_{\ell_k})^{\alpha_m}\omega_k^{-\nu_m}\).
The identities
\[
1-\alpha_mr_m=\alpha_m,
\qquad
2-\nu_m=r_m(1+\nu_m)=\kappa_m
\]
show that the exploration and nonparametric terms in \eqref{eq:technical-episode-envelope} are bounded by the right-hand side of \eqref{eq:technical-episode-rate}.
For the sparse term, the same schedule comparison first gives \(\ell_ks_0\Gamma_k/a_k\le
C\ell_ks_0\Gamma_k\omega_k^{\nu_m}/a_k^{\mathrm{ref}}\).
The coefficient of \(\omega_k^{\nu_m}\) satisfies
\[
\frac{\ell_ks_0\Gamma_k/a_k^{\mathrm{ref}}}
{(\ell_k\Lambda_{\ell_k})^{\alpha_m}}
\lesssim
\frac{s_0\Gamma_k}{\ell_k^{2\alpha_m-1}
\Lambda_{\ell_k}^{2\alpha_m}}.
\]
Because \(2\alpha_m-1>0\), the logarithmic growth of \(\Gamma_k\) gives \(\Gamma_k\ell_k^{-(2\alpha_m-1)}\le C\{1+\log(d+1)\}\).
Together with \(\Lambda_{\ell_k}\ge1+s_0\log(d+1)\), this shows that the ratio is uniformly bounded.
Therefore
\[
\ell_k\frac{s_0\Gamma_k}{a_k}
\le C(\ell_k\Lambda_{\ell_k})^{\alpha_m}\omega_k^{\nu_m}
\le C(\ell_k\Lambda_{\ell_k})^{\alpha_m}\omega_k^{\kappa_m},
\]
where the final constant uses the fixed floor \(\omega_k\ge\omega_{\min}\).

If \(a_k=\ell_k\), the exploitation terms vanish and the cap condition gives
\[
\omega_k^2\ell_k
\le C(\ell_k\Lambda_{\ell_k})^{\alpha_m}\omega_k^{2-\nu_m}
=C(\ell_k\Lambda_{\ell_k})^{\alpha_m}\omega_k^{\kappa_m}.
\]
Thus there is a constant \(C_{\mathrm{agg}}>0\), depending only on \(C_{\mathrm{env}}\) and the fixed schedule constants, such that every sequence satisfying \eqref{eq:technical-episode-envelope} obeys
\begin{equation}
r_k\le C_{\mathrm{agg}}\rho_A^4
(\ell_k\Lambda_{\ell_k})^{\alpha_m}\omega_k^{\kappa_m}.
\label{eq:technical-episode-rate}
\end{equation}
Finally, the doubling schedule implies \(\ell_{K_T}\le T+\ell_0-1\), \(\Lambda_{\ell_{K_T}}\le C\Lambda_T\), and
\[
\sum_{k=1}^{K_T}(\ell_k\Lambda_{\ell_k})^{\alpha_m}
\le C(T\Lambda_T)^{\alpha_m}.
\]
Because all terms are nonnegative, we may extend the sum from \(k_0\) back to episode \(1\).
Taking expectations and using \eqref{eq:effective-width-factor} gives, for either policy \(\pi\in\{\mathrm K,\mathrm U\}\),
\begin{equation}
\mathbb E_\pi\sum_{k=k_0}^{K_T}r_k
\le C_{\mathrm{agg}}\rho_A^4(\omega_{\pi,T})^{\kappa_m}
(T\Lambda_T)^{\alpha_m}.
\label{eq:technical-episode-sum}
\end{equation}
The implicit constants depend only on \(C_{\mathrm{env}}\) and the fixed schedule constants and are absorbed by \(C_{\mathrm{agg}}\).

Revenue curvature reduces one-period regret to squared price error.
We bound acquisition and deployment losses episode by episode, use Lemma~\ref{lem:confidence-recovery-main} to control invalid episodes, and apply \eqref{eq:technical-episode-sum} to sum the resulting envelopes.
Throughout, \(C\) denotes a positive constant that may change from line to line and depends only on the fixed model, regularity, and tuning constants, but not on \(\rho_A\) or the realized exploration widths.

For both regret proofs, assume that an optimal report in \eqref{eq:buyer-report-problem} is selected by a fixed measurable rule using only the buyer's true features and past information.
For an exploitation period \(t\), introduce the enlarged analysis filtration
\begin{equation*}
\widetilde{\mathcal F}_{t-1}
=\mathcal F_{t-1}\vee\sigma(\mathbf x_t^0,\mathbf x_t,\mathrm{id}_t).
\end{equation*}
This filtration augments the seller's history by the current true and reported features and the buyer identifier; it is used only for conditioning in the proof.
Under this reporting convention and the noise-independence conditions, \(z_t\) remains independent of \(\widetilde{\mathcal F}_{t-1}\).
The price, the oracle price, and any pre-price matched-pair estimate are \(\widetilde{\mathcal F}_{t-1}\)-measurable, whereas the purchase outcome \(y_t\) is not.

The reference schedule and its comparison with the realized SALE schedule are
defined in \eqref{eq:reference-exploration-schedule}--%
\eqref{eq:sale-schedule-comparability}.
We first decompose regret by episode.
Recall that \(K_T\) is the last episode that begins by period \(T\).
In that episode, intersect the exploration and exploitation sets below with \([T]\).
Since every upper bound below is nonnegative, we may bound this truncated episode by its full scheduled episode.
For each episode \(k\), define its start time and its exploration and exploitation blocks by
\[
S_k:=1+\sum_{j<k}\ell_j,\qquad
I_k=\{S_k,\ldots,S_k+a_k-1\},\qquad
I_k'=\{S_k+a_k,\ldots,S_k+\ell_k-1\}
\]
As proof-only shorthand, write \(\Delta_t:=p_t^\ast\mathbb I(p_t^\ast\le v_t)
-p_t\mathbb I(p_t\le v_t)\) for the realized one-period revenue difference.
Write \(\mathbb E_{\mathrm K}\) for expectation under SALE-K.
Then
\begin{equation*}
\begin{gathered}
L_{1,k}:=\sum_{t\in I_k\cap[T]}\Delta_t,\qquad
L_{2,k}:=\sum_{t\in I_k'\cap[T]}\Delta_t,\\
\Reg_{\mathrm K}(T)=\mathbb E_{\mathrm K}\sum_{k=1}^{K_T}
\bigl(L_{1,k}+L_{2,k}\bigr).
\end{gathered}
\end{equation*}

The schedule is predictable: \(\omega_k\) and \(a_k\) are \(\mathcal H_{k-1}\)-measurable.
Hence the one-episode concentration arguments apply after conditioning on \(\mathcal H_{k-1}\).
The comparability relation \eqref{eq:sale-schedule-comparability} and the floor \(\omega_k\ge\omega_{\min}\) allow the finite-sample conditions to be checked against the deterministic reference schedule.
Apply Lemma~\ref{lem:finite-localization} with \(M\) chosen to dominate
the constants in Theorems~\ref{lem:sparse-valuation-main}--\ref{lem:noise-function-main} and Lemma~\ref{lem:pricing-function-main} and the reference-schedule restricted-eigenvalue bound.
Use the resulting cutoff \(k_0\) throughout this proof.

We next collect the events used in the episode bound.
From episode \(k_0\) onward, the local nonparametric, compact-range, kernel, and sparse restricted-eigenvalue conditions hold, while \eqref{eq:cutoff-size} bounds the preceding episodes.
Let \(\mathcal H_k^+:=\mathcal F_{S_k+a_k-1}\) denote the post-acquisition history in episode \(k\).
Recall \(\mathcal J_k\) from \eqref{eq:main-joint-good-event}, and let \(\mathcal V_k=\{\boldsymbol\theta_0\in\mathcal C_{k-1}\}\) be its pre-acquisition coverage component.

The event \(\mathcal V_k\) is \(\mathcal H_{k-1}\)-measurable and implies \(\mathcal G_k\).
The event \(\mathcal J_k\) is \(\mathcal H_k^+\)-measurable and adds the sparse and nonparametric estimation bounds obtained from the acquisition block.
Thus \(\mathcal J_k\subseteq\mathcal V_k\subseteq\mathcal G_k\): the first event is used for deployment, while \(\mathcal V_k\) is the event available before acquisition begins.
The noise-estimation bounds on \(J\) and the fixed inverse-map margins ensure that \(\widehat g_k\) and \(\widehat g_k'\) are well defined on \(I_g\) whenever \(\mathcal J_k\) holds.
On its complement, the policy still posts a price in \([0,B]\).
Applying Lemma~\ref{lem:confidence-recovery-main} with \(k_\star=k_0\) gives
\begin{equation}
\mathbb E\left[
\sum_{k\ge k_0}\ell_k\mathbb I(\mathcal J_k^c)
\right]\le C_5.
\label{eq:regeneration-bad-event-bound}
\end{equation}
The episodes before \(k_0\) will be bounded at the end of the proof using the uniform one-period regret bound and \eqref{eq:cutoff-size}.

We first bound regret by squared price error.
Recall \(\operatorname{rev}_t\) from \eqref{revenuet}.
The valuation-support condition, \eqref{eq:oracle-map-representation}, and Assumption~\ref{ass:virtual-valuation} place \(p_t^*=g(\boldsymbol\theta_0^\top\mathbf x_t^0)\) and \(p_t^*-\boldsymbol\theta_0^\top\mathbf x_t^0\) in fixed compact interior intervals.
Consequently, \(\operatorname{rev}_t'(p_t^*)=0\), and Assumption~\ref{ass:noise-regularity} uniformly bounds \(\operatorname{rev}_t''(p)=-2F'(p-\boldsymbol\theta_0^\top\mathbf x_t^0)
-pF''(p-\boldsymbol\theta_0^\top\mathbf x_t^0)\) in a fixed neighborhood of \(p_t^*\).
Taylor's theorem applies in that neighborhood.
Outside it, bounded one-period revenue and the fixed positive neighborhood radius give the same bound after increasing the constant.
Thus there is a revenue-curvature constant \(C_{\mathrm{rev}}>0\), depending only on the fixed compact-range and noise-regularity bounds, such that, uniformly over \(p\in[0,B]\),
\begin{equation}
0\le \operatorname{rev}_t(p_t^*)-\operatorname{rev}_t(p)
\le C_{\mathrm{rev}}(p-p_t^*)^2.
\label{eq:quadratic-revenue-reduction}
\end{equation}
Since the current noise is independent of the pre-price information, conditioning first on \(\widetilde{\mathcal F}_{t-1}\) gives
\begin{equation}
\mathbb E[\Delta_t\mid\widetilde{\mathcal F}_{t-1}]
=\operatorname{rev}_t(p_t^*)-\operatorname{rev}_t(p_t)
\le C_{\mathrm{rev}}(p_t-p_t^*)^2.
\label{eq:conditional-regret-reduction}
\end{equation}
Conditional bounds at an earlier information time follow by the tower property.

We next bound acquisition regret.
On \(\mathcal V_k\), the endpoint bounds in \eqref{eq:coverage-residual-margin} and Assumption~\ref{ass:virtual-valuation} imply, almost surely,
\[
p_t^\ast
=\boldsymbol\theta_0^\top\mathbf x_t^0
+\phi^{-1}(-\boldsymbol\theta_0^\top\mathbf x_t^0)
\in[L_k(\mathbf x_{t,\mathrm V}^0),U_k(\mathbf x_{t,\mathrm V}^0)].
\]
Recall that \(W_k(\mathbf x_{\mathrm V})=U_k(\mathbf x_{\mathrm V})-
L_k(\mathbf x_{\mathrm V})\).
Conditional on \(\mathcal H_{k-1}\) and \(\mathbf x_t^0\), direct integration over the uniform exploration price gives
\begin{equation*}
\mathbb E\!\left[(p_t-p_t^\ast)^2
\mid\mathbf x_t^0,\mathcal H_{k-1}\right]
 =\frac{\{p_t^\ast-L_k(\mathbf x_{t,\mathrm V}^0)\}^3
 +\{U_k(\mathbf x_{t,\mathrm V}^0)-p_t^\ast\}^3}
{3W_k(\mathbf x_{t,\mathrm V}^0)}
\le\frac{W_k(\mathbf x_{t,\mathrm V}^0)^2}{3}
\le\frac{B^2\omega_k^2}{3}.
\end{equation*}
Because \(\mathcal V_k\), \(a_k\), and \(\omega_k\) are \(\mathcal H_{k-1}\)-measurable, combining this display with \eqref{eq:quadratic-revenue-reduction} and using the tower property yields
\begin{equation}
\mathbb E_{\mathrm K}
\left[L_{1,k}\mathbb I(\mathcal V_k)\right]
\le
\frac{C_{\mathrm{rev}}B^2}{3}\mathbb E_{\mathrm K}
\left[\omega_k^2a_k\mathbb I(\mathcal V_k)\right].
\label{eq:proof-knownA-explore-regret}
\end{equation}
This calculation does not condition exploration loss on the estimation bounds inside \(\mathcal J_k\), which are generated by the exploration observations themselves.
Their failures enter only in the post-exploration deployment split below.

We then turn to deployment.
Fix \(t\in I_k'\).
Both \(p_t\) and \(p_t^*\) are \(\widetilde{\mathcal F}_{t-1}\)-measurable, so \eqref{eq:conditional-regret-reduction} applies.
Define the strategic correction coefficient by
\[
\chi_0:=\boldsymbol\beta_{\mathrm R,0}^\top
\mathbf A^{-1}\boldsymbol\beta_{\mathrm R,0}.
\]
By \eqref{eq:population-corrected-valuation},
\begin{equation}
\boldsymbol\theta_0^\top\mathbf x_t
+\chi_0g'(\boldsymbol\theta_0^\top\mathbf x_t)
=\boldsymbol\theta_0^\top\mathbf x_t^0,
\qquad
p_t^\ast=g(\boldsymbol\theta_0^\top\mathbf x_t^0).
\label{eq:proof-knownA-oracle-price}
\end{equation}

For \(k\ge k_0\), work on \(\mathcal J_k\) and define
\[
\widehat\chi_k=\widehat{\boldsymbol\beta}_{\mathrm R,k}^\top
\mathbf A^{-1}\widehat{\boldsymbol\beta}_{\mathrm R,k}.
\]
When \(\widehat{\boldsymbol\theta}_k^\top\mathbf x_t\in I_g\), the policy evaluates \(\widehat g_k\) at
\[
\widehat{\boldsymbol\theta}_k^\top\mathbf x_t
+\widehat\chi_k\widehat g_k'
(\widehat{\boldsymbol\theta}_k^\top\mathbf x_t);
\]
otherwise it uses the fallback specified in Appendix~\ref{app:pricing-map-evaluation}.

The valuation-support condition and Assumption~\ref{ass:manipulation-cost} place \(\boldsymbol\theta_0^\top\mathbf x_t^0\) and \(\boldsymbol\theta_0^\top\mathbf x_t\) at a fixed positive distance from \(\partial I_g\).
If \(\widehat{\boldsymbol\theta}_k^\top\mathbf x_t\notin I_g\), this distance gives
\[
\mathbb I\{\widehat{\boldsymbol\theta}_k^\top\mathbf x_t\notin I_g\}
\le C|(
\widehat{\boldsymbol\theta}_k-\boldsymbol\theta_0)^\top\mathbf x_t|^2.
\]
If \(\widehat{\boldsymbol\theta}_k^\top\mathbf x_t\in I_g\) but the corrected argument above lies outside \(I_g\), its exit indicator is bounded by the squared distance between that argument and \(\boldsymbol\theta_0^\top\mathbf x_t^0\).
The squared difference between the fallback price and \(p_t^\ast\) is at most \(B^2\).
On the remaining branch, projection onto \([0,B]\), the uniform plug-in bound, and Lipschitz continuity of \(g\) give
\[
|p_t-p_t^*|\le\|\widehat g_k-g\|_{\infty,I_g}
+C\left|
\widehat{\boldsymbol\theta}_k^\top\mathbf x_t
+\widehat\chi_k\widehat g_k'(\widehat{\boldsymbol\theta}_k^\top\mathbf x_t)
-\boldsymbol\theta_0^\top\mathbf x_t^0
\right|.
\]
Combining the three cases yields
\begin{equation}
(p_t-p_t^\ast)^2
\le C\bigg\{\|\widehat g_k-g\|_{\infty,I_g}^2
+|(\widehat{\boldsymbol\theta}_k-\boldsymbol\theta_0)^\top\mathbf x_t|^2
+\left|
\widehat{\boldsymbol\theta}_k^\top\mathbf x_t
+\widehat\chi_k\widehat g_k'(\widehat{\boldsymbol\theta}_k^\top\mathbf x_t)
-\boldsymbol\theta_0^\top\mathbf x_t^0
\right|^2
\mathbb I\{\widehat{\boldsymbol\theta}_k^\top\mathbf x_t\in I_g\}\bigg\}.
\label{eq:knownA-safeguarded-price}
\end{equation}

To control the correction term, Equation~\eqref{eq:manipulation}, boundedness of \(g'\), and Assumption~\ref{ass:covariance} imply, for every \(\mathcal F_{t-1}\)-measurable vector \(\mathbf h\in\mathbb R^{d+1}\),
\begin{equation}
\mathbb E\bigl[(\mathbf h^\top\mathbf x_t)^2\mid\mathcal F_{t-1}\bigr]
\le
2\mathbf h^\top\boldsymbol\Sigma_t\mathbf h
+C\|\mathbf A^{-1}\boldsymbol\beta_{\mathrm R,0}\|_2^2\|\mathbf h\|_2^2
\le C\rho_A^2\|\mathbf h\|_2^2.
\label{eq:reported-prediction}
\end{equation}
The map \(\mathbf b\mapsto \mathbf b^\top\mathbf A^{-1}\mathbf b\) is Lipschitz on the bounded parameter region, so
\[
|\widehat\chi_k-\chi_0|
\le C\rho_A\|\widehat{\boldsymbol\theta}_k-\boldsymbol\theta_0\|_2,
\qquad |\chi_0|\vee|\widehat\chi_k|\le C\rho_A.
\]

On \(\{\widehat{\boldsymbol\theta}_k^\top\mathbf x_t\in I_g\}\), \(g'\) is bounded and Lipschitz on the relevant compact interval, and a three-term product decomposition gives
\[
\bigl|\widehat\chi_k\widehat g_k'(\widehat{\boldsymbol\theta}_k^\top\mathbf x_t)
-\chi_0g'(\boldsymbol\theta_0^\top\mathbf x_t)\bigr|
\le C\rho_A\left\{
\|\widehat{\boldsymbol\theta}_k-\boldsymbol\theta_0\|_2
+\sup_{u\in I_g}|\widehat g_k'(u)-g'(u)|
+|(\widehat{\boldsymbol\theta}_k-\boldsymbol\theta_0)^\top\mathbf x_t|
\right\}.
\]
Combining this inequality with
\[
\widehat{\boldsymbol\theta}_k^\top\mathbf x_t
+\widehat\chi_k\widehat g_k'(\widehat{\boldsymbol\theta}_k^\top\mathbf x_t)
-\boldsymbol\theta_0^\top\mathbf x_t^0
=(\widehat{\boldsymbol\theta}_k-\boldsymbol\theta_0)^\top\mathbf x_t
+\widehat\chi_k\widehat g_k'(\widehat{\boldsymbol\theta}_k^\top\mathbf x_t)
-\chi_0g'(\boldsymbol\theta_0^\top\mathbf x_t),
\]
then applying \eqref{eq:reported-prediction} and Minkowski's inequality, gives
\begin{equation*}
\left\{\mathbb E\left[\left|
\widehat{\boldsymbol\theta}_k^\top\mathbf x_t
+\widehat\chi_k\widehat g_k'(\widehat{\boldsymbol\theta}_k^\top\mathbf x_t)
-\boldsymbol\theta_0^\top\mathbf x_t^0
\right|^2
\mathbb I\{\widehat{\boldsymbol\theta}_k^\top\mathbf x_t\in I_g\}
\mid\mathcal F_{t-1}\right]\right\}^{1/2}
\le C\rho_A^2\left(
\|\widehat{\boldsymbol\theta}_k-\boldsymbol\theta_0\|_2
+\sup_{u\in I_g}|\widehat g_k'(u)-g'(u)|
\right).
\end{equation*}
The event \(\mathcal J_k\) supplies the sparse-estimation bound in Theorem~\ref{lem:sparse-valuation-main} and both pricing-map bounds in Lemma~\ref{lem:pricing-function-main}.
Since \(b_k\le1\), the derivative rate also dominates the function-estimation rate.
Substituting these bounds into \eqref{eq:knownA-safeguarded-price} and the preceding bound gives
\begin{equation}
\left\{\mathbb E\bigl[(p_t-p_t^\ast)^2
\mid\mathcal F_{t-1}\bigr]\right\}^{1/2}
\le C\rho_A^2\left(
 \frac{\lambda_k\sqrt{s_0}}{\lambda_{\min}}+b_k^{m-1}
 +\sqrt{\frac{\Lambda_{\ell_k}}{a_k b_k^5q_k}}
 \right).
\label{eq:proof-knownA-price-bound}
\end{equation}

This price bound gives the deployment-regret envelope for episode \(k\).
Because \(\mathcal J_k\) is \(\mathcal H_k^+\)-measurable and \(\mathcal H_k^+\subseteq\mathcal F_{t-1}\) for every \(t\in I_k'\), we may square \eqref{eq:proof-knownA-price-bound},
multiply by \(\mathbb I(\mathcal J_k)\), apply the conditional reduction \eqref{eq:conditional-regret-reduction}, and sum over the deployment block.
The tower property then gives
\begin{equation}
\mathbb E_{\mathrm K}\!\left[L_{2,k}\mathbb I(\mathcal J_k)\right]
\le C\rho_A^4\mathbb E_{\mathrm K}\!\left[
(\ell_k-a_k)\left\{
\frac{s_0\lambda_k^2}{\lambda_{\min}^2}+b_k^{2m-2}
+\frac{\Lambda_{\ell_k}}{a_kb_k^5q_k}
\right\}\mathbb I(\mathcal J_k)\right].
\label{eq:proof-knownA-exploit-regret}
\end{equation}

We now substitute the statistical rates.
Recall \(\Gamma_k=1+\log(d+1)+\log\ell_k\), and \(\Lambda_{\ell_k}=1+\log\ell_k+s_0\log(d+1)\).
By the choice of \(k_0\) above, the statistical premises hold throughout the present episode.
The definition of \(\lambda_k\) gives
\begin{equation}
\frac{s_0\lambda_k^2}{\lambda_{\min}^2}
=\frac{c_\lambda^2}{\lambda_{\min}^2}
\frac{s_0\Gamma_k}{a_k}.
\label{eq:proof-knownA-theta-rate}
\end{equation}
Since
\[
b_k=\left(\frac{\omega_k}{a_k}\right)^{1/(2m+3)},
\]
we have
\begin{equation}
b_k^{2m-2}
=
\omega_k^{r_m}a_k^{-r_m},
\qquad
\frac{\omega_k}{a_k b_k^5}
=
\omega_k^{r_m}a_k^{-r_m}.
\label{eq:proof-knownA-bandwidth-balance}
\end{equation}
Since \(q_k^{-1}\le B\omega_k\), the last identity also gives
\[
\frac{1}{a_kb_k^5q_k}
\le B\omega_k^{r_m}a_k^{-r_m}.
\]
Substitution into \eqref{eq:proof-knownA-exploit-regret}, together with \(\ell_k-a_k\le\ell_k\mathbb I\{a_k<\ell_k\}\), gives
\begin{equation}
\mathbb E_{\mathrm K}\!\left[L_{2,k}\mathbb I(\mathcal J_k)\right]
\le C\rho_A^4\mathbb E_{\mathrm K}\!\left[
\mathbb I\{a_k<\ell_k\}\ell_k
\left\{
\frac{s_0\Gamma_k}{a_k}
+\Lambda_{\ell_k}\omega_k^{r_m}a_k^{-r_m}
\right\}\mathbb I(\mathcal J_k)\right].
\label{eq:proof-knownA-exploit-bound}
\end{equation}
Both terms in \eqref{eq:proof-knownA-exploit-bound} are retained in the episode envelope; no pointwise comparison between them is needed.

It remains to control the complement of the joint event and aggregate the episode bounds.
The acquisition and deployment losses must be split at their respective information times.
Since \(\mathcal V_k^c\subseteq\mathcal J_k^c\) and the one-period realized revenue difference is at most \(B\), pathwise,
\[
L_{1,k}\mathbb I\{\mathcal V_k^c\}
+L_{2,k}\mathbb I(\mathcal J_k^c)
\le B\left\{a_k\mathbb I\{\mathcal V_k^c\}+(\ell_k-a_k)\mathbb I(\mathcal J_k^c)\right\}
\le B\ell_k\mathbb I(\mathcal J_k^c).
\]
Therefore the recovery bound \eqref{eq:regeneration-bad-event-bound} directly yields
\begin{equation}
\mathbb E_{\mathrm K}\left[
\sum_{k\ge k_0}\left\{L_{1,k}\mathbb I\{\mathcal V_k^c\}
+L_{2,k}\mathbb I(\mathcal J_k^c)\right\}\right]
\le B\mathbb E_{\mathrm K}\left[\sum_{k\ge k_0}\ell_k\mathbb I(\mathcal J_k^c)\right]
\le BC_5.
\label{eq:bad-event-regret-regeneration}
\end{equation}
This bound includes the episode in which a concentration failure occurs and all subsequent episodes needed to restore coverage.
The same choice of \(k_0\) gives \eqref{eq:cutoff-size}.
Since one-period regret is at most \(B\), the preceding episodes contribute at most \(B\sum_{k<k_0}\ell_k\le2B\ell_{k_0}\).

We now combine the episode terms.
Choose \(C_{\mathrm{env}}\) to dominate the fixed coefficients in \eqref{eq:proof-knownA-explore-regret} and \eqref{eq:proof-knownA-exploit-bound}, and define
\begin{equation*}
r_k:=C_{\mathrm{env}}\rho_A^4\left[
\omega_k^2a_k+
\mathbb I\{a_k<\ell_k\}\ell_k
\left\{\Lambda_{\ell_k}\omega_k^{r_m}a_k^{-r_m}
+\frac{s_0\Gamma_k}{a_k}\right\}\right].
\end{equation*}
If \(a_k=\ell_k\), the deployment block is empty; otherwise \eqref{eq:proof-knownA-exploit-bound} applies.
Hence \eqref{eq:proof-knownA-explore-regret} and \eqref{eq:proof-knownA-exploit-bound} imply
\[
\mathbb E_{\mathrm K}\!\left[
L_{1,k}\mathbb I(\mathcal V_k)
+L_{2,k}\mathbb I(\mathcal J_k)
\right]
\le \mathbb E_{\mathrm K}[r_k].
\]
Here \(\rho_A\ge1\) absorbs the fixed acquisition coefficient.
By construction, \(r_k\) satisfies \eqref{eq:technical-episode-envelope}, so \eqref{eq:technical-episode-rate} gives
\begin{equation*}
r_k
\le
C_{\mathrm{agg}}\rho_A^4(\ell_k\Lambda_{\ell_k})^{\alpha_m}\omega_k^{\kappa_m}.
\end{equation*}
Equations \eqref{eq:technical-episode-sum} and \eqref{eq:effective-width-factor}, together with the bad-event and early-episode bounds, give
\begin{equation*}
\Reg_{\mathrm K}(T)
\le
C_{\mathrm{agg}}\rho_A^4(\omega_{\mathrm K,T})^{\kappa_m}
(T\Lambda_T)^{\alpha_m}
+2B\ell_{k_0}+BC_5.
\end{equation*}
By \eqref{eq:cutoff-size}, the two horizon conditions in Theorem~\ref{thm:main_hd}, and the fixed lower bound \(\omega_{\mathrm K,T}\ge\omega_{\min}\), the cutoff term in the preceding display is absorbed by the leading rate after increasing \(c_0\) to dominate the fixed cutoff multiplier.
The fixed recovery term is absorbed by \(C_{\mathrm K}\).
Enlarging \(C_{\mathrm K}\) proves \eqref{eq:main-knownA-regret}; \(\omega_{\mathrm K,T}\le1\) then gives \eqref{eq:main-knownA-worst-case}.\hfill\(\square\)

\subsection{Proof of Lemma~\ref{lem:gamma-main}}
\label{app:proof-gamma}
Uniformly over periods \(t\in I_k'\) at which the matched-pair estimator is available, the matched least-squares identity reduces the estimation error in \eqref{eq:main-gamma-rate} to the error in \(g'\) evaluated at the reported features.
The matched archive may depend on buyer features, so its empirical covariance cannot be treated as if it came from an independently selected sample.
We instead establish a prediction bound over every chronological prefix of the truthful exploration observations.
Uniformity over the Lasso cone makes this bound valid for the random estimation error and every feature-dependent matched subset.

The reporting problem also supplies the feature bound used below.
Comparing any selected minimizer of \eqref{eq:buyer-report-problem} with the truthful report and using \(0\le g\le B\) gives
\[
\|\mathbf x_{t,\mathrm R}-\mathbf x_{t,\mathrm R}^0\|_2
\le \sqrt{2B/\lambda_{\min}(\mathbf A)},
\qquad
\|\mathbf x_t\|_\infty\le
W_{\mathrm{rep}}:=W_x+\sqrt{2B/\lambda_{\min}(\mathbf A)}.
\]
Thus, \(W_{\mathrm{rep}}\) is a uniform coordinate bound for every reported augmented feature vector.
Assumption~\ref{ass:manipulation-cost} also gives a fixed constant \(c_{\mathrm{int}}>0\) such that \(\boldsymbol\theta_0^\top\mathbf x_t\) is at least \(c_{\mathrm{int}}\) from the boundary of \(I_g\) for every optimal report.

To control feature-dependent matching, let \(S=\supp(\boldsymbol\theta_0)\) and recall the Lasso cone \(\mathcal C(S,3)=\{\mathbf v:\|\mathbf v_{S^c}\|_1
\le3\|\mathbf v_S\|_1\}\).
Define the exploration-prefix remainder
\begin{equation*}
\varepsilon_Q(n)
:=C_Q(d+1)^2W_x^4s_0^2
\exp\left\{-\frac{c_Qn}{W_x^4s_0^2}\right\},
\end{equation*}
where \(C_Q,c_Q>0\) are fixed numerical constants.

Let \(\mathbf x_1^0,\mathbf x_2^0,\ldots\) be the truthful exploration observations in chronological order, with observation times \(T_1,T_2,\ldots\).
For \(N\ge1\), write \(\overline{\boldsymbol\Sigma}^{(N)}:=N^{-1}\sum_{q=1}^N\boldsymbol\Sigma_{T_q}\) for the predictable Gram matrix of the first \(N\) records and define
\begin{equation*}
\mathcal Q_N
:=
\left\{
\left\|
\frac1N\sum_{q=1}^N\mathbf x_q^0\mathbf x_q^{0\top}-\overline{\boldsymbol\Sigma}^{(N)}
\right\|_{\max}
\le\frac1{16s_0}
\right\}.
\end{equation*}

\begin{lemma}
\label{lem:truthful-prefix-prediction}
Under Assumptions~\ref{ass:bounded-sparse} and \ref{ass:covariance}, for every integer \(n_0\ge1\),
\begin{equation}
\mathbb P\{\mathcal Q_N^c\text{ for some }N\ge n_0\}
\le\varepsilon_Q(n_0).
\label{eq:truthful-prefix-probability}
\end{equation}
Moreover, on \(\mathcal Q_N\), every \(\mathbf v\in\mathcal C(S,3)\) satisfies
\begin{equation}
\frac1N\sum_{q=1}^N(\mathbf v^\top\mathbf x_q^0)^2
\le(\lambda_{\max}+1)\|\mathbf v\|_2^2.
\label{eq:truthful-prefix-prediction}
\end{equation}
\end{lemma}

The proof of Lemma~\ref{lem:truthful-prefix-prediction} is given in Appendix~\ref{app:proof-truthful-prefix}.
Let \(c_g>0\) be the lower bound on \(|g'|\) over the range covered by matched records.
Recall that \(R_{2,k}\) and \(R_{1,k}\) are the \(\ell_2\)- and \(\ell_1\)-radii of the Lasso error, while \(H_k(\delta_k)\) is the stochastic term in the kernel bound.
The definition of \(\lambda_k\), the bandwidth identity, \(q_k\ge1/B\), and \(H_k(\delta_k)^2\le C\Lambda_{\ell_k}\) give
\[
W_{\mathrm{rep}}R_{1,k}\lesssim
\frac{W_{\mathrm{rep}}b_k}{\sqrt{B_6}},
\qquad
R_{2,k}\lesssim\frac{b_k}{\sqrt{B_6}},
\qquad
b_k^{m-1}\le(a_kb_k^5)^{-1/2},
\qquad
\frac{H_k(\delta_k)}{\sqrt{a_kb_k^5q_k}}
\lesssim\frac1{\sqrt{B_6}}.
\]
Let \(C_{\mathrm{der}}>0\) be the fixed perturbation multiplier supplied by the Lipschitz and uniform derivative bounds;
it depends only on the compact-interval regularity constants.
Choosing \(B_6\) sufficiently large first gives
\begin{equation}
W_{\mathrm{rep}}R_{1,k}\le\frac{c_{\mathrm{int}}}{2}.
\label{eq:matched-input-interior}
\end{equation}
It also gives
\begin{equation}
C_{\mathrm{der}}\left\{
b_k^{m-1}
+\frac{H_k(\delta_k)}{\sqrt{a_kb_k^5q_k}}
+R_{2,k}+W_{\mathrm{rep}}R_{1,k}
\right\}
\le\frac{c_g}{2}.
\label{eq:finite-sample-gamma}
\end{equation}

Work on the intersection of the score and restricted-eigenvalue events used in the proof of Theorem~\ref{lem:sparse-valuation-main}, the kernel event used in the proof of Lemma~\ref{lem:pricing-function-main}, and \(\bigcap_{N\ge a_k^{\mathrm{ref}}}\mathcal Q_N\).
On \(\mathcal G_k\), these events supply, respectively, the Lasso cone and norm bounds, the uniform pricing-map derivative bound, and prediction control over every truthful prefix.
On \(\mathcal G_k\) and this intersection, the remaining argument is deterministic.
Fix a period \(t\in I_k'\) at which the matched-pair estimator is available.
Let \(\mathcal M\) contain the pairs available before the current price, put \(n_t=|\mathcal M|\ge1\), and write
\[
g_i'=g'(\boldsymbol\theta_0^\top\mathbf x_i),
\qquad
\widehat g_i'=\widehat g_k'(\widehat{\boldsymbol\theta}_k^\top\mathbf x_i).
\]
Equation~\eqref{eq:delta} gives \(\boldsymbol\delta_i=\boldsymbol\gamma_0g_i'\) exactly, so there is no additional regression error.

The derivative condition supplies the lower bound \(c_g\).
Continuity on the compact interval \(I_g\) also supplies a fixed upper bound \(\bar c_g\), so \(c_g\le |g'(\boldsymbol\theta_0^\top\mathbf x_i)|\le \bar c_g\) for every matched record \(i\).
The \(\ell_1\) bound in \eqref{eq:raw-lasso-rate} and \eqref{eq:matched-input-interior} keep \(\widehat{\boldsymbol\theta}_k^\top\mathbf x_i\) in \(I_g\).
Lipschitz continuity of \(g'\) and \eqref{eq:gprime-lemma-rate} then bound the derivative error by the left-hand side of \eqref{eq:finite-sample-gamma}.
Hence
\[
\max_{i\in\mathcal M}|\widehat g_i'-g_i'|\le\frac{c_g}{2}.
\]
It follows that
\[
\min_{i\in\mathcal M}|\widehat g_i'|\ge\frac{c_g}2,
\qquad
\sum_{i\in\mathcal M}(\widehat g_i')^2
\ge\frac{c_g^2}{4}n_t.
\]

The vector least-squares identity gives
\begin{equation*}
\|\widehat{\boldsymbol\gamma}_t-\boldsymbol\gamma_0\|_2^2
=
\|\boldsymbol\gamma_0\|_2^2\left|
\frac{\sum_{i\in\mathcal M}
\widehat g_i'(g_i'-\widehat g_i')}
{\sum_{i\in\mathcal M}(\widehat g_i')^2}
\right|^2
\le
C\|\boldsymbol\gamma_0\|_2^2\frac1{n_t}
\sum_{i\in\mathcal M}|g_i'-\widehat g_i'|^2 .
\end{equation*}
The same denominator bound also gives
\begin{equation*}
\|\widehat{\boldsymbol\gamma}_t\|_2
=\|\boldsymbol\gamma_0\|_2
\left|
\frac{\sum_{i\in\mathcal M}\widehat g_i'g_i'}
{\sum_{i\in\mathcal M}(\widehat g_i')^2}
\right|
\le C\|\boldsymbol\gamma_0\|_2
\le C\rho_A.
\end{equation*}

It remains to control the prediction error on the selected records.
Recall \(N_k=\sum_{j\le k}a_j\) and \(n_t\) from Definition~\ref{def:repeat-buyer-rate}.
Since SALE-U retains one pair per matched ID, \(n_t=|\mathcal M|\) immediately before pricing.
Equation~\eqref{eq:effective-repeat-rate} gives \(\max\{1,n_t\}\ge\tau N_k\) almost surely, simultaneously over all exploitation decisions.
In particular, \(n_t\ge\tau N_k\) whenever the matched-pair estimator is available.
At every exploitation decision in episode \(k\), take \(N=N_k\).
Since \(N\ge a_k\ge a_k^{\mathrm{ref}}\), Lemma~\ref{lem:truthful-prefix-prediction} applies with \(n_0=a_k^{\mathrm{ref}}\).
The truthful records in \(\mathcal M\) form a subset of this prefix.
Since the summands are nonnegative, the full-prefix bound controls their sum.
Its uniformity over \(\mathcal C(S,3)\) allows us to substitute the random Lasso error even when the matched subset depends on the features.
By \eqref{eq:effective-repeat-rate}, whenever \(n_t\ge1\),
\begin{equation*}
\frac{N}{n_t}\le\frac1\tau.
\end{equation*}

We now transfer the prefix bound to the matched sample.
Embed the reportable displacement in the full feature space by setting \(\boldsymbol\gamma_0^{\mathrm{aug}}
=(\mathbf0_{d_{\mathrm V}}^\top,\boldsymbol\gamma_0^\top,0)^\top\).
For every \(\mathbf v\in\mathcal C(S,3)\), the identity \(\mathbf x_i=\mathbf x_i^0+\boldsymbol\gamma_0^{\mathrm{aug}}g_i'\), the bound \(N/n_t\le1/\tau\), and \(|g_i'|\le \bar c_g\) give
\begin{equation}
\begin{aligned}
\frac1{n_t}\sum_{i\in\mathcal M}(\mathbf v^\top\mathbf x_i)^2
&\le \frac2{n_t}\sum_{i\in\mathcal M}(\mathbf v^\top\mathbf x_i^0)^2
+2\bar c_g^2(\mathbf v^\top\boldsymbol\gamma_0^{\mathrm{aug}})^2\\
&\lesssim \frac1{\tau N}\sum_{q=1}^N(\mathbf v^\top\mathbf x_q^0)^2
+\|\boldsymbol\gamma_0^{\mathrm{aug}}\|_2^2\|\mathbf v\|_2^2
\le \frac{C\rho_A^2}{\tau}\|\mathbf v\|_2^2 .
\end{aligned}
\label{eq:matched-prediction-bound}
\end{equation}
The last inequality uses \(\rho_A\ge1\), \(\tau\le1\), and \(\|\boldsymbol\gamma_0^{\mathrm{aug}}\|_2\le W_\theta\rho_A\).
Thus feature-dependent matching contributes the effective-sample factor \(1/\tau\), but does not add a polynomial factor in the ambient dimension.

By Lipschitz continuity of \(g'\),
\[
|g_i'-\widehat g_i'|
\le
\sup_{u\in I_g}|\widehat g_k'(u)-g'(u)|
+C|(\widehat{\boldsymbol\theta}_k-\boldsymbol\theta_0)^\top\mathbf x_i|.
\]
Applying \eqref{eq:matched-prediction-bound} to \(v=\widehat{\boldsymbol\theta}_k-\boldsymbol\theta_0\), and using \(\|\boldsymbol\gamma_0\|_2\le W_\theta\rho_A\), we obtain
\[
\|\widehat{\boldsymbol\gamma}_t-\boldsymbol\gamma_0\|_2^2
\le \frac{C\rho_A^4}{\tau}\left\{
\|\widehat{\boldsymbol\theta}_k-\boldsymbol\theta_0\|_2^2+
\sup_{u\in I_g}|\widehat g_k'(u)-g'(u)|^2
\right\}.
\]
Equations \eqref{eq:raw-lasso-rate} and \eqref{eq:gprime-lemma-rate} control the parameter and derivative errors, respectively.
Substitution into the preceding display gives
\begin{equation}
\|\widehat{\boldsymbol\gamma}_t-\boldsymbol\gamma_0\|_2^2
\le \frac{C_6\rho_A^4}{\tau}\left\{
R_{2,k}^2+b_k^{2m-2}
+\frac{H_k(\delta_k)^2}{a_kb_k^5q_k}
\right\}.
\label{eq:semi-gamma-rate}
\end{equation}
Because the same events control every truthful prefix and every matched subset, \eqref{eq:semi-gamma-rate} holds simultaneously for all \(t\in I_k'\) at which the matched-pair estimator is available.
No union bound over \(t\) is needed.
Using \(R_{2,k}^2=9s_0\lambda_k^2/\lambda_{\min}^2\) and \(H_k(\delta_k)^2\le C\Lambda_{\ell_k}\) gives \eqref{eq:main-gamma-rate}.

It remains to bound the complement of the preceding intersection.
The shared restricted-eigenvalue event contributes \(\varepsilon_{\mathrm{RE}}(a_k^{\mathrm{ref}})\), the score and kernel events contribute \(\delta_k\) each, and the prefix event contributes \(\varepsilon_Q(a_k^{\mathrm{ref}})\).
Schedule comparability, the second sample-size condition in Lemma~\ref{lem:gamma-main}, and \(b_k\le1\) give
\[
a_k^{\mathrm{ref}}
\ge\frac{B_6}{C_\omega}s_0^2
\{1+\log\ell_k+\log(d+1)\}.
\]
Thus \eqref{eq:RE-remainder} is at most \(\delta_k\) after a fixed increase of \(B_6\).
The prefix remainder satisfies
\[
\varepsilon_Q(a_k^{\mathrm{ref}})
\le C_Q(d+1)^2W_x^4s_0^2
\exp\left[-\frac{c_QB_6}{C_\omega W_x^4}
\{1+\log\ell_k+\log(d+1)\}\right].
\]
Since \(s_0\le d+1\), another fixed increase of \(B_6\) makes this bound at most \(\delta_k\).
Hence
\[
\varepsilon_{\mathrm{RE}}(a_k^{\mathrm{ref}})
+\varepsilon_Q(a_k^{\mathrm{ref}})+2\delta_k
\le4\delta_k\le4\ell_k^{-2}.
\]
Taking expectations of the conditional restricted-eigenvalue, score, and kernel bounds gives the corresponding unconditional bounds for their failure events intersected with \(\mathcal G_k\).
The prefix bound is unconditional, and no independence among these events is used.
This proves Lemma~\ref{lem:gamma-main}.
\hfill\(\square\)

\subsection{Proof of Theorem~\ref{thm:main_hd_unknownA}}

Use the episode losses \(L_{1,k},L_{2,k}\), the pre-exploration event \(\mathcal V_k\), and the joint estimation event \(\mathcal J_k\) from the SALE-K proof.
The exploration, recovery, and early-episode bounds are unchanged.
The only new step is to control deployment regret when the strategic adjustment is estimated from matched records.
The generic-constant convention from the SALE-K proof remains in force, and \(\mathbb E_{\mathrm U}\) denotes expectation under SALE-U.

For SALE-U, choose the fixed threshold in Lemma~\ref{lem:finite-localization} large enough also to dominate \(B_6\), and denote the resulting, possibly larger, cutoff again by \(k_0\).
Lemma~\ref{lem:gamma-main} then applies in every episode \(k\ge k_0\).
This changes only the fixed cutoff multiplier and is independent of \(\tau\).
Recall \(N_k\) and \(n_t\) from Definition~\ref{def:repeat-buyer-rate}, and let \(\mathcal M_t\) be the matched archive immediately before pricing.
Thus \(n_t=|\mathcal M_t|\), and \eqref{eq:effective-repeat-rate} holds simultaneously over deployment decisions.

The policy enters the direct branch on \(I_{k,\mathrm{dir}}'=\{t\in I_k':\mathrm{id}_t\in\mathcal M_t\}\).
Let \(I_{k,\mathrm{cor}}'\) contain the periods in which the policy uses the adjustment estimator in \eqref{eq:gamma_est}.
The remaining periods use the temporary uncorrected price.
These three sets partition the deployment block and correspond exactly to the branches of Algorithm~\ref{alg:unknownA}.
On \(\mathcal J_k\), the fixed calibration ensures that the matched-pair estimator is available whenever \(n_t\ge1\), so the correction branch is then \(\{t\in I_k':\mathrm{id}_t\notin\mathcal M_t,\ n_t\ge1\}\);
consequently, a temporary uncorrected decision on \(\mathcal J_k\) has \(n_t=0\).

For each deployment period \(t\), let \(\mathcal A_{k,t}\) be the event that either the policy does not enter the correction branch or its adjustment estimate satisfies \eqref{eq:main-gamma-rate}.
Set \(\mathcal A_k=\bigcap_{t\in I_k'}\mathcal A_{k,t}\).
The uniform conclusion of Lemma~\ref{lem:gamma-main} gives
\begin{equation}
\mathbb P(\mathcal G_k\cap\mathcal A_k^c)\le4\ell_k^{-2}.
\label{eq:unknownA-adjustment-failure}
\end{equation}
As in the SALE-K proof, we first condition on \(\widetilde{\mathcal F}_{t-1}\) to apply the regret reduction and then use the tower property to return to \(\mathcal F_{t-1}\).

We first control the correction branch.
Since \(-\boldsymbol\beta_{\mathrm R,0}^\top\boldsymbol\gamma_0=\chi_0\), \eqref{eq:proof-knownA-oracle-price} gives
\[
\boldsymbol\theta_0^\top\mathbf x_t-
\boldsymbol\beta_{\mathrm R,0}^\top\boldsymbol\gamma_0
g'(\boldsymbol\theta_0^\top\mathbf x_t)
=\boldsymbol\theta_0^\top\mathbf x_t^0.
\]
For \(t\in I_{k,\mathrm{cor}}'\), the policy uses the estimated argument
\[
\widehat{\boldsymbol\theta}_k^\top\mathbf x_t-
\widehat{\boldsymbol\beta}_{\mathrm R,k}^\top
\widehat{\boldsymbol\gamma}_t
\widehat g_k'(\widehat{\boldsymbol\theta}_k^\top\mathbf x_t).
\]
On \(\mathcal J_k\cap
\{\widehat{\boldsymbol\theta}_k^\top\mathbf x_t\in I_g\}\), add and subtract the products formed by replacing \(\widehat{\boldsymbol\gamma}_t\) and then \(\widehat{\boldsymbol\beta}_{\mathrm R,k}\) by their population values.
The bounded parameter space, \(\|\boldsymbol\gamma_0\|_2\le W_\theta\rho_A\), and Lipschitz continuity of \(g'\) then give
\begin{equation}
\begin{aligned}
\left|
\widehat{\boldsymbol\theta}_k^\top\mathbf x_t-
\widehat{\boldsymbol\beta}_{\mathrm R,k}^\top
\widehat{\boldsymbol\gamma}_t
\widehat g_k'(\widehat{\boldsymbol\theta}_k^\top\mathbf x_t)
-\boldsymbol\theta_0^\top\mathbf x_t^0
\right|
\le C\bigl\{&\rho_A|(\widehat{\boldsymbol\theta}_k-\boldsymbol\theta_0)^\top\mathbf x_t|
+\rho_A\|\widehat{\boldsymbol\beta}_{\mathrm R,k}
-\boldsymbol\beta_{\mathrm R,0}\|_2\\
&+\|\widehat{\boldsymbol\gamma}_t-\boldsymbol\gamma_0\|_2
+\rho_A\|\widehat g_k'-g'\|_{\infty,I_g}\bigr\}.
\end{aligned}
\label{eq:unknownA-corrected-index-decomposition}
\end{equation}
The common squared statistical error in this display is
\begin{equation*}
E_k:=\frac{s_0\lambda_k^2}{\lambda_{\min}^2}
+b_k^{2m-2}+\frac{\Lambda_{\ell_k}}{a_kb_k^5q_k}.
\end{equation*}
On \(\mathcal J_k\), Theorem~\ref{lem:sparse-valuation-main} controls the two parameter errors, while Lemma~\ref{lem:pricing-function-main} controls the pricing-map derivative.
On \(\mathcal A_{k,t}\), Lemma~\ref{lem:gamma-main} controls the adjustment error.
Applying the reported-feature prediction bound \eqref{eq:reported-prediction} to the first term in \eqref{eq:unknownA-corrected-index-decomposition} therefore gives
\begin{equation*}
\begin{aligned}
&\left\{\mathbb E\left[\left|
\widehat{\boldsymbol\theta}_k^\top\mathbf x_t-
\widehat{\boldsymbol\beta}_{\mathrm R,k}^\top
\widehat{\boldsymbol\gamma}_t
\widehat g_k'(\widehat{\boldsymbol\theta}_k^\top\mathbf x_t)
-\boldsymbol\theta_0^\top\mathbf x_t^0
\right|^2
\mathbb I\{t\in I_{k,\mathrm{cor}}'\}
\mathbb I\{\widehat{\boldsymbol\theta}_k^\top\mathbf x_t\in I_g\}
\mathbb I(\mathcal J_k\cap\mathcal A_{k,t})
\mid\mathcal F_{t-1}\right]\right\}^{1/2}
\\
&\qquad\le C\rho_A^2(1+\tau^{-1/2})E_k^{1/2}
\mathbb I(\mathcal J_k).
\end{aligned}
\end{equation*}
The three-case safeguard leading to \eqref{eq:knownA-safeguarded-price} then yields constants \(C_{\mathrm U,0},C_{\mathrm U,1}>0\), independent of \(\tau\), such that
\begin{equation}
\mathbb E\left[(p_t-p_t^\ast)^2
\mathbb I\{t\in I_{k,\mathrm{cor}}'\}
\mathbb I(\mathcal J_k\cap\mathcal A_{k,t})
\mid\mathcal F_{t-1}\right]
\le\rho_A^4\left(C_{\mathrm U,0}
+\frac{C_{\mathrm U,1}}{\tau}\right)E_k\mathbb I(\mathcal J_k).
\label{eq:unknownA-corrected-price}
\end{equation}
The additional factor \(1/\tau\) arises only from estimating \(\boldsymbol\gamma_0\).

We next consider the direct branch, where SALE-U retrieves the truthful feature vector.
Multiplication by the direct-branch indicator can only decrease the conditional prediction error.
Hence Assumption~\ref{ass:covariance} and Theorem~\ref{lem:sparse-valuation-main} give, on \(\mathcal J_k\),
\[
\mathbb E\left[\{(\widehat{\boldsymbol\theta}_k-
\boldsymbol\theta_0)^\top\mathbf x_t^0\}^2
\mathbb I\{t\in I_{k,\mathrm{dir}}'\}\mathbb I(\mathcal J_k)
\mid\mathcal F_{t-1}\right]
\le\frac{9\lambda_{\max}s_0\lambda_k^2}{\lambda_{\min}^2}
\mathbb I(\mathcal J_k).
\]
Combining this bound with Lemma~\ref{lem:pricing-function-main} and the same safeguard argument gives
\begin{equation}
\mathbb E\left[(p_t-p_t^\ast)^2
\mathbb I\{t\in I_{k,\mathrm{dir}}'\}\mathbb I(\mathcal J_k)
\mid\mathcal F_{t-1}\right]
\le C_{\mathrm U,0}\rho_A^4E_k\mathbb I(\mathcal J_k).
\label{eq:unknownA-direct-price}
\end{equation}

We now combine the two regular branches.
The acquisition calculation in \eqref{eq:proof-knownA-explore-regret} remains valid with \(\mathbb E_{\mathrm U}\).
For deployment, apply \eqref{eq:conditional-regret-reduction} to the two price bounds \eqref{eq:unknownA-corrected-price} and \eqref{eq:unknownA-direct-price}, and sum over the block.
This gives
\begin{equation*}
\begin{aligned}
&\mathbb E_{\mathrm U}[L_{1,k}\mathbb I(\mathcal V_k)]
+\mathbb E_{\mathrm U}\sum_{t\in I_k'}\Delta_t\mathbb I(\mathcal J_k)
\left[\mathbb I\{t\in I_{k,\mathrm{dir}}'\}
+\mathbb I\{t\in I_{k,\mathrm{cor}}'\}\mathbb I(\mathcal A_{k,t})\right]\\
&\qquad\le\rho_A^4\left(C_{\mathrm U,0}
+\frac{C_{\mathrm U,1}}{\tau}\right)
\mathbb E_{\mathrm U}\left[\omega_k^2a_k+(\ell_k-a_k)E_k\right].
\end{aligned}
\end{equation*}
Substituting
\eqref{eq:proof-knownA-theta-rate}--%
\eqref{eq:proof-knownA-bandwidth-balance} bounds its integrand by a fixed multiple of
\[
\omega_k^2a_k+
\mathbb I\{a_k<\ell_k\}\ell_k\left\{
\Lambda_{\ell_k}\omega_k^{r_m}a_k^{-r_m}+
\frac{s_0\Gamma_k}{a_k}\right\},
\]
which is the episode envelope in \eqref{eq:technical-episode-envelope}.

It remains to control the temporary uncorrected branch and the two failure events.
On \(\mathcal J_k\), any period in that branch has \(n_t=0\).
Definition~\ref{def:repeat-buyer-rate} then gives \(\tau N_k\le1\).
Since \(N_k\ge a_k\), it follows that \(1/\tau\ge a_k\).
Such a period can occur only when \(a_k<\ell_k\).
The uncapped schedule gives
\[
\frac1\tau(\ell_k\Lambda_{\ell_k})^{\alpha_m}\omega_k^{\kappa_m}
\gtrsim(\ell_k\Lambda_{\ell_k})^{2\alpha_m}
\omega_k^{\kappa_m-\nu_m}\gtrsim\ell_k,
\]
where the last comparison uses \(2\alpha_m-1=5/(4m+1)>0\) and \(\omega_k\ge\omega_{\min}\).
Thus, on \(\mathcal J_k\), the trivial loss on the temporary uncorrected branch is absorbed by the same \(\tau^{-1}\)-weighted episode envelope.

For adjustment failures, the three branch indicators and \(\mathcal A_k=\bigcap_t\mathcal A_{k,t}\) give the pathwise bound
\[
\sum_{t\in I_{k,\mathrm{cor}}'}\Delta_t
\mathbb I(\mathcal J_k)\mathbb I(\mathcal A_{k,t}^c)
\le B\ell_k\mathbb I(\mathcal J_k\cap\mathcal A_k^c).
\]
Since \(\mathcal J_k\subseteq\mathcal G_k\), \eqref{eq:unknownA-adjustment-failure} and the geometric episode schedule imply
\begin{equation}
\sum_{k\ge k_0}B\ell_k
\mathbb P(\mathcal J_k\cap\mathcal A_k^c)
\le4B\sum_{k\ge k_0}\ell_k^{-1}\le C.
\label{eq:unknownA-adjustment-failure-cost}
\end{equation}
Exactly as in \eqref{eq:bad-event-regret-regeneration},
\begin{equation}
\mathbb E_{\mathrm U}\sum_{k\ge k_0}
\left\{L_{1,k}\mathbb I(\mathcal V_k^c)
+L_{2,k}\mathbb I(\mathcal J_k^c)\right\}
\le BC_5.
\label{eq:unknownA-common-failure-cost}
\end{equation}

Applying \eqref{eq:technical-episode-sum} to the regular-branch and empty-archive envelopes and adding \eqref{eq:unknownA-adjustment-failure-cost}, \eqref{eq:unknownA-common-failure-cost}, and the early-episode loss gives, for sufficiently large fixed \(C_{\mathrm U,0},C_{\mathrm U,1}\),
\begin{equation*}
\Reg_{\mathrm U}(T)
\le\rho_A^4\left(C_{\mathrm U,0}+\frac{C_{\mathrm U,1}}{\tau}\right)
(\omega_{\mathrm U,T})^{\kappa_m}(T\Lambda_T)^{\alpha_m}
+2B\ell_{k_0}+BC_5+C.
\end{equation*}
The cutoff bound, with the fixed SALE-U threshold chosen above, and the two horizon conditions in Theorem~\ref{thm:main_hd} absorb the cutoff term into the leading rate.
This enlargement changes only the fixed onset constant and is independent of \(\tau\).
The two fixed remainder terms are absorbed by \(C_{\mathrm U,0}\).
This proves \eqref{eq:main-unknownA-regret}; using \(\omega_{\mathrm U,T}\le1\) gives \eqref{eq:main-unknownA-worst-case}.

\phantomsection\label{app:proof-unknownA-moment}
For the moment condition in Remark~\ref{rem:liu-repeat-rate}, work instead on \(\mathcal J_k\cap\mathcal Q_{N_k}\).
The definition of \(\mathcal Q_{N_k}\) and \(R_{1,k}=4\sqrt{s_0}R_{2,k}\) bound the mean squared prediction error over the first \(N_k\) truthful records by \((\lambda_{\max}+1)R_{2,k}^2\).
Retaining \(N_k/n_t\) in the matched-record calculation and using \(q_k^{-1}\le B\omega_k\) therefore gives, whenever \(n_t\ge1\),
\[
\|\widehat{\boldsymbol\gamma}_t-\boldsymbol\gamma_0\|_2^2
\le C\rho_A^4\left(1+\frac{N_k}{\max\{1,n_t\}}\right)
\left\{\frac{s_0\lambda_k^2}{\lambda_{\min}^2}
+b_k^{2m-2}+\frac{B\Lambda_{\ell_k}\omega_k}{a_kb_k^5}\right\}.
\]
The braced expression and \(N_k\) are \(\mathcal F_{S_k-1}\)-measurable.
The denominator argument above still gives \(\|\widehat{\boldsymbol\gamma}_t\|_2\le C\rho_A\), with a constant independent of \(n_t\).

Apply \eqref{eq:conditional-regret-reduction} before bounding the event indicators.
For the prediction term, discard the prefix-event indicator and use \eqref{eq:reported-prediction}; for the adjustment term, use the preceding bound and the moment condition.
The tower property yields
\[
\mathbb E_{\mathrm U}\!\left[\sum_{t\in I_k'}\Delta_t\,
\mathbb I(\mathcal J_k\cap\mathcal Q_{N_k})
\,\middle|\,\mathcal F_{S_k-1}\right]
\le C\rho_A^4(\ell_k-a_k)(1+\tau^{-1})
\left\{\frac{s_0\lambda_k^2}{\lambda_{\min}^2}
+b_k^{2m-2}+\frac{B\Lambda_{\ell_k}\omega_k}{a_kb_k^5}\right\}
+\frac{B(\ell_k-a_k)}{\tau N_k}.
\]
The last term covers the empty archive, since \(\mathbb P(n_t=0\mid\mathcal F_{S_k-1})\le1/(\tau N_k)\), and is absorbed by the sparse term in the episode envelope.
Lemma~\ref{lem:truthful-prefix-prediction} and the existing calibration make \(\sum_{k\ge k_0}B\ell_k\mathbb P(\mathcal Q_{N_k}^c)\) finite; the other failure and recovery costs are unchanged.
The same episode summation proves \eqref{eq:main-unknownA-regret} and \eqref{eq:main-unknownA-worst-case} under the moment condition.
\hfill\(\square\)

\section{Proofs of Technical Lemmas}
\label{app:technical-proofs}

\subsection{Proof of Lemma~\ref{lem:kernel-stochastic}}
\label{app:proof-kernel-stochastic}

Work on \(\mathcal G_k\) under the conditional law of Appendix~\ref{app:proof-noise-function}, which conditions on \(\mathcal H_{k-1}\) and the realized exploration contexts.
The argument is for conditionally almost every such context path.
Fix \(j\in\{0,1,2\}\).
We prove the bound for \(h_k^{(j)}\); the argument for \(f_k^{(j)}\) is identical after replacing \(y_t\) by \(1\).
Write \(b=b_k\) and \(K_j=K^{(j)}\).
Allocate failure probability \(\delta/6\) to this estimator--derivative pair, and define the centered process
\begin{equation}
Z_j(u,\boldsymbol\theta)
:=\frac1n\sum_{t\in I_k}\left[
\frac{y_t}{b^{j+1}}K_j\!\left(\frac{w_t(\boldsymbol\theta)-u}{b}\right)
-\mathbb E\!\left\{
\frac{y_t}{b^{j+1}}K_j\!\left(\frac{w_t(\boldsymbol\theta)-u}{b}\right)
\right\}\right].
\label{eq:localized-generated-process}
\end{equation}

The first step controls the occupancy of every kernel window.
Recall that \([-1,1]\) contains the supports of \(K,K'\), and \(K''\).
On \(\mathcal G_k\), the endpoint bounds in \eqref{eq:coverage-residual-margin} give \(U_k(\mathbf x_{t,\mathrm V}^0)-L_k(\mathbf x_{t,\mathrm V}^0)\ge2\delta_z\) on the context paths under consideration.
Conditional on \(\mathbf x_t^0\), the posted price is uniform on an interval of width \(W_k(\mathbf x_{t,\mathrm V}^0)\).
Intersecting a window with that interval can only reduce its probability.
Hence, for every fixed \(a>0\),
\begin{equation}
\sup_{u\in\mathbb R}\frac1n\sum_{t\in I_k}\mathbb P\!\left(|w_t(\boldsymbol\theta_0)-u|\le ab\right)
\le \frac{2ab}{n}\sum_{t\in I_k}W_k(\mathbf x_{t,\mathrm V}^0)^{-1}=2abq_k.
\label{eq:localized-window-probability}
\end{equation}

To control local occupancy, set
	\[
	N_b:=\sup_{u\in I_z}
	\sum_{t\in I_k}
\mathbb I\{|w_t(\boldsymbol\theta_0)-u|\le 5b\}.
	\]
Cover \(I_z\) by a deterministic grid of mesh \(b\).
Every interval in the definition of \(N_b\) is contained in one centered at a grid point with radius \(6b\).
The grid has at most \(C/b\) points, where \(C\) depends only on \(I_z\).
Apply Bernstein's inequality for independent, nonidentically distributed bounded variables \citep[Theorem~2.10]{BoucheronLugosiMassart2013} at each grid point.
A union bound and \eqref{eq:localized-window-probability} then give, for every \(x\ge1\),
	\begin{equation*}
	\mathbb P\left\{
	N_b>C\left(nbq_k+x+\log\frac{2}{b}\right)
	\right\}
	\le e^{-x}.
	\end{equation*}
Integrating this tail bound and using the sample-size condition \(nbq_k\ge C\log\{96(d+1)n/(b\delta)\}\) yields
	\begin{equation}
	\mathbb E\sqrt{N_b}\le C\sqrt{nbq_k}.
	\label{eq:localized-window-occupancy-mean}
	\end{equation}

With the occupancy bound in hand, we discretize the index set and control the process at \(\boldsymbol\theta_0\).
Let \(\mathcal U_n\) be a deterministic grid of \(I_z\) with mesh \(b/n^2\), and note that
	\begin{equation*}
	|\mathcal U_n|\le \frac{Cn^2}{b},
	\qquad
	\log\{2|\mathcal U_n|\}\le C\left\{\log n+\log\frac1b\right\}.
	\end{equation*}
	The Lipschitz property of \(K_j\) and the bound \(|y_t|\le1\) imply
	\begin{equation}
	\sup_{u\in I_z,\,\boldsymbol\theta\in\Theta_k}|Z_j(u,\boldsymbol\theta)|
	\le
	\max_{u\in\mathcal U_n}\sup_{\boldsymbol\theta\in\Theta_k}|Z_j(u,\boldsymbol\theta)|
	+\frac{C}{n^2b^{j+1}}.
	\label{eq:fine-residual-grid-remainder}
	\end{equation}
	Thus it suffices to control the process on \(\mathcal U_n\).

	At \(\boldsymbol\theta_0\), the summand in \eqref{eq:localized-generated-process} is bounded by \(C b^{-(j+1)}\).
	Equation \eqref{eq:localized-window-probability} also gives the uniform variance bound
	\begin{equation*}
	\sup_{u\in I_z}
	\frac1n\sum_{t\in I_k}\mathbb E\left\{
	\left[
	\frac{K_j((w_t(\boldsymbol\theta_0)-u)/b)}{b^{j+1}}y_t
	\right]^2
	\right\}
	\le \frac{Cq_k}{b^{2j+1}}.
	\end{equation*}
	Apply the same Bernstein inequality and take a union bound over \(\mathcal U_n\).
	Except on an event of probability at most \(\delta/12\),
\begin{equation}
\max_{u\in\mathcal U_n}|Z_j(u,\boldsymbol\theta_0)|
\le C\left\{\sqrt{\frac{q_k[\log\{2|\mathcal U_n|\}+\log(12/\delta)]}{nb^{2j+1}}}
+\frac{\log\{2|\mathcal U_n|\}+\log(12/\delta)}{nb^{j+1}}\right\}.
\label{eq:anchor-grid-process-bound}
\end{equation}

The preceding bound uses the true residuals.
We now control the perturbation from estimating \(\boldsymbol\theta_0\).
The admissible parameter errors form the set
	\[
	\mathcal D_k:=\{\boldsymbol\Delta:\boldsymbol\theta_0+\boldsymbol\Delta\in\Theta_k\}.
	\]
	Then \(\|\boldsymbol\Delta\|_1\le R_{1,k}\) for every \(\boldsymbol\Delta\in\mathcal D_k\).
	For a grid point \(u\in\mathcal U_n\), define the resulting kernel increment by
	\begin{align*}
	\Delta K_{t,u}(\boldsymbol\Delta)
	:=y_t\Bigg\{
	K_j\!\left(\frac{w_t(\boldsymbol\theta_0+\boldsymbol\Delta)-u}{b}\right)
	-
	K_j\!\left(\frac{w_t(\boldsymbol\theta_0)-u}{b}\right)
	\Bigg\}.
	\end{align*}
By the localization condition \(W_xR_{1,k}\le b\), \(|(\mathbf x_t^0)^\top\boldsymbol\Delta|\le W_xR_{1,k}\le b\).
Therefore \(\Delta K_{t,u}(\boldsymbol\Delta)=0\) unless \(|w_t(\boldsymbol\theta_0)-u|\le2b\).
On this enlarged window, the map \(z\mapsto K_j((w_t(\boldsymbol\theta_0)-u-z)/b)-K_j((w_t(\boldsymbol\theta_0)-u)/b)\) is \(C/b\)-Lipschitz and vanishes at zero.
For observations outside the enlarged window, this map is identically zero on \(|z|\le W_xR_{1,k}\).
Inside the window, restrict it to the same interval and extend it to the real line with the same Lipschitz constant and value at the origin.
For every \(\boldsymbol\Delta\in\mathcal D_k\), evaluating the extension at \(z=(\mathbf x_t^0)^\top\boldsymbol\Delta\) and multiplying by \(y_t\) recovers \(\Delta K_{t,u}(\boldsymbol\Delta)\), so the contraction principle \citep[Theorem~4.12]{LedouxTalagrand1991} applies.

	Let \(\{\epsilon_t^{\mathrm{rad}}\}\) be independent Rademacher variables, and let \(\mathbb E_{\mathrm{rad}}\) denote expectation over them conditional on the data.
	For \(u\in\mathcal U_n\), write \(A_u\) for the Rademacher supremum
	\begin{align*}
	A_u
	&:=
	\sup_{\boldsymbol\Delta\in\mathcal D_k}
	\left|\sum_{t\in I_k}\epsilon_t^{\mathrm{rad}}
	\Delta K_{t,u}(\boldsymbol\Delta)\right|.
	\end{align*}
The same contraction principle and \(\ell_1/\ell_\infty\) duality give
\begin{equation*}
\mathbb E_{\mathrm{rad}}A_u
\le\frac{CR_{1,k}}{b}\mathbb E_{\mathrm{rad}}\left\|
\sum_{t\in I_k}\epsilon_t^{\mathrm{rad}}\mathbb I\{|w_t(\boldsymbol\theta_0)-u|\le5b\}y_t\mathbf x_t^0\right\|_\infty
\le\frac{CW_xR_{1,k}}{b}\sqrt{N_b\log\{2(d+1)\}}.
\end{equation*}
For fixed data and \(u\), \(A_u\) is a convex function of the Rademacher vector.
The following bound shows that its Euclidean Lipschitz constant is at most \(C\sqrt{N_b}\):
\[
|\Delta K_{t,u}(\boldsymbol\Delta)|
\le\frac{C}{b}|(\mathbf x_t^0)^\top\boldsymbol\Delta|\mathbb I\{|w_t(\boldsymbol\theta_0)-u|\le5b\}
\le C\mathbb I\{|w_t(\boldsymbol\theta_0)-u|\le5b\}.
\]
	The convex concentration inequality for bounded product variables \citep[Theorem~6.10]{BoucheronLugosiMassart2013} controls each \(A_u\) around its conditional mean.
	Applying it to the Rademacher vector and taking a union bound over \(\mathcal U_n\) gives
	\begin{equation*}
	\mathbb E_{\mathrm{rad}}\max_{u\in\mathcal U_n}A_u
	\le
	C\sqrt{N_b}
	\left\{
		\frac{W_xR_{1,k}}{b}\sqrt{\log\{2(d+1)\}}
	+
	\sqrt{\log\{2|\mathcal U_n|\}}
	\right\}.
	\end{equation*}
The standard symmetrization inequality for independent observations \citep[Lemma~11.4]{BoucheronLugosiMassart2013}, \eqref{eq:localized-window-occupancy-mean}, and the preceding bound now imply
\begin{equation}
\mathbb E\max_{u\in\mathcal U_n}\sup_{\boldsymbol\Delta\in\mathcal D_k}
\left|\frac1n\sum_{t\in I_k}\frac{\Delta K_{t,u}(\boldsymbol\Delta)-\mathbb E\{\Delta K_{t,u}(\boldsymbol\Delta)\}}{b^{j+1}}\right|
\le C\left\{\frac{R_{1,k}\sqrt{q_k\log\{2(d+1)\}}}{\sqrt n\,b^{j+3/2}}
+\sqrt{\frac{q_k\log\{2|\mathcal U_n|\}}{nb^{2j+1}}}\right\}.
\label{eq:plugin-residual-expected-supremum}
\end{equation}

The last display controls the expected supremum.
To obtain a high-probability bound, consider the symmetric class consisting of \(\pm b^{-(j+1)}\Delta K_{t,u}(\boldsymbol\Delta)\), parameterized by \(u\in\mathcal U_n\) and \(\boldsymbol\Delta\in\mathcal D_k\).
Its envelope is at most \(Cb^{-(j+1)}\).
Either kernel term can be nonzero only on an interval of conditional probability \(O(b)\).
Hence \eqref{eq:localized-window-probability} gives
	\begin{equation*}
	\sup_{u\in\mathcal U_n,\,\boldsymbol\Delta\in\mathcal D_k}
	\frac1n\sum_{t\in I_k}\mathbb E\left\{
	\left[\frac{\Delta K_{t,u}(\boldsymbol\Delta)}{b^{j+1}}\right]^2
	\right\}
	\le \frac{Cq_k}{b^{2j+1}}.
	\end{equation*}
	By continuity in \(\boldsymbol\Delta\), the supremum can be taken over a countable dense subset of \(\mathcal D_k\).
	The concentration inequality of \citet[Theorem~1.1(c)]{KleinRio2005}, together with \(2\sqrt{ab}\le a+b\), shows that, except on an event of probability at most \(\delta/12\),
	the supremum in \eqref{eq:plugin-residual-expected-supremum} is at most
	\begin{equation}
	C\left\{
	\frac{R_{1,k}\sqrt{q_k\log\{2(d+1)\}}}{\sqrt n\,b^{j+3/2}}
	+
	\sqrt{\frac{q_k\log\{2|\mathcal U_n|\}}{n b^{2j+1}}}
	+
	\sqrt{\frac{q_k\log(12/\delta)}{n b^{2j+1}}}
	+
	\frac{\log(12/\delta)}{n b^{j+1}}
	\right\},
\label{eq:plugin-residual-bousquet}
		\end{equation}

Combining \eqref{eq:fine-residual-grid-remainder}, \eqref{eq:anchor-grid-process-bound}, and \eqref{eq:plugin-residual-bousquet} gives, outside an event of probability \(\delta/6\),
\begin{equation*}
\sup_{u\in I_z,\,\boldsymbol\theta\in\Theta_k}|Z_j(u,\boldsymbol\theta)|
\le C\left\{
\sqrt{\frac{q_k\log\{96(d+1)n/(b\delta)\}}{nb^{2j+1}}}
+\frac{R_{1,k}\sqrt{q_k\log\{2(d+1)\}}}{\sqrt n\,b^{j+3/2}}
+\frac{\log\{48n/(b\delta)\}}{nb^{j+1}}
\right\}.
\end{equation*}
The localization condition \(W_xR_{1,k}\le b\) absorbs the \(R_{1,k}\)-term.
The effective-sample condition absorbs the Bernstein linear term into the square-root term and implies \(\log(1/b)\lesssim\log n\).
Consequently,
\begin{equation*}
\sup_{u\in I_z,\,\boldsymbol\theta\in\Theta_k}|Z_j(u,\boldsymbol\theta)|
\le C\sqrt{\frac{q_k\log\{12(d+1)n/\delta\}}{nb^{2j+1}}}.
\end{equation*}
This process is \(h_k^{(j)}-\mathbb Eh_k^{(j)}\), up to the sign \((-1)^j\).
Repeating the argument with \(y_t\) replaced by \(1\) gives the corresponding bound for \(f_k^{(j)}-\mathbb Ef_k^{(j)}\).
Each of the six estimator--derivative combinations fails with probability at most \(\delta/6\).
Since \(n=a_k\) and \(\log\{12(d+1)n/\delta\}\le C H_k(\delta)^2\), a union bound over these combinations proves \eqref{eq:kernel-stochastic-unified}.
The effective-sample threshold and the deviation multiplier obtained above depend only on the primitive regularity bounds and the kernel;
choose \(B_{\mathrm{emp}}\) and \(C_{\mathrm{emp}}\) to dominate them, respectively.
\hfill\(\square\)

\subsection{Proof of Lemma~\ref{lem:sparse-confidence-reentry}}
\label{app:proof-sparse-confidence-reentry}

Write \(n=a_k\), let \(S=\supp(\boldsymbol\theta_0)\), and use \(\overline P_k,\beta_k,\sigma_k,e_k\) as defined in Appendix~\ref{app:proof-confidence-dynamics-main}.
Put \(\boldsymbol\Delta=\widehat{\boldsymbol\theta}_k-\boldsymbol\theta_0\), and write
\[
Y_t=\mathbf x_t^{0\top}\boldsymbol\theta_0+\beta_k(\mathbf x_t^0)+\epsilon_t,
\qquad
\mathbb E(\epsilon_t\mid \mathbf x_t^0,\mathcal F_{t-1})=0.
\]
Let \(D_k=\|\widehat{\boldsymbol\Sigma}_k-\overline{\boldsymbol\Sigma}_k\|_{\max}\).
Conditional Hoeffding--Azuma bounds, a union bound, and the penalty lower bound in Theorem~\ref{lem:sparse-valuation-main} imply that, with conditional probability at least \(1-\delta_k\),
\begin{equation}
2\left\|\frac1n\sum_{t\in I_k}\mathbf x_t^0\epsilon_t\right\|_\infty
+2\left\|\frac1n\sum_{t\in I_k}\left[
\mathbf x_t^0\beta_k(\mathbf x_t^0)
-\mathbb E\{\mathbf x_t^0\beta_k(\mathbf x_t^0)\mid\mathcal F_{t-1}\}
\right]\right\|_\infty
+2W_\theta D_k
\le\frac{\lambda_k}{4}.
\label{eq:misspecified-score-event}
\end{equation}
This concentration is history-uniform: all summands are conditionally centered and bounded under Assumptions~\ref{ass:bounded-sparse} and \ref{ass:covariance}.
Since both parameters belong to \(\Theta\), \(\|\boldsymbol\Delta\|_1\le2W_\theta\).
Penalized optimality, \eqref{eq:misspecified-score-event}, and \(|\overline P_k\{\beta_k(\mathbf x_t^0)\mathbf x_t^{0\top}\boldsymbol\Delta\}|\le\sigma_k e_k\) therefore give
\begin{equation}
e_k^2+\frac{3\lambda_k}{4}\|\boldsymbol\Delta_{S^c}\|_1
\le\frac{5\lambda_k}{4}\|\boldsymbol\Delta_S\|_1+2\sigma_ke_k.
\label{eq:misspecified-basic-inequality}
\end{equation}
Indeed, the empirical quadratic form is converted to its population version by
\[
e_k^2\le\boldsymbol\Delta^\top\widehat{\boldsymbol\Sigma}_k\boldsymbol\Delta
+D_k\|\boldsymbol\Delta\|_1^2,
\]
and, using \(\|\boldsymbol\Delta\|_1\le2W_\theta\), the last term is controlled together with the linear score term by \eqref{eq:misspecified-score-event}.

The prediction and \(\ell_1\) bounds now follow from the basic inequality.
Because \(\|\boldsymbol\Delta_S\|_1\le\sqrt{s_0}\,e_k/\sqrt{\lambda_{\min}}\), dropping the off-support term in \eqref{eq:misspecified-basic-inequality} yields
\[
e_k\le2\sigma_k+
\frac{5}{4\sqrt{\lambda_{\min}}}\sqrt{s_0}\lambda_k,
\]
which is \eqref{eq:misspecified-prediction-bound} with \(C_\Sigma=5/(4\sqrt{\lambda_{\min}})\).
Dropping instead the left quadratic term gives
\[
\|\boldsymbol\Delta\|_1
\le\frac83e_k\left(
\frac{\sqrt{s_0}}{\sqrt{\lambda_{\min}}}
+\frac{\sigma_k}{\lambda_k}\right)
\le C_{\mathrm{mis}}\left(s_0\lambda_k+\frac{\sigma_k^2}{\lambda_k}\right),
\]
where the last inequality uses the prediction bound and \(2ab\le a^2+b^2\).

To recover confidence-set coverage, take \(c_{\mathrm{ent}}=1/(4\sqrt{\lambda_{\min}})\).
Under \(\sigma_k\le c_{\mathrm{ent}}\sqrt{s_0}\lambda_k\),
\[
\begin{aligned}
\|\boldsymbol\Delta\|_2
&\le\frac{7}{4\lambda_{\min}}\sqrt{s_0}\lambda_k
<C_\theta\sqrt{s_0}\lambda_k,\\
\|\boldsymbol\Delta\|_1
&\le\frac{35}{6\lambda_{\min}}s_0\lambda_k
<4C_\theta s_0\lambda_k.
\end{aligned}
\]
Thus \(\boldsymbol\theta_0\in\mathcal C_k\) because \(C_\theta\ge3/\lambda_{\min}\).
\hfill\(\square\)

\subsection{Proof of Lemma~\ref{lem:finite-localization}}
\label{app:proof-finite-localization}

The first-crossing property of \(k_0\) in \eqref{eq:appendix-cutoff-definition} and the doubling schedule give \eqref{eq:cutoff-size}.
Since \(s_0\le d+1\), this bound also gives
\begin{equation}
\log\ell_{k_0}\le C\{1+\log(d+1)\},
\qquad
\log(d+1)+\log(1/\delta_{k_0})
\le C\{1+\log(d+1)\}.
\label{eq:tight-cutoff-log-comparison}
\end{equation}
Moreover, \(\Lambda_{\ell_{k_0}}\asymp1+s_0\log(d+1)\).
The first term in \eqref{eq:appendix-cutoff-definition} places the reference schedule past its initial full-exploration regime.
Consequently, the reference schedule in \eqref{eq:reference-exploration-schedule} satisfies, for every \(k\ge k_0\),
\[
a_k^{\mathrm{ref}}\asymp(\ell_k\Lambda_{\ell_k})^{\alpha_m},
\qquad
a_k^{\mathrm{ref}}\le a_k\le C_\omega a_k^{\mathrm{ref}}.
\]

We first verify the two statistical thresholds in Lemma~\ref{lem:finite-localization}.
Since \(b_k=(\omega_k/a_k)^{1/(2m+3)}\), schedule comparability and \(\omega_k\ge\omega_{\min}\) give
\[
a_kb_k^5
\gtrsim\bigl(a_k^{\mathrm{ref}}\bigr)^{\frac{2m-2}{2m+3}}
\asymp(\ell_k\Lambda_{\ell_k})^{\frac{2m-2}{4m+1}}.
\]
At \(k=k_0\), the first term in \eqref{eq:appendix-cutoff-definition}, together with \eqref{eq:tight-cutoff-log-comparison}, makes the last quantity an arbitrarily large fixed multiple of \(\Lambda_{\ell_{k_0}}\) as \(C_{\mathrm{cut}}\) increases.
Thus \(a_{k_0}b_{k_0}^5\ge M\Lambda_{\ell_{k_0}}\) after one fixed enlargement.

Similarly,
\[
a_kb_k^2
\gtrsim\bigl(a_k^{\mathrm{ref}}\bigr)^{\frac{2m+1}{2m+3}}
\asymp(\ell_k\Lambda_{\ell_k})^{\frac{2m+1}{4m+1}}.
\]
The second term in \eqref{eq:appendix-cutoff-definition} and the identities
\[
\frac{2m+1}{4m+1}\left(\frac{6m+1}{2m+1}+1\right)=2,
\qquad
\frac{2m+1}{4m+1}\left(\frac{2m}{2m+1}+1\right)=1
\]
give
\[
a_{k_0}b_{k_0}^2
\ge M s_0^2\{1+\log\ell_{k_0}+\log(d+1)\}
\]
after increasing \(C_{\mathrm{cut}}\) by another fixed factor.
The ratios in these two inequalities grow as positive powers of \(\ell_k\), up to logarithmic factors.
We may enlarge \(C_{\mathrm{cut}}\) once so that \(k_0\) lies beyond their fixed monotonicity thresholds; both inequalities then persist for every \(k\ge k_0\).

These thresholds verify the remaining statistical conditions.
The first, together with \(q_k\ge1/B\), \(H_k(\delta_k)^2\lesssim\Lambda_{\ell_k}\), and \eqref{eq:main-to-kernel-conditions}, controls the kernel stochastic and bias terms.
The second controls \(R_{2,k}\), and \eqref{eq:main-to-residual-localization} gives the two generated-residual localization conditions.
Together, the thresholds verify the effective sample-size, denominator, inverse-map, and derivative requirements, including their fixed interior margins.
Since \(b_k\le1\), the second threshold also yields
\[
a_k\ge M s_0^2\{1+\log\ell_k+\log(d+1)\}.
\]
Together with \(a_k^{\mathrm{ref}}\ge a_k/C_\omega\), this verifies the restricted-eigenvalue sample-size condition on the reference schedule and gives \(\varepsilon_{\mathrm{RE}}(a_k^{\mathrm{ref}})\le\delta_k\).
The penalty requirement follows from the fixed choice of \(c_\lambda\).
Hence all finite-sample conditions in Theorems~\ref{lem:sparse-valuation-main}--\ref{lem:noise-function-main} and Lemma~\ref{lem:pricing-function-main} hold for every \(k\ge k_0\).
Since \(q_F\) is fixed, one final fixed enlargement ensures \(k_0\ge q_F\).
\hfill\(\square\)

\subsection{Proof of Lemma~\ref{lem:truthful-prefix-prediction}}
\label{app:proof-truthful-prefix}

The decision to explore at time \(t\) is measurable before the current context is observed, so each \(T_q\) is predictable, with \(\{T_q=t\}\in\mathcal F_{t-1}\).
The matrix \(\boldsymbol\Sigma_{T_q}\) is \(\mathcal F_{T_q-1}\)-measurable and satisfies \(\mathbb E[\mathbf x_{T_q}^0\mathbf x_{T_q}^{0\top}\mid\mathcal F_{T_q-1}]
=\boldsymbol\Sigma_{T_q}\).
Hence the tower property implies that every entry of \(\mathbf x_{T_q}^0\mathbf x_{T_q}^{0\top}-\boldsymbol\Sigma_{T_q}\) is a bounded martingale difference with respect to \(\{\mathcal F_{T_q}\}_{q\ge1}\).
Entrywise Hoeffding--Azuma bounds and a union bound over the \((d+1)^2\) entries and all \(N\ge n_0\) give
\[
\mathbb P\{\mathcal Q_N^c\text{ for some }N\ge n_0\}
\le
2(d+1)^2\sum_{N\ge n_0}
\exp\left\{-\frac{N}{512W_x^4s_0^2}\right\}
\le\varepsilon_Q(n_0).
\]
This proves \eqref{eq:truthful-prefix-probability}.
If \(\mathbf v\in\mathcal C(S,3)\), then \(\|\mathbf v\|_1\le4\sqrt{s_0}\|\mathbf v\|_2\).
On \(\mathcal Q_N\), this cone inequality gives
\[
\frac1N\sum_{q=1}^N(\mathbf v^\top\mathbf x_q^0)^2
\le \mathbf v^\top\overline{\boldsymbol\Sigma}^{(N)}\mathbf v+
\|\mathbf v\|_1^2\left\|
\frac1N\sum_{q=1}^N\mathbf x_q^0\mathbf x_q^{0\top}-\overline{\boldsymbol\Sigma}^{(N)}
\right\|_{\max}
\le(\lambda_{\max}+1)\|\mathbf v\|_2^2,
\]
using \(\lambda_{\max}(\overline{\boldsymbol\Sigma}^{(N)})\le\lambda_{\max}\).
This proves \eqref{eq:truthful-prefix-prediction}.
\hfill\(\square\)
\endgroup
\end{document}